\documentclass[11pt]{article}
\usepackage[preprint]{acl}
\usepackage{iftex}
\ifXeTeX
\usepackage{fontspec}
\usepackage[english]{babel}
\babelprovide[import]{hebrew}
\usepackage{ucharclasses}
\newfontfamily\cjkfont[NFSSFamily=piicjkfont]{fonts/DroidSansFallbackFull.ttf}
\setTransitionsForCJK{\begingroup\cjkfont}{\endgroup}
\newfontfamily\koreanfont[NFSSFamily=piikoreanfont]{fonts/NanumGothic.ttf}
\setTransitionTo{HangulSyllables}{\begingroup\koreanfont}
\setTransitionFrom{HangulSyllables}{\endgroup}
\newfontfamily\symbolfont[NFSSFamily=piisymbolfont]{fonts/NotoSansMono-Regular.ttf}
\setTransitionTo{MiscellaneousMathematicalSymbolsA}{\begingroup\symbolfont}
\setTransitionFrom{MiscellaneousMathematicalSymbolsA}{\endgroup}
\fi
\usepackage{microtype}
\usepackage{amsmath,amssymb,graphicx}
\setkeys{Gin}{keepaspectratio}
\usepackage{longtable,booktabs,array,calc}
\ifXeTeX\else
\usepackage{tabularx}
\fi

\usepackage{caption}
\usepackage{etoolbox}
\usepackage{placeins}
\newif\ifACLmainmatter
\ACLmainmattertrue
\newcounter{ACLmainfloat}
\newcommand{\ACLmarkmainfloat}{\ifACLmainmatter
  \stepcounter{ACLmainfloat}\label{acl-main-float-\arabic{ACLmainfloat}}\fi}
\AtBeginEnvironment{figure}{\ACLmarkmainfloat}
\AtBeginEnvironment{figure*}{\ACLmarkmainfloat}
\AtBeginEnvironment{table}{\ACLmarkmainfloat}
\AtBeginEnvironment{table*}{\ACLmarkmainfloat}
\ifXeTeX
\usepackage{zref-savepos}
\newwrite\ACLlayoutfile
\AtBeginDocument{%
  \immediate\openout\ACLlayoutfile=\jobname.layout
  \immediate\write\ACLlayoutfile{height=\number\textheight}%
  \immediate\write\ACLlayoutfile{width=\number\textwidth}%
  \immediate\write\ACLlayoutfile{column=\number\columnwidth}%
  \immediate\write\ACLlayoutfile{gap=\number\columnsep}%
  \immediate\write\ACLlayoutfile{left=\number\dimexpr1in+\hoffset+\oddsidemargin\relax}%
  \immediate\write\ACLlayoutfile{top=\number\dimexpr\paperheight-1in-\voffset-\topmargin-\headheight-\headsep\relax}%
  \immediate\write\ACLlayoutfile{baseline=\number\baselineskip}%
  \immediate\closeout\ACLlayoutfile
}
\else
\newcommand{\zsavepos}[1]{}
\fi
\makeatletter
\newif\ifACLwidetable
\newif\ifACLcontinuedtable
\ACLwidetabletrue
\ifXeTeX
\BeforeBeginEnvironment{longtable}{\ifACLwidetable\begin{table*}[tp]\begin{minipage}{\textwidth}\else\begin{table}[tp]\begin{minipage}{\columnwidth}\small\fi\ifACLcontinuedtable\ContinuedFloat\fi\edef\ACLrestorecolumn{\global\@colroom=\the\@colroom\relax\global\vsize=\the\vsize\relax}\@twocolumnfalse}
\AfterEndEnvironment{longtable}{\ACLrestorecolumn\end{minipage}\ifACLwidetable\end{table*}\else\end{table}\fi}
\fi
\makeatother
\usepackage{adjustbox}
\usepackage{fancyvrb}
\usepackage{fvextra}
\DefineVerbatimEnvironment{Highlighting}{Verbatim}{breaklines,commandchars=\\\{\}}
\usepackage{color}
\usepackage{fancyvrb}

\DefineVerbatimEnvironment{Highlighting}{Verbatim}{commandchars=\\\{\}}
\usepackage{framed}
\definecolor{shadecolor}{RGB}{241,243,245}
\newenvironment{Shaded}{\begin{snugshade}}{\end{snugshade}}

\newcommand{\DataTypeTok}[1]{\textcolor[rgb]{0.68,0.00,0.00}{#1}}

\newcommand{\FunctionTok}[1]{\textcolor[rgb]{0.28,0.35,0.67}{#1}}

\newcommand{\OtherTok}[1]{\textcolor[rgb]{0.00,0.23,0.31}{#1}}

\newcommand{\StringTok}[1]{\textcolor[rgb]{0.13,0.47,0.30}{#1}}

\AtBeginEnvironment{Shaded}{\columnwidth=\linewidth\raggedright}

\providecommand{\real}[1]{#1}
\providecommand{\toprule}{\hline}
\title{Strong Multilingual Privacy Tagging at Encoder Speed}
\author{Jonathan Graehl \\
  RWS Language Weaver \\
  \small\href{https://orcid.org/0009-0008-0255-0811}{ORCID: 0009-0008-0255-0811}}
\hypersetup{pdfauthor={Jonathan Graehl},pdftitle={Strong Multilingual
Privacy Tagging at Encoder Speed}}
\date{}
\makeatletter
\@ifpackageloaded{caption}{}{\usepackage{caption}}
\AtBeginDocument{%
\ifdefined\contentsname
  \renewcommand*\contentsname{Table of contents}
\else
  \newcommand\contentsname{Table of contents}
\fi
\ifdefined\listfigurename
  \renewcommand*\listfigurename{List of Figures}
\else
  \newcommand\listfigurename{List of Figures}
\fi
\ifdefined\listtablename
  \renewcommand*\listtablename{List of Tables}
\else
  \newcommand\listtablename{List of Tables}
\fi
\ifdefined\figurename
  \renewcommand*\figurename{Figure}
\else
  \newcommand\figurename{Figure}
\fi
\ifdefined\tablename
  \renewcommand*\tablename{Table}
\else
  \newcommand\tablename{Table}
\fi
}
\@ifpackageloaded{float}{}{\usepackage{float}}
\floatstyle{ruled}
\@ifundefined{c@chapter}{\newfloat{codelisting}{h}{lop}}{\newfloat{codelisting}{h}{lop}[chapter]}
\floatname{codelisting}{Listing}

\makeatother
\makeatletter
\@ifpackageloaded{caption}{}{\usepackage{caption}}
\@ifpackageloaded{subcaption}{}{\usepackage{subcaption}}
\makeatother
\ifXeTeX
\newenvironment{RTL}{\begin{flushright}\beginR}{\endR\end{flushright}}
\else
\newenvironment{RTL}{\begin{flushright}}{\end{flushright}}
\newenvironment{otherlanguage}[1]{}{}
\fi
\begin{document}
\maketitle
\begin{abstract}
Privacy redaction must reliably remove personal information while
preserving the relationships expressed in the text. We develop a
general-purpose multilingual named-entity tagger with distinctions
fine-grained enough to support a broad range of redaction policies, and
methods for cheaply learning additional distinctions. We fine-tune the
multilingual encoder XLM\nobreak\hbox{-}\nobreak{}R-large on text in 35
languages annotated by prompted frontier language models, plus existing
human-annotated corpora with labels mapped onto our entity types. We
evaluate on human-annotated test sets in seven languages and on
LLM-annotated text in 35 languages. On the human-annotated sets, at
operating points fixed on development data, our tagger reaches 88.2
redaction F1, against 69.1 for published GLiNER2 and 61.4 for Microsoft
Presidio, and its architecture runs 4.9 times faster than GLiNER2's on
CPU. Compared with frontier-model annotations alone, adding the mapped
human-annotated corpora to training raised redaction F1 on those sets by
ten points, with essentially no change on the LLM-annotated text.
Open-weight LLMs fitting on one 96-GB GPU were less accurate than our
tagger, whether prompted to tag text or used as encoders beneath a
trained tagging layer, and roughly 30 to 1,100 times slower. We release
code, prompts and training recipes, with data-acquisition scripts and
source links.\footnote{\url{https://github.com/graehl/multilingual-pii-span-tagging},
  \href{https://github.com/graehl/multilingual-pii-span-tagging/tree/v1.5}{release
  v1.5}.}
\end{abstract}

\section{Introduction}\label{improving-redaction-with-task-matched-training-data}

Redaction must remove personal information such as names, addresses and
identifiers from text in many languages while leaving the rest readable.
What must be removed depends on the policy: a court judgment, a clinical
note and a support ticket call for different entity types, so a reusable
tagger needs distinctions fine enough for each policy to choose among
(Figure~\ref{fig-ont-inventory}). Published open-weight taggers are
useful starting points, but their entity types, languages and annotation
conventions often differ from the intended application. We compare our
tagger with these published models (incumbents) across 35 languages.

We fine-tune XLM\nobreak\hbox{-}\nobreak{}R-large
\citep{conneau2020-xlmr} with an affine token-classification head. Its
training data combine silver labels, produced by prompting frontier
LLMs, with gold human-annotated corpora whose labels are mapped to ours
(mapped gold). The decoder combines evidence from related entity types,
and a fitted ±1-character span-boundary model repairs subword
granularity errors (Section~\ref{sec-character-boundaries}). With each
system's decision threshold fixed on development data, our tagger
reaches 88.2 F1 over redacted regions of human-annotated test data in
seven languages, against 69.1 for published GLiNER2
\citep{zaratiana2026-gliner2-pii} and 61.4 for Microsoft Presidio
(Section~\ref{sec-priority9-comparison}), and its architecture runs 4.9
times faster than GLiNER2's on CPU
(Section~\ref{sec-deployment-efficiency}).

\section{Related work}\label{sec-related-work}

Our comparisons include the span-extraction model GLiNER2-PII
\citep{zaratiana2026-gliner2-pii}, OpenAI Privacy Filter
\citep{openai2026-privacy-filter-card}, OpenMed's multilingual
\citep{openmed2026-privacy-filter-multilingual-v2} and Nemotron-v2
fine-tunes of it, and Microsoft Presidio, which combines pattern
recognizers with spaCy entity models. Their claimed coverage is seven
languages for GLiNER2-PII, primarily English for OpenAI Privacy Filter,
and 16 and 20 languages for the two OpenMed variants, respectively. We
score every system on all seven languages of our gold evaluation, run
Presidio in each language's configuration, and align label definitions.
GLiNER2 is the strongest published incumbent and our recurring
comparison across experiments.

Prior work also motivates task-specific encoder fine-tuning. PIIBench
reports poor exact-span transfer from existing systems, while a
follow-on study obtains stronger results with a DeBERTa token classifier
trained for its labels \citep{jha2026-piibench, jha2026-deberta-pii}. In
a separate study, XLM\nobreak\hbox{-}\nobreak{}R fine-tuned on roughly
100--1,000 examples, depending on domain, surpasses the zero-shot OpenAI
Privacy Filter under a looser span-overlap criterion
\citep{uppala2026-opf-eval}.

Our approach builds on multilingual encoder pretraining
\citep{conneau2020-xlmr}, translation-based annotation transfer
\citep{chen2023-easyproject, garciaferrero2023-tprojection}, and
learning from incomplete annotations
\citep{mayhew2019-partial-ner, effland2021-expected-entity-ratio}.
FiNERweb provides a precedent for multilingual annotation by language
models \citep{golde2026-finerweb}. Our masked-language-model experiments
draw on domain-adaptive pretraining and efforts to preserve
cross-lingual capability during continued training
\citep{gururangan2020-dont-stop, liu2021-preserving-crosslinguality}.

\section{Data and labels}\label{sec-data}

\subsection{Label inventories and
alignment}\label{learning-and-evaluating-across-annotation-conventions}

\begin{figure*}[tp]

\centering{

\ifXeTeX
\includegraphics[width=1\linewidth,height=6in,alt={Packed circles group the 29 shared primary labels into 12 families; person\_reference and organization\_reference are marked as Ont3 additions.}]{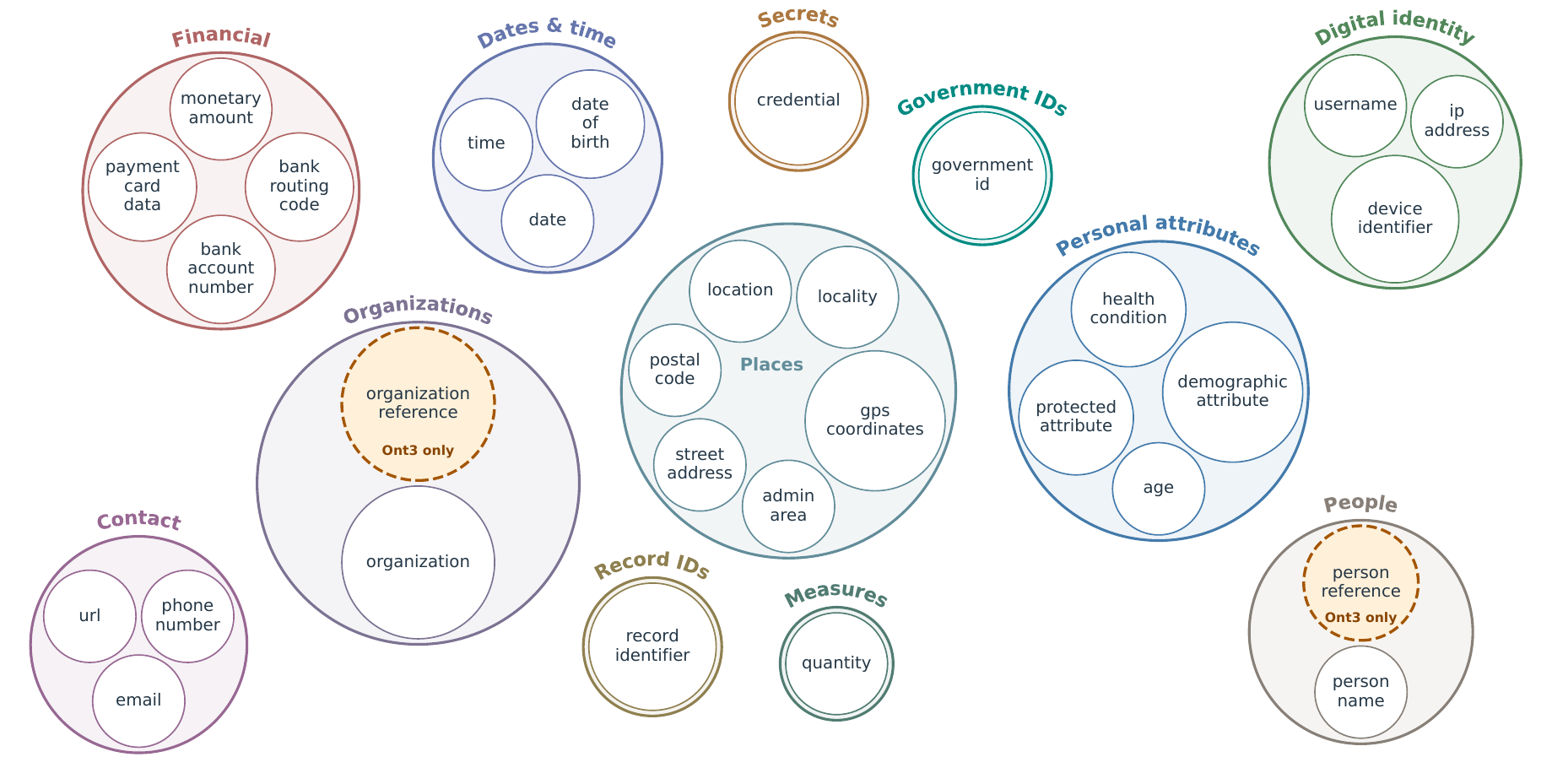}
\else
\includegraphics[alt={Packed circles group the 29 shared primary labels into 12 families; person\_reference and organization\_reference are marked as Ont3 additions.}]{figures/ont2-ont3-nested-families-inline-v1.pdf}
\fi

}

\caption{\label{fig-ont-inventory}Primary labels shared by Ont2 and
Ont3, grouped by semantic family. Dashed amber circles mark Ont3's two
optional reference labels; there are no Ont2-only primary labels.
Identifier-kind and name-component refinements are in Appendix
Figure~\ref{fig-ont-primary}.}

\end{figure*}

Our training and evaluation corpora use different label sets, so we
built three successive inventories. \textbf{Ont1} normalized label
spelling across the available gold corpora and kept every distinction
that was not clearly one-to-one as a separate label, giving 116 labels
(Appendix Figure~\ref{fig-ont1-union}). These labels were not disjoint
subclasses: a clinician's surname combines a person role with a name
component. \textbf{Ont2} therefore has 29 primary entity types in 12
semantic families, with consistent annotation guidelines and with name
components and roles moved to separate refinements. \textbf{Ont3} keeps
those 29 types and adds two optional
\textbf{\texttt{\_\allowbreak{}reference} labels},
\texttt{person\_\allowbreak{}reference} and
\texttt{organization\_\allowbreak{}reference}, for expressions referring
to people or organizations rather than naming them
(Figure~\ref{fig-ont-inventory}).

Corpora and models differ in their entity types and span boundaries. For
training, a source label can supervise a declared set of acceptable
target labels. For evaluation, we map all models' predictions and the
gold labels to the same categories and report exclusions. For example,
the ADDRESS label of the MAPA corpus \citep{arranz2022-mapa} permits
locality, postal code and street address as targets, with one designated
fallback (Appendix Figure~\ref{fig-gold-mappings}). The accepted
interpretations must agree between training and evaluation: a label
permitted by the training objective should not become an error solely
because evaluation narrows that set.

\begin{figure}[tp]

\centering{

\ifXeTeX
\includegraphics[width=1\linewidth,height=6in,alt={Given and family names are subspans of Dr Maria Delgado. The care-provider role attaches to the whole person name.}]{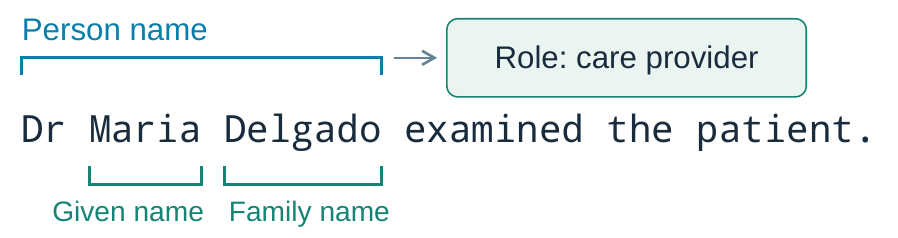}
\else
\includegraphics[alt={Given and family names are subspans of Dr Maria Delgado. The care-provider role attaches to the whole person name.}]{figures/name-refinements-v1.pdf}
\fi

}

\caption{\label{fig-name-refinements}Components locate subspans of a
name; a role labels the whole name span (illustrative).}

\end{figure}

Component refinements identify subspans, such as the given name within a
person's full name. Role refinements label the whole name span with
contextual information, such as whether the person is a patient
(Figure~\ref{fig-name-refinements}).

\subsection{Training data}\label{sec-ontology-comparison}

Our first models were trained mainly on mapped gold: human annotations
converted to Ont2 types through reviewed label mappings (Appendix
Figure~\ref{fig-gold-mappings}). Their high scores on data from the
training corpora did not carry over to fresh legal text from the HUDOC
case-law database \citep{echr2026-hudoc}, which differed in annotation
conventions and document type (Section~\ref{sec-ont2-transfer}). For
Ont3 we therefore directly annotated general web text in 35 languages
from FineWeb \citep{penedo2024-fineweb} and FineWeb2
\citep{penedo2025-fineweb2}, plus HUDOC court judgments and PubMed
Central clinical case reports \citep{nlm2026-pmc-oa-subset}, using
frontier LLMs. Our silver evaluation text
(Section~\ref{sec-evaluation-data}) comes from these sources but is held
out from training. Appendix Table~\ref{tbl-resource-access} lists every
corpus we used, with its license.

\protect\phantomsection\label{language-specific-demonstration-shots-for-annotation}{}

\begin{figure}[tp]

\centering{

\ifXeTeX
\includegraphics[width=1\linewidth,height=6in,alt={Fixed English instructions give the rules, including one tag set per occurrence of each surface; a French demonstration shows the town name Maringues twice, untagged inside the organization name Mairie de Maringues and labeled locality after the postal code.}]{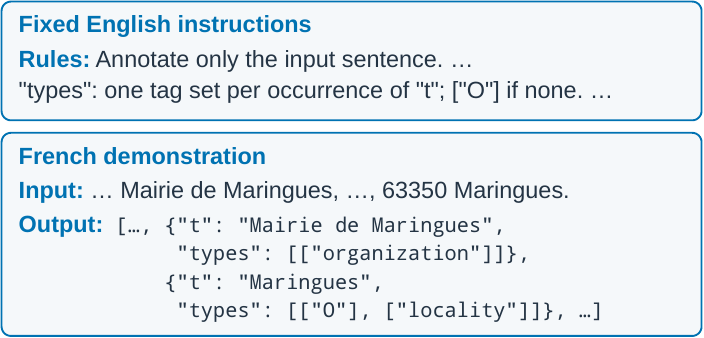}
\else
\includegraphics[alt={Fixed English instructions give the rules, including one tag set per occurrence of each surface; a French demonstration shows the town name Maringues twice, untagged inside the organization name Mairie de Maringues and labeled locality after the postal code.}]{figures/annotation-prompt-compact-v1.pdf}
\fi

}

\caption{\label{fig-annotation-prompts}Prompt assembly: fixed English
instructions plus a demonstration in the target language (French shown).
Ellipses omit other rules and annotations.}

\end{figure}

Each annotation prompt pairs shared definitions of 31 entity types (the
29 Ont2 types plus \texttt{organization\_\allowbreak{}reference} and
\texttt{person\_\allowbreak{}reference}) with worked examples in the
target language (Figure~\ref{fig-annotation-prompts}). The annotator
marks exact spans in one sentence, using its source paragraph to resolve
references (an annotated example: Section~\ref{sec-annotation-prompts}).

The annotator copies a surface and labels its occurrences; software
locates them in the input and computes the span offsets. This avoids
asking the model to count characters or reproduce the caller's offset
convention. The aligner checks the copied text and occurrence count.

\subsubsection{Adding mapped human
annotations}\label{reintroducing-mapped-gold-supervision}

Human annotations teach conventions that differ from those of frontier
annotators. We add MAPA, AQMAR \citep{mohit2012-aqmar}, OpenNER
\citep{palenmichel2025-openner} and Wojood \citep{jarrar2022-wojood} to
training through their label mappings. Each corpus annotates only some
of our types, so its unlabeled text counts as a negative only for types
that one of its labels maps to; we call these the types the corpus
covers.

\subsection{Evaluation data and scoring}\label{sec-evaluation-data}

We construct fresh evaluation data, \textbf{Silver}, from previously
unannotated text: 1,201 segments in 35 languages, including text
selected for rare entity types. Of these, 659 (\textbf{Silver-dev}) are
used for model selection and 542 (\textbf{Silver-test}) are not.
GPT-6-Luna \citep{openai2026-gpt6-luna} labeled most of Silver;
Silver-dev received costlier multi-model annotation
(Section~\ref{sec-evaluation-details}). We also evaluate on
\textbf{Gold-7}, 1,283 human-annotated segments from the test splits of
four public corpora in English, German, Spanish, French, Arabic,
Portuguese and Chinese, mapping their label inventories to shared
categories. Their presence in incumbents' training data is unknown.

We score \textbf{redaction F1} over untyped redacted regions: since
every type is redacted, a wrong type costs nothing, and corpora need not
agree on labels or on one span versus several adjacent ones. We sum
matched, predicted and gold region counts rather than averaging F1.
Scoring excludes corpus-unannotated types; name-kind subspans are not
scored. Optional reference labels are neutral in scoring
(Section~\ref{sec-span-views}). To tolerate small boundary misses, a
match must cover at least 80\% of both redaction regions, with each
region matched at most once. \textbf{Exact-boundary} redaction F1
requires identical boundaries, \textbf{character F1} scores redacted
characters, and \textbf{typed-span F1} requires identical boundaries and
the same entity type.

Our checks of silver labels against gold use the exact-boundary and
character variants, through corpus-specific many-to-many label mappings
(Appendix Figure~\ref{fig-gold-mappings}). We regard frontier-model
agreement around 90\% with human annotation as sufficient for this role.
For comparison, human annotators of the 274 multiply annotated
court-case documents in the Text Anonymization Benchmark (TAB) agree at
82.6\% typed-span F1 on TAB's own entity types \citep{pilan2022-tab}.

We drop training candidates whose lexical or embedding nearest neighbors
in the evaluation text are near-duplicates.

\section{Method}\label{sec-method}

\subsection{Token classification and
decoding}\label{affine-token-classification-with-bioes-span-tags}

Our model uses XLM\nobreak\hbox{-}\nobreak{}R-large
\citep{conneau2020-xlmr} and applies one affine layer to each token's
final encoder representation. Its scores describe both entity type and
position within a span: \textbf{B}eginning, \textbf{I}nside,
\textbf{O}utside, \textbf{E}nd, or \textbf{S}ingle token (BIOES). A
constrained decoder combines these scores into labeled spans. Adding a
constant bias to the outside (\textbf{O}) label's score before decoding
predicts fewer entities as it rises, trading recall for precision as a
confidence threshold does.

Our rationale for a simple head is that full encoder fine-tuning already
provides trainable nonlinear depth for the task. All heads use 10\%
dropout. We compared seven alternatives to the final-layer affine head:
four layer-concatenation variants, a nonlinear GELU head with layer
normalization, and two rank-constrained heads. None gave a reproducible
gain. The best fed the classifier a middle encoder layer's states
alongside the final layer's; its initial gain of 1.6 validation F1
points became a 0.3-point loss on repeat, so we treated it as selection
noise and retained the final-layer affine head.\footnote{The seven
  alternatives exclude exploratory runs;
  Section~\ref{sec-head-alternatives} gives the full search and
  corrected comparison conditions.}

Separate small heads, trained on the same annotations, predict
refinements such as name components and an identifier's kind (passport
or tax number). Most published taggers lack such outputs, and we do not
score them; they show that one encoder can add per-type outputs beyond a
single BIOES tag set (Appendix Figure~\ref{fig-ont-primary}).

\subsection{Coarse--fine
decoding}\label{coarsefine-aware-constrained-decoding}

Constrained BIOES decoding selects the highest-scoring legal sequence,
requiring the same entity type throughout a span; over per-token scores
it performed comparably to a CRF on three of four datasets
\citep{lester2020-constrained-decoding}. Our class-aware variant allows
related fine types to support a shared boundary: a hospital name can
receive alternating support for healthcare organization and generic
organization. For each boundary position, it combines scores within a
compatible broad class using a temperature-controlled log-mean-exp,
normalized by the number of member types. It then finds the
highest-scoring legal sequence over those broad classes, retaining the
highest-scoring fine member at each token. Within each selected span,
character-weighted votes from those token labels determine its final
fine type. That type need never win as a complete, matching-type BIOES
path on its own. For a fixed label inventory, decoding scales linearly
with token count.

For token \(t\), boundary position \(b\in\{B,I,E,S\}\), compatible
fine-type group \(C\), and temperature \(T>0\), the shared score is

\[
G_{t,b,C}=T\log\left(\frac{1}{|C|}\sum_{f\in C}
\exp\left(\frac{z_{t,b,f}}{T}\right)\right),
\]

where \(z_{t,b,f}\) is the encoder head's logit for that boundary and
type. Relative to matching-type Viterbi, coarse-class decoding raises
typed overlap F1 by 0.3 points when class scores take the maximum, and
by 0.5 with the log-mean-exp above. Moving each label's score 95\% of
the way toward its class's shared score before ordinary BIOES decoding
performs about the same (Section~\ref{sec-decoder-contrast}).

\subsection{Character-boundary
refinement}\label{sec-character-boundaries}

A subword classifier can place span endpoints only at token boundaries,
which need not coincide with the text that should be redacted. For
example, the token \texttt{2.} can join an identifier's final digit to
sentence punctuation: redacting the digit while preserving the period
requires an endpoint inside that token. In languages with little
whitespace, including Chinese and Japanese, subwords can also cross
name-component boundaries. Only 92\% of gold spans had both endpoints on
the tokenizer's boundaries in an early tokenizer audit
(Section~\ref{sec-boundary-refiner-details}).

To recover precise character spans without changing the encoder, we add
a small linear ranker after BIOES decoding, trained on gold endpoints
near the tagger's own span proposals. For each proposed endpoint it
scores the current position and its two one-character neighbors, using
the predicted Ont3 entity type, endpoint side and nearby characters. Its
\(2^{18}\) hashed weights occupy about 1 MB. A dynamic program chooses
endpoint pairs while preserving each span's type, proposal count and
uniqueness, and preventing new overlaps between previously disjoint
spans.

On our final model (O4, Section~\ref{sec-training-recipe}), refinement
raises exact typed-span F1 from 74.6 to 76.3 on Silver: +1.68 points
(paired 95\% interval {[}+1.01, +2.41{]}). Exact-boundary redaction F1
rises from 77.7 to 80.2; on Gold-7 it is essentially unchanged, 87.87 to
87.80 (Section~\ref{sec-boundary-refiner-details}).

Word-level pooling combines subword representations before labeling
\citep{acs2021-subword-pooling}. Our refiner instead keeps the tagger's
subword sequence and learns character endpoints, including positions
inside tokens. A deterministic English word-edge rule was competitive on
the native development subset but less reliable on punctuation-bearing
identifiers and names (Section~\ref{sec-boundary-refiner-details}).

\subsection{Training recipe}\label{sec-training-recipe}

Our first model, \textbf{O2}, uses the Ont2 inventory in 20 languages
and was trained mainly on mapped gold and translated data (Appendix
Table~\ref{tbl-ont2-ont3-methods}).

\textbf{O3} uses Ont3 in 35 languages and trains on silver text plus
40\% draws from mapped gold (Section~\ref{sec-ontology-comparison}). It
adds three training choices to the affine head, legal BIOES decoding and
±1-character span-boundary model: outside (\textbf{O}) tokens get loss
weight 0.75, counted only for types their corpus covers; each language
has a learned output bias; and training mixes in neighboring-sentence
inputs, which raised silver redaction F1 by 1.1 points (95\% interval
{[}+0.01, +2.18{]}; Section~\ref{sec-neighboring-context}).

Our final model, \textbf{O4}, adds 49,812 unique sentence-oriented
training texts with 378,431 labeled XLM\nobreak\hbox{-}\nobreak{}R
subword tokens to O3's silver data and raises the mapped-gold share to
50\%. Repricing the recorded annotation attempts, including retries and
rejected outputs, at current GPT-6-Luna rates gives about \$53
(Section~\ref{sec-o4-annotation-recipe}). With the added data,
neighboring context no longer helped, so O4 reads each sentence alone:
giving it the preceding sentence lowered its model-selection score (70\%
gold redaction F1; Section~\ref{sec-o4-annotation-recipe}) by 0.44
points {[}−0.90, 0.00{]}, and mixing contextual inputs into training
lowered it by 0.30. Register tokens, soft prompts and an extra
entity-versus-outside loss did not help either
(Section~\ref{sec-failed-ideas}).

\ACLwidetabletrue\ACLcontinuedtablefalse

\ifXeTeX
\begin{longtable}[]{@{}
  >{\raggedright\arraybackslash}p{(\linewidth - 8\tabcolsep) * \real{0.1667}}
  >{\raggedleft\arraybackslash}p{(\linewidth - 8\tabcolsep) * \real{0.2222}}
  >{\raggedleft\arraybackslash}p{(\linewidth - 8\tabcolsep) * \real{0.2222}}
  >{\raggedleft\arraybackslash}p{(\linewidth - 8\tabcolsep) * \real{0.2222}}
  >{\raggedright\arraybackslash}p{(\linewidth - 8\tabcolsep) * \real{0.1667}}@{}}
\caption{Both models use Ont3 in 35 languages and train on the same
80,099 unique mapped human-annotated texts. Counts deduplicate language
and text before sampling; labeled tokens include optional
references.}\label{tbl-o4-data}\tabularnewline
\toprule\noalign{}
\begin{minipage}[b]{\linewidth}\raggedright
Recipe
\end{minipage} & \begin{minipage}[b]{\linewidth}\raggedleft
Unique silver texts
\end{minipage} & \begin{minipage}[b]{\linewidth}\raggedleft
Labeled tokens
\end{minipage} & \begin{minipage}[b]{\linewidth}\raggedleft
Mapped-gold draws
\end{minipage} & \begin{minipage}[b]{\linewidth}\raggedright
Input
\end{minipage} \\
\midrule\noalign{}
\endfirsthead
\toprule\noalign{}
\begin{minipage}[b]{\linewidth}\raggedright
Recipe
\end{minipage} & \begin{minipage}[b]{\linewidth}\raggedleft
Unique silver texts
\end{minipage} & \begin{minipage}[b]{\linewidth}\raggedleft
Labeled tokens
\end{minipage} & \begin{minipage}[b]{\linewidth}\raggedleft
Mapped-gold draws
\end{minipage} & \begin{minipage}[b]{\linewidth}\raggedright
Input
\end{minipage} \\
\midrule\noalign{}
\endhead
\bottomrule\noalign{}
\endlastfoot
O3 & 38,700 & 477,685 & 40\% & Neighboring sentences \\
O4 & 88,512 & 856,116 & 50\% & Isolated sentence \\
\end{longtable}
\else
\begin{longtable}[]{@{}
  >{\raggedright\arraybackslash}X
  >{\raggedleft\arraybackslash}X
  >{\raggedleft\arraybackslash}X
  >{\raggedleft\arraybackslash}X
  >{\raggedright\arraybackslash}X@{}}
\caption{Both models use Ont3 in 35 languages and train on the same
80,099 unique mapped human-annotated texts. Counts deduplicate language
and text before sampling; labeled tokens include optional
references.}\label{tbl-o4-data}\tabularnewline
\toprule\noalign{}
\raggedright
Recipe
 & \raggedleft
Unique silver texts
 & \raggedleft
Labeled tokens
 & \raggedleft
Mapped-gold draws
 & \raggedright
Input
 \\
\midrule\noalign{}
\endfirsthead
\toprule\noalign{}
\raggedright
Recipe
 & \raggedleft
Unique silver texts
 & \raggedleft
Labeled tokens
 & \raggedleft
Mapped-gold draws
 & \raggedright
Input
 \\
\midrule\noalign{}
\endhead
\bottomrule\noalign{}
\endlastfoot
O3 & 38,700 & 477,685 & 40\% & Neighboring sentences \\
O4 & 88,512 & 856,116 & 50\% & Isolated sentence \\
\end{longtable}
\fi

\subsection{Incomplete teacher
annotations}\label{learning-from-incomplete-teacher-annotations}

Ignoring unmarked tokens in sparse LLM annotations risks lowering
precision: false positives there incur no labeling loss. We denote their
non-entity (\textbf{O}) loss weight by \(\lambda\), ranging from
\(\lambda=0\) (ignored) to \(\lambda=1\) (ordinary negative
supervision), balancing this risk against learning the annotator's
omissions. Marked spans and fully annotated examples retain their usual
supervision. Separately, reducing the weight of known \textbf{O} targets
to 0.75 on fully annotated data improved Ont2 F1, primarily by shifting
the precision--recall operating point
(Section~\ref{sec-operating-points}); here we test whether unmarked
tokens should be treated as negatives at all.

We annotated 1,652 examples with Gemma\nobreak\hbox{-}\nobreak{}4-31B-IT
\citep{gemmateam2026-gemma4}, whose annotations reach about 80\% of
Luna's coverage on Ont3 text. With these examples at 10\% training
sampling mass, \(\lambda=0\) gave better span and character F1 than
\(\lambda\in\{0.05,0.20,1\}\): compared with \(\lambda=1\), it increased
character recall by 2.8 points and character F1 by 1.5 points, while
reducing precision by 0.3 points, in a two-seed contrast on the
twenty-language development set
(Section~\ref{sec-teacher-omission-controls}). Boundary conventions can
also mimic omissions, so the objective accepts declared internal name
splits or merges while retaining outer boundaries
(Section~\ref{sec-teacher-boundaries}).

\protect\phantomsection\label{sec-operating-point-control}
The training data mix, the background (\textbf{O}) loss weight and an
inference-time \textbf{O} bias all shift the model's propensity to label
rather than withhold. We favor the \textbf{O} loss weight: it
compensates during learning for a predictable mismatch between training
and deployment entity density without discarding useful examples, and it
can change the learned decision function rather than uniformly shifting
a fixed model's scores. The \textbf{O} bias needs no retraining, so we
use it for cheap operating-point sweeps. Evidence is in
Section~\ref{sec-operating-points}.

\subsection{Adapting GLiNER2}\label{adapting-gliner2-to-ont3}

GLiNER2 predicts only whole-word spans, so to transfer it to Ont3 we
snap training spans to word boundaries. We also keep familiar label
spellings, initialize new label tokens from related pretrained
embeddings, and shuffle the order of the entity types in each training
example, as the original GLiNER recipe does
\citep{zaratiana2024-gliner}. GLiNER2's trainer leaves this shuffling
off by default and its paper does not mention it
\citep{zaratiana2025-gliner2}; our first adaptation, without it,
collapsed after a few hundred updates
(Section~\ref{sec-gliner2-adaptation}). With shuffling, longer training
is stable but still costs some Gold-7 quality. GLiNER2's original
training data is not publicly released, so we could not mix it back in.
Of 13 checkpoints scored on Silver-dev, we select the one after 2,500 of
6,000 updates. Fine-tuning is effective for full Ont3 labeling: from the
published weights with transferred labels, which already name every Ont3
type, it raises exact-region F1 from 57.4 to 62.8 on Gold-7 (unused in
selection) and from 50.1 to 58.9 on Silver-dev. This model, GL4, reaches
65.1 redaction F1 on Gold-7 and 66.9 on Silver-test at its Silver-dev
threshold, against 69.1 and 72.9 for the published model. GL4's task is
harder: it must predict all of our types, whereas published GLiNER2 is
not scored on gold types it has no label for, an advantage no other
system receives.

\section{Results}\label{sec-results}

\subsection{Main comparison}\label{sec-priority9-comparison}

An \textbf{operating point} is the precision--recall tradeoff obtained
at a chosen final confidence threshold. We vary the \textbf{O} bias for
XLM\nobreak\hbox{-}\nobreak{}R, the confidence threshold for GLiNER2,
and the score threshold for Microsoft Presidio 2.2.364, using the
multilingual configuration described in
Section~\ref{sec-presidio-configuration}. Each system's operating point
is fixed on Silver-dev: among the settings whose redaction F1 there is
not reliably below the best (paired bootstrap over documents), we take
the center of the contiguous range, which is less sensitive to
development noise than the single best setting
(Section~\ref{sec-operating-points}).

\begin{figure*}[tp]

\centering{

\ifXeTeX
\includegraphics[width=1\linewidth,height=7.5in,alt={Gold-7 aggregate and seven language panels, then Silver-test\textquotesingle s seven-language subset and all 542 segments. Hollow rings mark each curve\textquotesingle s operating point fixed on Silver-dev; O4\textquotesingle s label sits at its ring.}]{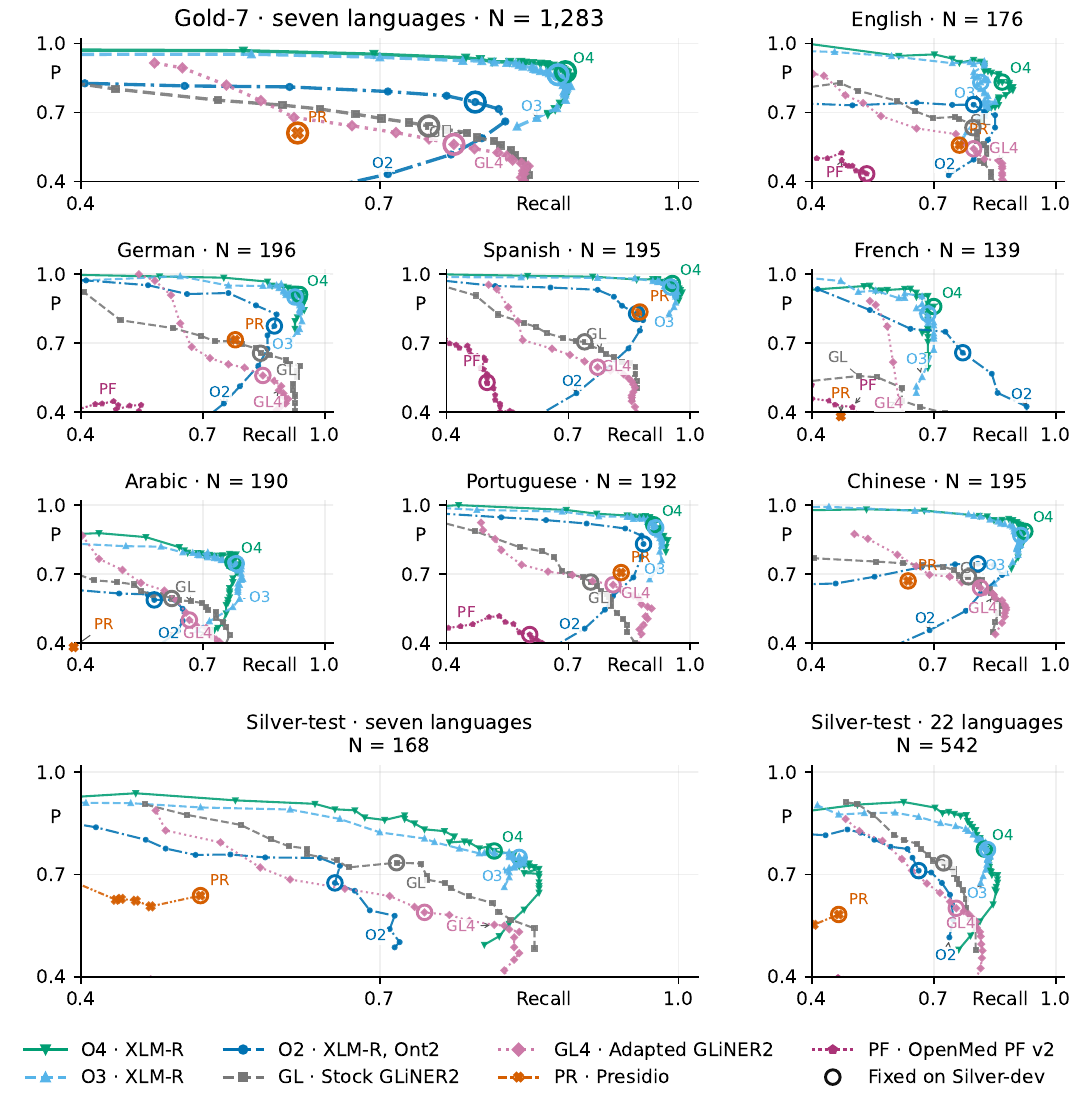}
\else
\includegraphics[alt={Gold-7 aggregate and seven language panels, then Silver-test\textquotesingle s seven-language subset and all 542 segments. Hollow rings mark each curve\textquotesingle s operating point fixed on Silver-dev; O4\textquotesingle s label sits at its ring.}]{figures/o4-human-ont3-overlap80-tr-v1.pdf}
\fi

}

\caption{\label{fig-final20-priority9-overlap}Precision--recall as each
system's confidence threshold or outside-label bias varies, on Gold-7
(top three rows: all seven languages, then each) and Silver-test (bottom
row: its 168 segments in the same seven languages at left, all 542
segments in 22 languages at right). Rings mark each system's operating
point, fixed on Silver-dev; O4 was selected on Silver-dev and Gold-7,
GL4 on Silver-dev. Untyped redaction regions match at 80\% overlap of
both (exact matches: Section~\ref{sec-evaluation-details}); name
components and other refinements are not scored. GL and GL4: published
and fine-tuned GLiNER2; O2--O4: XLM\nobreak\hbox{-}\nobreak{}R variants
(O2 on Ont2); PR: multilingual Presidio; PF: OpenMed multilingual-v2
Privacy Filter. Gold spans outside GL's label coverage are unscored for
GL alone; Appendix Figure~\ref{fig-gliner-categories} compares GL and O4
on the same restricted task. Both axes start at 0.4. At its ring, O4
reaches 88.2 F1 on Gold-7 and 79.9 on Silver-test.}

\end{figure*}

On Gold-7, with operating points fixed on Silver-dev, O4 reaches 88.2
redaction F1, against 69.1 for published GLiNER2 and 61.4 for Presidio
(Figure~\ref{fig-final20-priority9-overlap}). GL4 is GLiNER2 fine-tuned
on our training data; it reaches 65.1. These comparisons allow differing
annotation conventions, and published GLiNER2 is not scored on gold
types it has no label for (Section~\ref{sec-span-views}).

OpenMed's multilingual v2 fine-tune of OpenAI Privacy Filter, the
strongest of the five Privacy Filter variants we tested, reaches 37.0 F1
on Gold-7 and 43.8 on Silver-test at its fixed \textbf{O}
bias.\footnote{On Gold-7, multilingual v2 dominates the original filter
  and both Nemotron fine-tunes; multilingual v1 reaches at most 33.0 F1.
  Multilingual v2's model card lists all seven languages.

  The original filter intentionally omits organizations and general
  locations. Its task is narrower than the full redaction inventory;
  Appendix Figure~\ref{fig-privacy-filter-categories} gives its
  supported-category comparison.}

\subsection{Gains from mapped gold and added silver
data}\label{sec-training-results}

O3's 40\% mapped-gold training draw followed a matched experiment in
which it raised exact-boundary redaction F1 on gold by ten points
without reducing it on silver evaluation text
(Section~\ref{sec-mapped-gold-replay}).

With the added silver data (Table~\ref{tbl-o4-data}) and at fixed zero
\textbf{O} bias, O4 raises pooled typed-span F1 on Silver from O3's 74.5
to 76.3: +1.79 points (paired document-bootstrap 95\% interval {[}+0.51,
+3.13{]}). On Gold-7, exact-boundary redaction F1 rises from 86.5 to
87.8 (+1.25, interval {[}+0.19, +2.34{]}). Its gains on Silver are
smaller: 79.7 to 80.2 (+0.55, interval {[}−0.62, +1.74{]}).

We expect additional frontier annotation to improve performance further,
and encourage practitioners to adapt the ontology and annotate their own
domains.

\subsection{Encoder and LLM
alternatives}\label{choosing-the-pretrained-encoder}

XLM\nobreak\hbox{-}\nobreak{}R-large scored highest of three encoders
trained on the same 1,854 clinical and legal documents, with checkpoints
selected without using the evaluation sets
(Section~\ref{sec-encoder-screen}). With matched affine heads, it
exceeds LaBSE \citep{feng2022-labse} by 4.4 F1 points on a six-language
diagnostic set and by 2.5 on synthetic clinical records in ten
languages, two earlier evaluation sets. On O4's full training data, an
affine head lets mmBERT-base \citep{marone2025-mmbert}, the largest
released mmBERT, gain 2.1 Gold-7 exact-boundary redaction F1 points over
its stock head (95\% interval {[}0.95, 3.38{]}), but it remains below O4
on silver text (Appendix Table~\ref{tbl-mmbert-o4}). The larger
XLM\nobreak\hbox{-}\nobreak{}R-XL \citep{goyal2021-xlmr-xl} (3.5B
parameters), fully unfrozen, did not improve on
XLM\nobreak\hbox{-}\nobreak{}R-large on Silver-dev (−0.9 typed-span F1,
95\% interval {[}−2.72, +0.87{]}), plausibly because its pretraining
processed fewer tokens of the same corpus (0.5T against 6T).

Causal language models have served as encoders, feeding their
final-token state to a classifier, since LSTMs
\citep{dai2015-semi-supervised} and then Transformer decoders
\citep{radford2018-gpt1}; per-token entity tagging from their states
followed \citep{peters2017-taglm, liu2019-linguistic-knowledge}. Frozen
local LLMs fitting on a single 96-GB GPU fell short in this role: task
instructions and a 512-unit GELU head combining final-layer states at
token offsets {[}−1, 0, +1{]} improved
Qwen3.8-\textbf{Flash\nobreak\hbox{-}\nobreak{}Next}
\citep{qwenteam2026-qwen38-flash-next}, but the best frozen LLM on
Silver-dev, Gemma\nobreak\hbox{-}\nobreak{}4-31B base, still trails O4
there by 4.5 points of character F1 (87.1 against 91.5) and 14.5 points
of typed-span F1 (62.4 against 76.9);
Flash\nobreak\hbox{-}\nobreak{}Next reaches 59.5
(Section~\ref{sec-frozen-llm-heads}).

Prompted directly, local models produced invalid annotations on 5--8\%
of Silver-dev. Even rescuing those failures with O4, their character F1
remained 1.4--5.1 points below O4's 91.5. Their estimated throughput was
2--15 labeled tokens/s versus the encoder's measured 2,744---roughly two
to three orders of magnitude slower (hardware and timing details in
Section~\ref{sec-local-annotators}).

\subsection{Deployment efficiency}\label{sec-deployment-efficiency}

To approach prompted frontier-model annotation quality with inexpensive
inference, we use an encoder that assigns token labels in one forward
pass. O4 labels each sentence in isolation.

Throughput counts tokens in the text being labeled, including
outside-entity text, with one common whitespace-delimited convention.

We measured serving throughput for the
XLM\nobreak\hbox{-}\nobreak{}R-large architecture and stock GLiNER2 on
the same inputs under each configuration (Table~\ref{tbl-serving-cost};
conditions in Section~\ref{sec-serving-conditions}). The table also
reports LLM encoding throughput for the BIOES-head comparisons in
Section~\ref{sec-frozen-llm-heads}.

\ACLwidetablefalse\ACLcontinuedtablefalse

\ifXeTeX
\begin{longtable}[]{@{}
  >{\raggedright\arraybackslash}p{(\linewidth - 6\tabcolsep) * \real{0.3100}}
  >{\raggedright\arraybackslash}p{(\linewidth - 6\tabcolsep) * \real{0.3700}}
  >{\raggedleft\arraybackslash}p{(\linewidth - 6\tabcolsep) * \real{0.1800}}
  >{\raggedleft\arraybackslash}p{(\linewidth - 6\tabcolsep) * \real{0.1400}}@{}}
\caption{Throughput per labeled input token, excluding instructions and
context, and Gold-7 redaction F1 of O4, published GLiNER2 and the LLM
heads. CPU is EPYC 7R13 with 16 threads. XLM\nobreak\hbox{-}\nobreak{}R
and GLiNER2 use document batches on GPU
(Section~\ref{sec-serving-conditions}). The LLM rows measure encoding
for simple BIOES heads
(Section~\ref{sec-frozen-llm-heads}).}\label{tbl-serving-cost}\tabularnewline
\toprule\noalign{}
\begin{minipage}[b]{\linewidth}\raggedright
System
\end{minipage} & \begin{minipage}[b]{\linewidth}\raggedright
Hardware
\end{minipage} & \begin{minipage}[b]{\linewidth}\raggedleft
Tokens/s
\end{minipage} & \begin{minipage}[b]{\linewidth}\raggedleft
F1
\end{minipage} \\
\midrule\noalign{}
\endfirsthead
\toprule\noalign{}
\begin{minipage}[b]{\linewidth}\raggedright
System
\end{minipage} & \begin{minipage}[b]{\linewidth}\raggedright
Hardware
\end{minipage} & \begin{minipage}[b]{\linewidth}\raggedleft
Tokens/s
\end{minipage} & \begin{minipage}[b]{\linewidth}\raggedleft
F1
\end{minipage} \\
\midrule\noalign{}
\endhead
\bottomrule\noalign{}
\endlastfoot
XLM\nobreak\hbox{-}\nobreak{}R & CPU & 675 & 88.2 \\
GLiNER2 & CPU & 138 & 69.1 \\
XLM\nobreak\hbox{-}\nobreak{}R & L40S & 2,744 & 88.2 \\
GLiNER2 & L40S & 1,405 & 69.1 \\
Flash\nobreak\hbox{-}\nobreak{}Next & RTX PRO 6000 & 29 & 65.1 \\
Gemma\nobreak\hbox{-}\nobreak{}4-31B & RTX PRO 6000 & 93 & 56.5 \\
\end{longtable}
\else
\begin{longtable}[]{@{}
  >{\raggedright\arraybackslash}X
  >{\raggedright\arraybackslash}X
  >{\raggedleft\arraybackslash}X
  >{\raggedleft\arraybackslash}X@{}}
\caption{Throughput per labeled input token, excluding instructions and
context, and Gold-7 redaction F1 of O4, published GLiNER2 and the LLM
heads. CPU is EPYC 7R13 with 16 threads. XLM\nobreak\hbox{-}\nobreak{}R
and GLiNER2 use document batches on GPU
(Section~\ref{sec-serving-conditions}). The LLM rows measure encoding
for simple BIOES heads
(Section~\ref{sec-frozen-llm-heads}).}\label{tbl-serving-cost}\tabularnewline
\toprule\noalign{}
\raggedright
System
 & \raggedright
Hardware
 & \raggedleft
Tokens/s
 & \raggedleft
F1
 \\
\midrule\noalign{}
\endfirsthead
\toprule\noalign{}
\raggedright
System
 & \raggedright
Hardware
 & \raggedleft
Tokens/s
 & \raggedleft
F1
 \\
\midrule\noalign{}
\endhead
\bottomrule\noalign{}
\endlastfoot
XLM\nobreak\hbox{-}\nobreak{}R & CPU & 675 & 88.2 \\
GLiNER2 & CPU & 138 & 69.1 \\
XLM\nobreak\hbox{-}\nobreak{}R & L40S & 2,744 & 88.2 \\
GLiNER2 & L40S & 1,405 & 69.1 \\
Flash\nobreak\hbox{-}\nobreak{}Next & RTX PRO 6000 & 29 & 65.1 \\
Gemma\nobreak\hbox{-}\nobreak{}4-31B & RTX PRO 6000 & 93 & 56.5 \\
\end{longtable}
\fi

The encoder delivered 4.9 times GLiNER2's CPU throughput and about twice
its GPU throughput in these configurations.

\section{Open questions}\label{sec-open-questions}

\textbf{Encoder capacity.} A fully unfrozen
XLM\nobreak\hbox{-}\nobreak{}R-XL did not improve on
XLM\nobreak\hbox{-}\nobreak{}R-large, but its pretraining processed
about a twelfth as many tokens of the same corpus. Larger, differently
pretrained or newer encoders, especially with more annotated data, may
still raise the ceiling.

\textbf{LLM representations.} Head and prompt changes improved taggers
on frozen LLM states, but the best, on
Gemma\nobreak\hbox{-}\nobreak{}4-31B base, was slower than the
XLM\nobreak\hbox{-}\nobreak{}R BIOES tagger and 4.5 character F1 worse
(87.1 against 91.5; Section~\ref{sec-frozen-llm-heads}). We did not
exhaust head and prompt choices; additional LLM layers as input and
continued LLM fine-tuning remain untested.

\textbf{Richer heads and labels.} An affine BIOES head reading only the
current token state, decoded with Viterbi, sufficed here, and deeper
heads or added encoder layers did not help. Joint span scoring as in
GLiNER2 could still add information. Can reliable subtype labels improve
coarse redaction beyond what a larger head learns from coarse labels
alone?

\textbf{Cheaper supervision.} At a fixed investment in clean Ont3
annotation, could noisier data, or data labeled under other ontologies,
admitted through partial supervision and label mappings, do better?
Adding mapped human annotations from corpora with other labeling schemes
already helps (Section~\ref{sec-ontology-comparison}).

\textbf{Per-language capacity.} If a more diverse training mixture
lowers quality on important languages, can automatically splitting
tagging heads or selected encoder parameters recover it while keeping
broad coverage?

\textbf{Cross-locale names.} Does a shared multilingual head still find
a name that is atypical of the surrounding text's language or locale?
Rerunning unchanged sentences with names from other locales would test
this directly.

\textbf{Machine translation.} Translating human-annotated data helped
extend coverage to 35 languages (Section~\ref{sec-transport-controls}),
but mismatched entity names across locales remained a quality constraint
(Section~\ref{sec-locale-repairs}). We found no easy remedy: achieving
adequate translation and entity quality cost about as much as annotating
native text with language-specific prompt demonstrations, which avoids
this mismatch. For which other tasks is cheap, high-quality translation
of annotated data sufficient to extend a multilingual encoder across
languages? Cheap, high-quality native annotation may leave less benefit
for more elaborate methods.

\section{Conclusion}\label{conclusion}

Task-matched supervision from prompted frontier models, combined with
human-annotated corpora whose labels are mapped to ours, trains an
XLM\nobreak\hbox{-}\nobreak{}R-large encoder with an affine BIOES head
(and a fitted ±1-character span-boundary model to repair subword
granularity errors) that reaches 88.2 redaction F1 on Gold-7 against
69.1 for published GLiNER2, at operating points fixed on Silver-dev.
Additional annotations and a larger share of mapped human annotations
improve typed-span F1, while gains on already strong redaction scores
are smaller. The encoder architecture delivers 4.9 times GLiNER2's CPU
throughput.

\FloatBarrier

\zsavepos{acl-main-end-position}\label{acl-main-text-end}
\ACLmainmatterfalse \label{acl-limitations-start}

\section{Limitations}\label{limitations}

Annotation variation affects exact-boundary scores. Across the 274 TAB
court-case documents with multiple human annotations
\citep{pilan2022-tab}, annotator pairs agree at 82.6\% typed-span F1 on
TAB's own entity types, pooled over all ordered pairs.

We release code, prompts and training recipes, with data-acquisition
scripts and source links, but not training data or weights. Access
conditions and training costs are in Section~\ref{sec-release-boundary}.

\section{Ethical considerations}\label{sec-ethical-considerations}

\textbf{Dual use.} A tagger that finds names, identifiers, contact
details and locations in 35 languages supports privacy protection and
equally supports targeted surveillance or profiling of the same text. We
release training recipes and pointers to the training data
(Section~\ref{sec-release-boundary}); the capability itself is not
specific to this work and is already available in the published models
we compare against.

\textbf{Annotation-service exposure.} Our training supervision comes
from prompted frontier language models hosted by external providers.
Annotating confidential text this way exposes that text to the
provider's session-data handling, and a redaction project cannot assume
the provider's use of session text is itself acceptable to the data
owner. In this work we annotated only text that was already public.
Customizing on confidential data first requires an annotator permissible
under laws and agreements. Public annotated corpora are also not a
stand-in for such data: the clinical corpora we use either document
manual placeholder or pseudonym substitutions or show signs of them.
Their surface forms and formats therefore differ from unredacted
records.

Fine-tuning the XLM\nobreak\hbox{-}\nobreak{}R BIOES tagger locally is
cheap and fast: a full Ont3 fit takes about 2.3 hours on one GPU,
roughly \$5 of rented time (Section~\ref{sec-training-cost}). The cost
of adapting to new data is essentially the cost of annotating it,
together with that annotation's confidentiality exposure. A locally run
annotator on an isolated machine removes that exposure. The local models
we tested produced useful but clearly weaker supervision than the
frontier annotator (Appendix Table~\ref{tbl-local-annotation}; the
partial-supervision teacher in
Section~\ref{sec-teacher-omission-controls}), and we expect that gap to
narrow as local models improve.

\textbf{Transfer.} The reported scores describe the evaluated languages,
domains and annotation conventions. No representation should be made
that this or any model performs comparably on languages, domains,
formats or entity conventions it was not trained and evaluated on; our
own history includes a high development score followed by a large drop
on fresh legal text of a different kind
(Section~\ref{sec-ont2-transfer}). A deployment in a new setting needs
its own evaluation; where that evaluation falls short, annotating
in-domain text and refitting is inexpensive.

\textbf{Mandated de-identification.} Regulations fix which identifiers
must be removed, and often for whom: the HIPAA Safe Harbor standard
enumerates identifier categories of protected health information, the
GDPR distinguishes pseudonymised from anonymised data, and the same
record may need different treatment for a treating clinician, an insurer
or a researcher. Which mentions count can also depend on the role of the
person named, for example a patient's name as opposed to a clinician's,
a distinction the clinical corpora used here already annotate. Meeting
such a mandate with one model may require human review, or an operating
point that over-redacts until the text is no longer usable. Text that
can no longer be interpreted correctly is itself a high-stakes harm, for
example when redacted case records go to an outsourced claims processor.
Rather than a global penalty on the \textbf{O} label, which trades
precision on every type for recall on the mandated ones, we suggest
using the ontology-migration flow demonstrated here
(Section~\ref{sec-ontology-transition}) to introduce the mandated role
subtypes as explicit labels, then raising only those labels' scores at
decoding, so that recall increases where the mandate requires it and
nowhere else.

\section*{Acknowledgements}\label{acknowledgements}
\addcontentsline{toc}{section}{Acknowledgements}

We thank Steve DeNeefe for detailed comments on the manuscript.

We thank the creators and maintainers of Nemotron-PII, OpenPII, TAB,
idner-news-2k, Wojood, MEDDOCAN, SPY, MultiGraSCCo, MAPA, KLUE, HiNER,
AQMAR, OpenNER, FineWeb and FineWeb2, and the providers of HUDOC
judgments and PubMed Central articles. These resources supported
training or evaluation; Appendix Table~\ref{tbl-resource-access} lists
the sources and access conditions.

Generative AI assistants contributed to code development, figure
preparation, and revision throughout the manuscript and appendices. Code
contributions are tracked through \texttt{Contributing-model} entries in
the development history. The author directed, supervised, reviewed, and
revised the contributions, and retains final responsibility (models used
to produce annotations are described separately).

\clearpage

\label{acl-references-start}

\bibliography{references}

\clearpage
\appendix
\renewcommand{\thesection}{\Alph{section}}

\label{acl-appendices-start}

\section{Appendices}\label{sec-appendices}

\subsection{Gold coverage and mappings}\label{sec-human-gold-coverage}

The Gold-7 corpora annotate different subsets of the redaction task. We
therefore exclude predictions outside each corpus's annotated categories
before comparing untyped redaction regions. GLiNER2 (GL) receives 43
requested labels; O2, O3 and O4 predict the same 29 primary types. All
are evaluated against the same corpus-specific coverage, including
acceptable alternatives for coarse gold labels
(Table~\ref{tbl-human-gold-mappings}).

\ACLwidetabletrue\ACLcontinuedtablefalse

\ifXeTeX
\begin{longtable}[]{@{}
  >{\raggedright\arraybackslash}p{(\linewidth - 8\tabcolsep) * \real{0.2500}}
  >{\raggedleft\arraybackslash}p{(\linewidth - 8\tabcolsep) * \real{0.1000}}
  >{\raggedleft\arraybackslash}p{(\linewidth - 8\tabcolsep) * \real{0.2000}}
  >{\raggedleft\arraybackslash}p{(\linewidth - 8\tabcolsep) * \real{0.1500}}
  >{\raggedleft\arraybackslash}p{(\linewidth - 8\tabcolsep) * \real{0.3000}}@{}}
\caption{Annotation coverage of the frozen Gold-7 test subset. Counts
refer to original annotated spans, before merging redaction regions.
Source-label counts describe the evaluated schema, not every label in
the original corpus. }\label{tbl-human-gold-mappings}\tabularnewline
\toprule\noalign{}
\begin{minipage}[b]{\linewidth}\raggedright
Gold-7 component
\end{minipage} & \begin{minipage}[b]{\linewidth}\raggedleft
Inputs
\end{minipage} & \begin{minipage}[b]{\linewidth}\raggedleft
Source labels used by the mapping
\end{minipage} & \begin{minipage}[b]{\linewidth}\raggedleft
Accepted primary types
\end{minipage} & \begin{minipage}[b]{\linewidth}\raggedleft
Gold spans with multiple accepted types / annotated spans
\end{minipage} \\
\midrule\noalign{}
\endfirsthead
\toprule\noalign{}
\begin{minipage}[b]{\linewidth}\raggedright
Gold-7 component
\end{minipage} & \begin{minipage}[b]{\linewidth}\raggedleft
Inputs
\end{minipage} & \begin{minipage}[b]{\linewidth}\raggedleft
Source labels used by the mapping
\end{minipage} & \begin{minipage}[b]{\linewidth}\raggedleft
Accepted primary types
\end{minipage} & \begin{minipage}[b]{\linewidth}\raggedleft
Gold spans with multiple accepted types / annotated spans
\end{minipage} \\
\midrule\noalign{}
\endhead
\bottomrule\noalign{}
\endlastfoot
MAPA & 189 & 6 & 13 & 59 / 126 \\
AQMAR & 134 & 3 & 8 & 82 / 173 \\
OpenNER collection & 904 & 3 & 8 & 350 / 999 \\
Wojood subset & 56 & 11 & 14 & 52 / 99 \\
\end{longtable}
\else
\begin{longtable}[]{@{}
  >{\raggedright\arraybackslash}X
  >{\raggedleft\arraybackslash}X
  >{\raggedleft\arraybackslash}X
  >{\raggedleft\arraybackslash}X
  >{\raggedleft\arraybackslash}X@{}}
\caption{Annotation coverage of the frozen Gold-7 test subset. Counts
refer to original annotated spans, before merging redaction regions.
Source-label counts describe the evaluated schema, not every label in
the original corpus. }\label{tbl-human-gold-mappings}\tabularnewline
\toprule\noalign{}
\raggedright
Gold-7 component
 & \raggedleft
Inputs
 & \raggedleft
Source labels used by the mapping
 & \raggedleft
Accepted primary types
 & \raggedleft
Gold spans with multiple accepted types / annotated spans
 \\
\midrule\noalign{}
\endfirsthead
\toprule\noalign{}
\raggedright
Gold-7 component
 & \raggedleft
Inputs
 & \raggedleft
Source labels used by the mapping
 & \raggedleft
Accepted primary types
 & \raggedleft
Gold spans with multiple accepted types / annotated spans
 \\
\midrule\noalign{}
\endhead
\bottomrule\noalign{}
\endlastfoot
MAPA & 189 & 6 & 13 & 59 / 126 \\
AQMAR & 134 & 3 & 8 & 82 / 173 \\
OpenNER collection & 904 & 3 & 8 & 350 / 999 \\
Wojood subset & 56 & 11 & 14 & 52 / 99 \\
\end{longtable}
\fi

For example, a coarse location label accepts six location-related
primary types; MAPA dates accept either date or date of birth, and
amounts accept quantity or monetary amount. These alternatives preserve
the source's uncertainty and award no additional type-match credit in
this untyped comparison. Replacing GLiNER2's multi-target prediction
mappings with their single designated targets changes no scored counts
at any plotted threshold on Gold-7, so this mapping flexibility does not
explain the gap between GL and O2--O4. Coverage masking does matter: in
a corpus that annotates only people, organizations and locations, an
unmarked date is no evidence of a false positive. Mappings do not repair
omitted annotations, incorrect extents or disagreements about which
mentions should be annotated.

\begin{figure*}[tp]

\centering{

\ifXeTeX
\includegraphics[width=1\linewidth,height=6in,alt={Matrix of MEDDOCAN, MultiGraSCCo, TAB, MAPA and KLUE labels and their permissible target types.}]{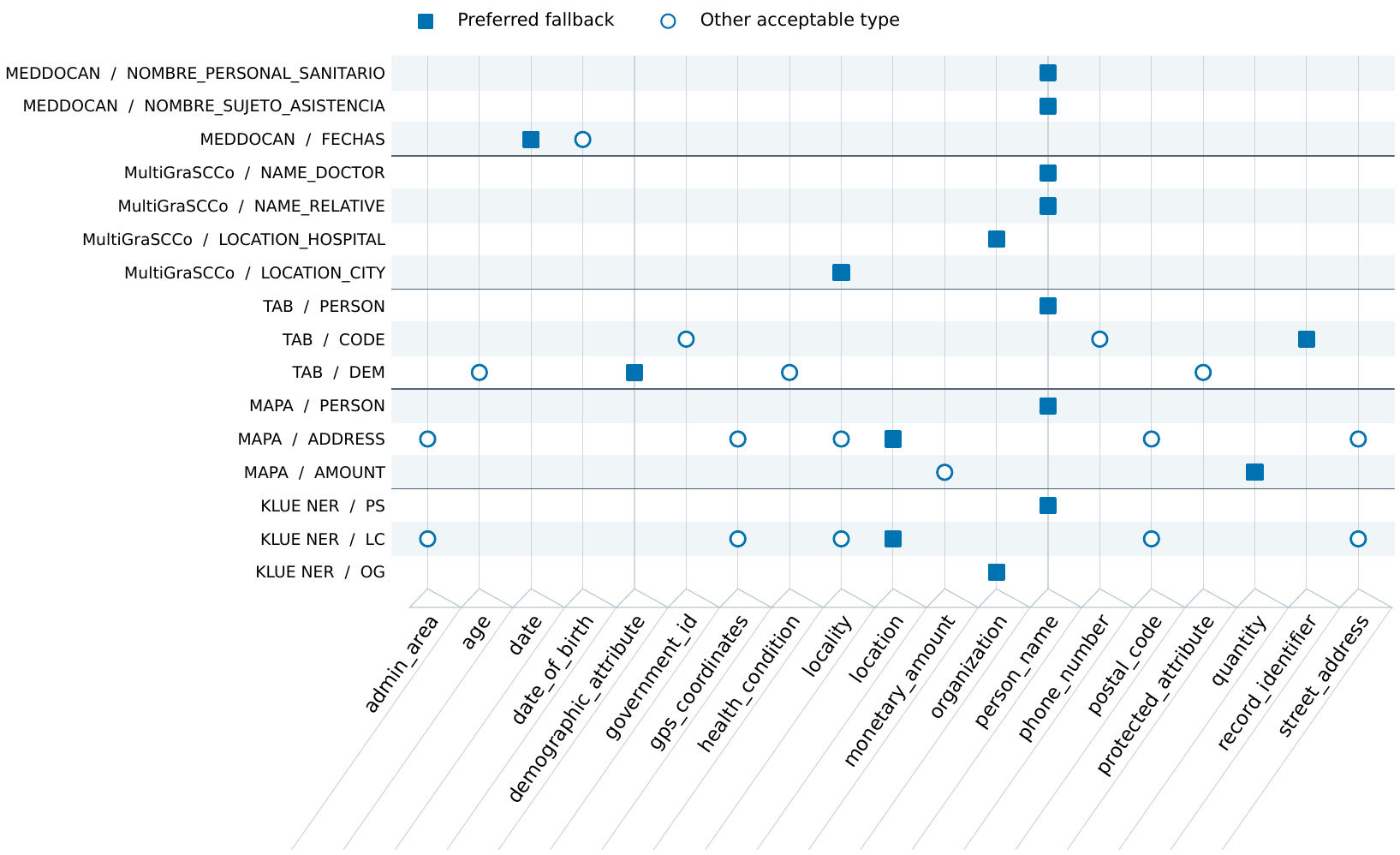}
\else
\includegraphics[alt={Matrix of MEDDOCAN, MultiGraSCCo, TAB, MAPA and KLUE labels and their permissible target types.}]{figures/ont2-gold-label-mappings-v1.pdf}
\fi

}

\caption{\label{fig-gold-mappings}Broad source labels can supervise
several target types: MAPA's ADDRESS permits locality, postal code and
street address, among others. Filled squares mark fallback labels; open
circles mark other acceptable labels. Shown: sixteen selected mappings
from five corpora to the Ont2 label inventory.}

\end{figure*}%

\subsection{Frontier annotation prompts}\label{sec-annotation-prompts}

Each annotation prompt pairs shared definitions of the 31 Ont3 entity
types with worked examples in the target language
(Figure~\ref{fig-annotation-prompts}). Given NFKC-normalized input, the
annotating model returns each distinct surface once, with one tag set
per occurrence of that surface in reading order (\texttt{{[}"O"{]}} for
an occurrence that is not an entity). The aligner locates the
occurrences, rejects any count mismatch and emits character spans for
non-\textbf{O} entries. Neither this copy-and-count offset design nor
the use of language-specific rather than shared demonstrations has been
isolated as a cause of annotation quality. Figure~\ref{fig-luna-result}
shows one Hebrew result from Luna
(GPT\nobreak\hbox{-}\nobreak{}5.6-Luna, the training-data annotator) in
which the same spelling is a person reference in one occurrence and an
ordinary adjective in the next.

\begin{figure*}[tp]

\centering{

\textbf{Input}

\begin{RTL}
\begin{otherlanguage}{hebrew}

בנוסף לכל זה מצאו הבלשים במקום בו לן החשוד חומר החשוד כסם מסוג מריחואנה
וציוד המשמש לגידול הסמים.

\end{otherlanguage}
\end{RTL}

\emph{English gloss:} The detectives also found, where \textbf{the
suspect} stayed, a substance \textbf{suspected} to be marijuana and
equipment for growing drugs.

\begin{center}\rule{0.5\linewidth}{0.5pt}\end{center}

\textbf{Luna output (raw)}

\begin{Shaded}
\begin{Highlighting}[]
\OtherTok{[}\FunctionTok{\{}\DataTypeTok{"t"}\FunctionTok{:}\StringTok{"הבלשים"}\FunctionTok{,}\DataTypeTok{"types"}\FunctionTok{:}\OtherTok{[[}\StringTok{"person\_reference"}\OtherTok{]]}\FunctionTok{\}}\OtherTok{,}
 \FunctionTok{\{}\DataTypeTok{"t"}\FunctionTok{:}\StringTok{"מקום"}\FunctionTok{,}\DataTypeTok{"types"}\FunctionTok{:}\OtherTok{[[}\StringTok{"location"}\OtherTok{]]}\FunctionTok{\}}\OtherTok{,}
 \FunctionTok{\{}\DataTypeTok{"t"}\FunctionTok{:}\StringTok{"החשוד"}\FunctionTok{,}\DataTypeTok{"types"}\FunctionTok{:}\OtherTok{[[}\StringTok{"person\_reference"}\OtherTok{],[}\StringTok{"O"}\OtherTok{]]}\FunctionTok{\}}\OtherTok{]}
\end{Highlighting}
\end{Shaded}

\textbf{Aligned spans (checked)}

\begin{verbatim}
[18,24) person_reference  הבלשים
[26,30) location          מקום
[37,42) person_reference  החשוד
[48,53) O, no span        החשוד
\end{verbatim}

}

\caption{\label{fig-luna-result}Luna's direct output, with line breaks
added, and the aligner's result. The same Hebrew spelling, החשוד, means
``the suspect'' in its first occurrence and ``suspected'' in its second;
its two tag sets label the person reference and leave the adjective
untagged (\textbf{O}). Both occurrences were found, the count matched,
and the aligner applied no surface repairs.}

\end{figure*}%

\subsection{Reproduction and release}\label{sec-release-boundary}

We release code, prompts and training recipes, including download and
preparation scripts and pinned dependencies, as release v1.5 of
\url{https://github.com/graehl/multilingual-pii-span-tagging}. We do not
release training data or model weights. Restricted sources can be
excluded; changing the mixture may change the results.

Once annotations are available, estimated training compute is \$9--18
for Ont2-equivalent exposure and \$4--5 for one Ont3 fit
(Section~\ref{sec-training-cost}). Obtaining and reviewing annotations
is a separate cost.

Table~\ref{tbl-resource-access} lists source terms checked on 22
September 2026. Links lead to distributions or access instructions.
Terms refer to the versions used here; component and source-text rights
still apply. The table includes evaluation and diagnostic sources as
well as training data.

\ACLwidetabletrue\ACLcontinuedtablefalse

\ifXeTeX
\begin{longtable}[]{@{}
  >{\raggedright\arraybackslash}p{(\linewidth - 2\tabcolsep) * \real{0.5000}}
  >{\raggedright\arraybackslash}p{(\linewidth - 2\tabcolsep) * \real{0.5000}}@{}}
\caption{Data sources and reuse conditions for the acquired versions.
}\label{tbl-resource-access}\tabularnewline
\toprule\noalign{}
\begin{minipage}[b]{\linewidth}\raggedright
Resource
\end{minipage} & \begin{minipage}[b]{\linewidth}\raggedright
Terms and access
\end{minipage} \\
\midrule\noalign{}
\endfirsthead
\toprule\noalign{}
\begin{minipage}[b]{\linewidth}\raggedright
Resource
\end{minipage} & \begin{minipage}[b]{\linewidth}\raggedright
Terms and access
\end{minipage} \\
\midrule\noalign{}
\endhead
\bottomrule\noalign{}
\endlastfoot
\href{https://huggingface.co/datasets/nvidia/Nemotron-PII}{Nemotron-PII}
\citep{nvidia2025-nemotron-pii},
\href{https://huggingface.co/datasets/ai4privacy/pii-masking-openpii-1m}{OpenPII
1M} and
\href{https://huggingface.co/datasets/ai4privacy/pii-masking-openpii-1.5m}{1.5M}
& CC BY 4.0; public downloads. \\
\href{https://huggingface.co/datasets/ai4privacy/pii-masking-200k/blob/691f4fc6b099563cecdc80c6d0444922229d23a4/LICENSE}{Ai4Privacy
200k} \citep{ai4privacy2023-pii-masking-200k} & Custom terms; for-profit
organizations with more than three staff need a corporate license.
Redistribution restrictions apply. \\
\href{https://huggingface.co/datasets/ai4privacy/pii-masking-health-phi-400k}{Ai4Privacy
health-PHI 400k} diagnostic sample & Gated access under custom
commercial terms. \\
\href{https://github.com/NorskRegnesentral/text-anonymization-benchmark}{TAB},
\href{https://github.com/khairunnisaor/idner-news-2k}{idner-news-2k}
\citep{khairunnisa2020-idner} and
\href{https://github.com/SinaLab/ArabicNER}{Wojood public sample} & MIT;
public repositories. Full Wojood is separate and was not used. \\
\href{https://huggingface.co/datasets/bigbio/meddocan}{MEDDOCAN}
\citep{marimon2019-meddocan},
\href{https://huggingface.co/datasets/mks-logic/SPY}{SPY}
\citep{savkin2025-spy},
\href{https://zenodo.org/records/19489040}{MultiGraSCCo}
\citep{baroud2026-multigrascco} and
\href{https://huggingface.co/datasets/joelniklaus/mapa}{MAPA EUR-Lex} &
CC BY 4.0; public distributions. Other MAPA packages may differ. \\
\href{https://huggingface.co/datasets/klue/klue}{KLUE NER}
\citep{park2021-klue} and
\href{https://github.com/cfiltnlp/HiNER}{HiNER} \citep{murthy2022-hiner}
& CC BY-SA 4.0; public distributions. \\
\href{https://huggingface.co/datasets/bltlab/open-ner-core-types}{AQMAR
through OpenNER} \citep{mohit2012-aqmar} & CC BY-SA 3.0 for the
annotations and the underlying Wikipedia text; citation requested. \\
\href{https://huggingface.co/datasets/bltlab/open-ner-core-types}{OpenNER
seven-corpus subset} & Collection: CC BY 4.0. AnCora and GermEval: CC BY
4.0; GSD and the four UNER components: CC BY-SA 4.0. Noncommercial (-NC)
components excluded. \\
\href{https://huggingface.co/datasets/HuggingFaceFW/fineweb}{FineWeb} /
\href{https://huggingface.co/datasets/HuggingFaceFW/fineweb-2}{FineWeb2}
& ODC-By 1.0; Common Crawl and underlying web-content terms also
apply. \\
\href{https://www.echr.coe.int/copyright-and-disclaimer}{HUDOC
judgments} & ECHR reproduction conditions and acknowledgment required;
translations may have separate terms. \\
\href{https://pmc.ncbi.nlm.nih.gov/tools/openftlist/}{PMC case reports}
& Article-specific licenses; use NCBI-approved retrieval services and
exclude incompatible articles. \\
Authored, translated and teacher-labeled records & Source-text and
model-service terms apply. \\
\end{longtable}
\else
\begin{longtable}[]{@{}
  >{\raggedright\arraybackslash}X
  >{\raggedright\arraybackslash}X@{}}
\caption{Data sources and reuse conditions for the acquired versions.
}\label{tbl-resource-access}\tabularnewline
\toprule\noalign{}
\raggedright
Resource
 & \raggedright
Terms and access
 \\
\midrule\noalign{}
\endfirsthead
\toprule\noalign{}
\raggedright
Resource
 & \raggedright
Terms and access
 \\
\midrule\noalign{}
\endhead
\bottomrule\noalign{}
\endlastfoot
\href{https://huggingface.co/datasets/nvidia/Nemotron-PII}{Nemotron-PII}
\citep{nvidia2025-nemotron-pii},
\href{https://huggingface.co/datasets/ai4privacy/pii-masking-openpii-1m}{OpenPII
1M} and
\href{https://huggingface.co/datasets/ai4privacy/pii-masking-openpii-1.5m}{1.5M}
& CC BY 4.0; public downloads. \\
\href{https://huggingface.co/datasets/ai4privacy/pii-masking-200k/blob/691f4fc6b099563cecdc80c6d0444922229d23a4/LICENSE}{Ai4Privacy
200k} \citep{ai4privacy2023-pii-masking-200k} & Custom terms; for-profit
organizations with more than three staff need a corporate license.
Redistribution restrictions apply. \\
\href{https://huggingface.co/datasets/ai4privacy/pii-masking-health-phi-400k}{Ai4Privacy
health-PHI 400k} diagnostic sample & Gated access under custom
commercial terms. \\
\href{https://github.com/NorskRegnesentral/text-anonymization-benchmark}{TAB},
\href{https://github.com/khairunnisaor/idner-news-2k}{idner-news-2k}
\citep{khairunnisa2020-idner} and
\href{https://github.com/SinaLab/ArabicNER}{Wojood public sample} & MIT;
public repositories. Full Wojood is separate and was not used. \\
\href{https://huggingface.co/datasets/bigbio/meddocan}{MEDDOCAN}
\citep{marimon2019-meddocan},
\href{https://huggingface.co/datasets/mks-logic/SPY}{SPY}
\citep{savkin2025-spy},
\href{https://zenodo.org/records/19489040}{MultiGraSCCo}
\citep{baroud2026-multigrascco} and
\href{https://huggingface.co/datasets/joelniklaus/mapa}{MAPA EUR-Lex} &
CC BY 4.0; public distributions. Other MAPA packages may differ. \\
\href{https://huggingface.co/datasets/klue/klue}{KLUE NER}
\citep{park2021-klue} and
\href{https://github.com/cfiltnlp/HiNER}{HiNER} \citep{murthy2022-hiner}
& CC BY-SA 4.0; public distributions. \\
\href{https://huggingface.co/datasets/bltlab/open-ner-core-types}{AQMAR
through OpenNER} \citep{mohit2012-aqmar} & CC BY-SA 3.0 for the
annotations and the underlying Wikipedia text; citation requested. \\
\href{https://huggingface.co/datasets/bltlab/open-ner-core-types}{OpenNER
seven-corpus subset} & Collection: CC BY 4.0. AnCora and GermEval: CC BY
4.0; GSD and the four UNER components: CC BY-SA 4.0. Noncommercial (-NC)
components excluded. \\
\href{https://huggingface.co/datasets/HuggingFaceFW/fineweb}{FineWeb} /
\href{https://huggingface.co/datasets/HuggingFaceFW/fineweb-2}{FineWeb2}
& ODC-By 1.0; Common Crawl and underlying web-content terms also
apply. \\
\href{https://www.echr.coe.int/copyright-and-disclaimer}{HUDOC
judgments} & ECHR reproduction conditions and acknowledgment required;
translations may have separate terms. \\
\href{https://pmc.ncbi.nlm.nih.gov/tools/openftlist/}{PMC case reports}
& Article-specific licenses; use NCBI-approved retrieval services and
exclude incompatible articles. \\
Authored, translated and teacher-labeled records & Source-text and
model-service terms apply. \\
\end{longtable}
\fi

In the 1,000-row Ai4Privacy health-PHI diagnostic sample, we found
inconsistent substituted values and untranslated English text under
other language codes. These findings concern that sample, not all
Ai4Privacy releases. Separately, Final20 briefly contained 21 Ai4Privacy
200k rows; excluding them left 6,979 records
(Section~\ref{sec-evaluation-sets}).

Before admitting additional data, we screen for overlap using lexical
matching and embedding similarity.

\subsubsection{Training-only cost}\label{sec-training-cost}

Table~\ref{tbl-training-cost} estimates instance rental once annotations
are available. Both fits use 64 sampled windows per update, including
repeated draws. We group sampled examples of similar token length into
physical batches to reduce padding, while retaining the requested
sampling mixture. Thus these costs reflect length-matched batching, not
arbitrary batches padded to the longest example.

\ACLwidetabletrue\ACLcontinuedtablefalse

\ifXeTeX
\begin{longtable}[]{@{}
  >{\raggedright\arraybackslash}p{(\linewidth - 8\tabcolsep) * \real{0.2000}}
  >{\raggedleft\arraybackslash}p{(\linewidth - 8\tabcolsep) * \real{0.2000}}
  >{\raggedright\arraybackslash}p{(\linewidth - 8\tabcolsep) * \real{0.2000}}
  >{\raggedleft\arraybackslash}p{(\linewidth - 8\tabcolsep) * \real{0.2000}}
  >{\raggedleft\arraybackslash}p{(\linewidth - 8\tabcolsep) * \real{0.2000}}@{}}
\caption{Training-only estimates at AWS Linux Spot prices observed on 22
September 2026: \$1.41--3.00/hour in us-west-2 for Ont2 and
\$1.87--2.37/hour in ap-south-1 for Ont3.
}\label{tbl-training-cost}\tabularnewline
\toprule\noalign{}
\begin{minipage}[b]{\linewidth}\raggedright
Training scope
\end{minipage} & \begin{minipage}[b]{\linewidth}\raggedleft
Updates
\end{minipage} & \begin{minipage}[b]{\linewidth}\raggedright
GPU / AWS instance
\end{minipage} & \begin{minipage}[b]{\linewidth}\raggedleft
Hours
\end{minipage} & \begin{minipage}[b]{\linewidth}\raggedleft
Estimated \$
\end{minipage} \\
\midrule\noalign{}
\endfirsthead
\toprule\noalign{}
\begin{minipage}[b]{\linewidth}\raggedright
Training scope
\end{minipage} & \begin{minipage}[b]{\linewidth}\raggedleft
Updates
\end{minipage} & \begin{minipage}[b]{\linewidth}\raggedright
GPU / AWS instance
\end{minipage} & \begin{minipage}[b]{\linewidth}\raggedleft
Hours
\end{minipage} & \begin{minipage}[b]{\linewidth}\raggedleft
Estimated \$
\end{minipage} \\
\midrule\noalign{}
\endhead
\bottomrule\noalign{}
\endlastfoot
Ont2-equivalent exposure & 33,300 & L40S / g6e.4xlarge & 6.0* &
8.5--18.0* \\
Ont3 fit from pretrained XLM\nobreak\hbox{-}\nobreak{}R & 16,000 & RTX
PRO 6000 Blackwell Server / g7e.2xlarge & 2.3 & 4.3--5.4 \\
\end{longtable}
\else
\begin{longtable}[]{@{}
  >{\raggedright\arraybackslash}X
  >{\raggedleft\arraybackslash}X
  >{\raggedright\arraybackslash}X
  >{\raggedleft\arraybackslash}X
  >{\raggedleft\arraybackslash}X@{}}
\caption{Training-only estimates at AWS Linux Spot prices observed on 22
September 2026: \$1.41--3.00/hour in us-west-2 for Ont2 and
\$1.87--2.37/hour in ap-south-1 for Ont3.
}\label{tbl-training-cost}\tabularnewline
\toprule\noalign{}
\raggedright
Training scope
 & \raggedleft
Updates
 & \raggedright
GPU / AWS instance
 & \raggedleft
Hours
 & \raggedleft
Estimated \$
 \\
\midrule\noalign{}
\endfirsthead
\toprule\noalign{}
\raggedright
Training scope
 & \raggedleft
Updates
 & \raggedright
GPU / AWS instance
 & \raggedleft
Hours
 & \raggedleft
Estimated \$
 \\
\midrule\noalign{}
\endhead
\bottomrule\noalign{}
\endlastfoot
Ont2-equivalent exposure & 33,300 & L40S / g6e.4xlarge & 6.0* &
8.5--18.0* \\
Ont3 fit from pretrained XLM\nobreak\hbox{-}\nobreak{}R & 16,000 & RTX
PRO 6000 Blackwell Server / g7e.2xlarge & 2.3 & 4.3--5.4 \\
\end{longtable}
\fi

* Ont2 extrapolates an 8,000-update, 5,194-second fit to the 33,300
retained updates across its training stages. Ont3 uses the measured full
16,000-update run on 47,641 windows with previous and next context.
These estimates exclude annotation, data preparation, experimental
search, auxiliary-model training and non-training infrastructure costs.

\subsection{Prompted local LLM annotators}\label{sec-local-annotators}

We evaluated prompted Gemma and Qwen models as locally runnable
alternatives to the frontier annotator.

On Silver-dev, 659 segments across 35 languages, we checked each model's
final answer for valid annotation structure and literal source alignment
before scoring. Failures route the entire segment to O4; valid empty
annotations remain unchanged. Character scores pool occupied character
positions across segments using the Ont3 annotations and the
optional-reference policy (Section~\ref{sec-span-views}). Prompt
versions and retry budgets differ: Gemma-12B allowed four retries,
Gemma-31B one, and Qwen none. All requests disabled thinking.

\ACLwidetabletrue\ACLcontinuedtablefalse

\ifXeTeX
\begin{longtable}[]{@{}
  >{\raggedright\arraybackslash}p{(\linewidth - 8\tabcolsep) * \real{0.3400}}
  >{\raggedleft\arraybackslash}p{(\linewidth - 8\tabcolsep) * \real{0.1600}}
  >{\raggedleft\arraybackslash}p{(\linewidth - 8\tabcolsep) * \real{0.1600}}
  >{\raggedleft\arraybackslash}p{(\linewidth - 8\tabcolsep) * \real{0.1700}}
  >{\raggedleft\arraybackslash}p{(\linewidth - 8\tabcolsep) * \real{0.1700}}@{}}
\caption{Annotation quality on the 659 Silver-dev segments. Asterisks
mark XLM\nobreak\hbox{-}\nobreak{}R substitution for invalid
annotations; without fallback, invalid outputs count as empty
predictions. Rescued subsets contain 33, 41 and 54 segments,
respectively. O4 uses zero \textbf{O} bias, isolated input and boundary
refinement. }\label{tbl-local-annotation}\tabularnewline
\toprule\noalign{}
\begin{minipage}[b]{\linewidth}\raggedright
Model
\end{minipage} & \begin{minipage}[b]{\linewidth}\raggedleft
Failed format (\%)
\end{minipage} & \begin{minipage}[b]{\linewidth}\raggedleft
Character F1
\end{minipage} & \begin{minipage}[b]{\linewidth}\raggedleft
With XLM\nobreak\hbox{-}\nobreak{}R fallback*
\end{minipage} & \begin{minipage}[b]{\linewidth}\raggedleft
Rescued subset F1*
\end{minipage} \\
\midrule\noalign{}
\endfirsthead
\toprule\noalign{}
\begin{minipage}[b]{\linewidth}\raggedright
Model
\end{minipage} & \begin{minipage}[b]{\linewidth}\raggedleft
Failed format (\%)
\end{minipage} & \begin{minipage}[b]{\linewidth}\raggedleft
Character F1
\end{minipage} & \begin{minipage}[b]{\linewidth}\raggedleft
With XLM\nobreak\hbox{-}\nobreak{}R fallback*
\end{minipage} & \begin{minipage}[b]{\linewidth}\raggedleft
Rescued subset F1*
\end{minipage} \\
\midrule\noalign{}
\endhead
\bottomrule\noalign{}
\endlastfoot
Gemma\nobreak\hbox{-}\nobreak{}4-12B & \textbf{5.01} & 83.84 & 86.38* &
84.51* \\
Gemma\nobreak\hbox{-}\nobreak{}4-31B-FP8 & \textbf{6.22} & 83.98 &
90.11* & 85.13* \\
Qwen3.8-Flash\nobreak\hbox{-}\nobreak{}Next-NVFP4 & \textbf{8.19} &
76.03 & 87.02* & 90.77* \\
O4, sentence only & --- & 91.53 & --- & --- \\
\end{longtable}
\else
\begin{longtable}[]{@{}
  >{\raggedright\arraybackslash}X
  >{\raggedleft\arraybackslash}X
  >{\raggedleft\arraybackslash}X
  >{\raggedleft\arraybackslash}X
  >{\raggedleft\arraybackslash}X@{}}
\caption{Annotation quality on the 659 Silver-dev segments. Asterisks
mark XLM\nobreak\hbox{-}\nobreak{}R substitution for invalid
annotations; without fallback, invalid outputs count as empty
predictions. Rescued subsets contain 33, 41 and 54 segments,
respectively. O4 uses zero \textbf{O} bias, isolated input and boundary
refinement. }\label{tbl-local-annotation}\tabularnewline
\toprule\noalign{}
\raggedright
Model
 & \raggedleft
Failed format (\%)
 & \raggedleft
Character F1
 & \raggedleft
With XLM\nobreak\hbox{-}\nobreak{}R fallback*
 & \raggedleft
Rescued subset F1*
 \\
\midrule\noalign{}
\endfirsthead
\toprule\noalign{}
\raggedright
Model
 & \raggedleft
Failed format (\%)
 & \raggedleft
Character F1
 & \raggedleft
With XLM\nobreak\hbox{-}\nobreak{}R fallback*
 & \raggedleft
Rescued subset F1*
 \\
\midrule\noalign{}
\endhead
\bottomrule\noalign{}
\endlastfoot
Gemma\nobreak\hbox{-}\nobreak{}4-12B & \textbf{5.01} & 83.84 & 86.38* &
84.51* \\
Gemma\nobreak\hbox{-}\nobreak{}4-31B-FP8 & \textbf{6.22} & 83.98 &
90.11* & 85.13* \\
Qwen3.8-Flash\nobreak\hbox{-}\nobreak{}Next-NVFP4 & \textbf{8.19} &
76.03 & 87.02* & 90.77* \\
O4, sentence only & --- & 91.53 & --- & --- \\
\end{longtable}
\fi

Failed format includes invalid labels, nonliteral spans, ordering and
forbidden overlap. Routing uses only the response and input text.

The Gemma prompts include surrounding paragraph context; Qwen receives
the preceding and following source sentences. The matched
XLM\nobreak\hbox{-}\nobreak{}R context comparisons appear in
Section~\ref{sec-neighboring-context}.

\ACLwidetabletrue\ACLcontinuedtablefalse

\ifXeTeX
\begin{longtable}[]{@{}
  >{\raggedright\arraybackslash}p{(\linewidth - 6\tabcolsep) * \real{0.4800}}
  >{\raggedleft\arraybackslash}p{(\linewidth - 6\tabcolsep) * \real{0.1400}}
  >{\raggedleft\arraybackslash}p{(\linewidth - 6\tabcolsep) * \real{0.1400}}
  >{\raggedleft\arraybackslash}p{(\linewidth - 6\tabcolsep) * \real{0.2400}}@{}}
\caption{Gemma\nobreak\hbox{-}\nobreak{}4-31B-IT serial generation
measured on a 96-GB RTX PRO 6000: 35 inputs in 35 languages, repeated
twice, 591 labeled input tokens per pass. Other prompted rates divide
labeled input tokens by summed request durations, including retries. The
encoder rate is measured throughput (Table~\ref{tbl-serving-cost}).
Parameter counts describe nominal backbones; the active count is shown
for Flash-Next's sparse main model, per token
(Table~\ref{tbl-flash-readouts}).
}\label{tbl-local-annotation-speed}\tabularnewline
\toprule\noalign{}
\begin{minipage}[b]{\linewidth}\raggedright
System
\end{minipage} & \begin{minipage}[b]{\linewidth}\raggedleft
Params (B)
\end{minipage} & \begin{minipage}[b]{\linewidth}\raggedleft
Active (B)
\end{minipage} & \begin{minipage}[b]{\linewidth}\raggedleft
Labeled input tokens/s
\end{minipage} \\
\midrule\noalign{}
\endfirsthead
\toprule\noalign{}
\begin{minipage}[b]{\linewidth}\raggedright
System
\end{minipage} & \begin{minipage}[b]{\linewidth}\raggedleft
Params (B)
\end{minipage} & \begin{minipage}[b]{\linewidth}\raggedleft
Active (B)
\end{minipage} & \begin{minipage}[b]{\linewidth}\raggedleft
Labeled input tokens/s
\end{minipage} \\
\midrule\noalign{}
\endhead
\bottomrule\noalign{}
\endlastfoot
Gemma\nobreak\hbox{-}\nobreak{}4-12B & 12 & --- & 2 \\
Gemma\nobreak\hbox{-}\nobreak{}4-31B-IT, bfloat16 & 30.7 & --- & 2.5 \\
Qwen3.8-Flash\nobreak\hbox{-}\nobreak{}Next-NVFP4 & 176 & 6 & 15 \\
XLM\nobreak\hbox{-}\nobreak{}R encoder & 0.55 & --- & 2,744 \\
\end{longtable}
\else
\begin{longtable}[]{@{}
  >{\raggedright\arraybackslash}X
  >{\raggedleft\arraybackslash}X
  >{\raggedleft\arraybackslash}X
  >{\raggedleft\arraybackslash}X@{}}
\caption{Gemma\nobreak\hbox{-}\nobreak{}4-31B-IT serial generation
measured on a 96-GB RTX PRO 6000: 35 inputs in 35 languages, repeated
twice, 591 labeled input tokens per pass. Other prompted rates divide
labeled input tokens by summed request durations, including retries. The
encoder rate is measured throughput (Table~\ref{tbl-serving-cost}).
Parameter counts describe nominal backbones; the active count is shown
for Flash-Next's sparse main model, per token
(Table~\ref{tbl-flash-readouts}).
}\label{tbl-local-annotation-speed}\tabularnewline
\toprule\noalign{}
\raggedright
System
 & \raggedleft
Params (B)
 & \raggedleft
Active (B)
 & \raggedleft
Labeled input tokens/s
 \\
\midrule\noalign{}
\endfirsthead
\toprule\noalign{}
\raggedright
System
 & \raggedleft
Params (B)
 & \raggedleft
Active (B)
 & \raggedleft
Labeled input tokens/s
 \\
\midrule\noalign{}
\endhead
\bottomrule\noalign{}
\endlastfoot
Gemma\nobreak\hbox{-}\nobreak{}4-12B & 12 & --- & 2 \\
Gemma\nobreak\hbox{-}\nobreak{}4-31B-IT, bfloat16 & 30.7 & --- & 2.5 \\
Qwen3.8-Flash\nobreak\hbox{-}\nobreak{}Next-NVFP4 & 176 & 6 & 15 \\
XLM\nobreak\hbox{-}\nobreak{}R encoder & 0.55 & --- & 2,744 \\
\end{longtable}
\fi

The serial Gemma check replays the saved annotation prompts through
Transformers in bfloat16. Each pass took 241 seconds, including full
prompt prefill and output generation, excluding model loading,
tokenization and validation; requests reuse only their own KV state.
This timing run uses no retries and does not replace the FP8 quality
evaluation above.

Adding estimated encoder time for the failed segments (their token
counts divided by the encoder's measured 2,744 tokens/s) leaves these
rounded rates unchanged for the original evaluated configurations.

\subsection{Frozen language-model token
heads}\label{sec-frozen-llm-heads}

These experiments cached hidden states and trained a BIOES
classification head; they did not generate annotation text. On 1,113
reused development sentences, rescored on the shared revised
annotations, scores use language-weighted typed character F1 and
optional references. In the matched Gemma\nobreak\hbox{-}\nobreak{}4-12B
comparison, adding the task instructions before each sentence raised
affine-head F1 from 26.9 to 59.0, using the same 10,427 training rows.
These were ordinary text instructions, distinct from the learned
soft-prompt vectors in Section~\ref{sec-prompt-controls}.

The best result in that comparison, 75.8, used
Gemma\nobreak\hbox{-}\nobreak{}4-31B layer-40 states, a 3,680-token
instruction prefix, no preceding sentence, and a nonlinear head with 512
hidden units. The fully fine-tuned XLM\nobreak\hbox{-}\nobreak{}R-large
control averaged 83.2 over five seeds (range 82.9--83.8, standard
deviation 0.38).

On the Ont3 annotation mixture, task instructions also improve frozen
Qwen3.8-Flash\nobreak\hbox{-}\nobreak{}Next representations. A nonlinear
head and neighboring token states provide further gains
(Table~\ref{tbl-flash-readouts}). Preceding and following sentences are
placed before the annotation target sentence in the causal model's
input; the head predicts only target-token labels. Adding neighboring
states to the nonlinear head improves typed span F1 by 5.7 points
(paired document-bootstrap 95\% interval {[}3.53, 7.94{]}). For
XLM\nobreak\hbox{-}\nobreak{}R with an affine head, neighboring states
change typed span F1 by −0.17 points.

\ACLwidetabletrue\ACLcontinuedtablefalse

\ifXeTeX
\begin{longtable}[]{@{}
  >{\raggedright\arraybackslash}p{(\linewidth - 10\tabcolsep) * \real{0.3200}}
  >{\raggedleft\arraybackslash}p{(\linewidth - 10\tabcolsep) * \real{0.0800}}
  >{\raggedright\arraybackslash}p{(\linewidth - 10\tabcolsep) * \real{0.2100}}
  >{\raggedright\arraybackslash}p{(\linewidth - 10\tabcolsep) * \real{0.1300}}
  >{\raggedleft\arraybackslash}p{(\linewidth - 10\tabcolsep) * \real{0.1200}}
  >{\raggedleft\arraybackslash}p{(\linewidth - 10\tabcolsep) * \real{0.1400}}@{}}
\caption{Flash\nobreak\hbox{-}\nobreak{}Next, Gemma and
XLM\nobreak\hbox{-}\nobreak{}R heads on the 659 Silver-dev segments from
468 source documents. Scores pool counts with optional references and
zero \textbf{O} bias. Offsets are relative to the target token; 0 uses
only its own state. LLM rows use the annotation prompt unless marked
unprompted; XLM\nobreak\hbox{-}\nobreak{}R uses no prompt. Parameter
counts are nominal backbone sizes in billions, excluding the BIOES head.
Flash\nobreak\hbox{-}\nobreak{}Next has 125B main-model parameters (6B
active per token) plus 51B n-gram embeddings; its unused 4B multi-token
prediction module is excluded. Flash and Gemma heads use the same 32,821
training targets, 279-target selector and annotation prompt; the fit
ceiling is ten epochs of 250 batches of 4,096 tokens. LLM heads train in
about 15 minutes or less on frozen, already-computed encodings. The
first three XLM\nobreak\hbox{-}\nobreak{}R heads receive 1,000 steps on
40,303 training windows with the encoder frozen and the same selector.
O4 is the fully fine-tuned serving model.
}\label{tbl-flash-readouts}\tabularnewline
\toprule\noalign{}
\begin{minipage}[b]{\linewidth}\raggedright
(Prompted) LLM / encoder
\end{minipage} & \begin{minipage}[b]{\linewidth}\raggedleft
Params (B)
\end{minipage} & \begin{minipage}[b]{\linewidth}\raggedright
Head
\end{minipage} & \begin{minipage}[b]{\linewidth}\raggedright
Input state offsets
\end{minipage} & \begin{minipage}[b]{\linewidth}\raggedleft
Typed span F1 (\%)
\end{minipage} & \begin{minipage}[b]{\linewidth}\raggedleft
Character F1 (\%)
\end{minipage} \\
\midrule\noalign{}
\endfirsthead
\toprule\noalign{}
\begin{minipage}[b]{\linewidth}\raggedright
(Prompted) LLM / encoder
\end{minipage} & \begin{minipage}[b]{\linewidth}\raggedleft
Params (B)
\end{minipage} & \begin{minipage}[b]{\linewidth}\raggedright
Head
\end{minipage} & \begin{minipage}[b]{\linewidth}\raggedright
Input state offsets
\end{minipage} & \begin{minipage}[b]{\linewidth}\raggedleft
Typed span F1 (\%)
\end{minipage} & \begin{minipage}[b]{\linewidth}\raggedleft
Character F1 (\%)
\end{minipage} \\
\midrule\noalign{}
\endhead
\bottomrule\noalign{}
\endlastfoot
Flash-Next unprompted & 176 & Affine & 0 & 39.14 & 78.59 \\
Qwen3.8-Flash-Next & 176 & Affine & 0 & 44.56 & 80.64 \\
Qwen3.8-Flash-Next & 176 & Affine & −1, 0, +1 & 53.00 & 82.96 \\
Qwen3.8-Flash-Next & 176 & 512-unit GELU & 0 & 53.80 & 85.25 \\
Qwen3.8-Flash-Next & 176 & 512-unit GELU & −1, 0, +1 & 59.54 & 86.61 \\
Gemma\nobreak\hbox{-}\nobreak{}4-31B base & 30.7 & 512-unit GELU & −1,
0, +1 & 62.41 & 87.07 \\
Gemma\nobreak\hbox{-}\nobreak{}4-31B IT & 30.7 & 512-unit GELU & −1, 0,
+1 & 61.12 & 86.16 \\
XLM\nobreak\hbox{-}\nobreak{}R & 0.55 & Affine & 0 & 72.36 & 91.47 \\
XLM\nobreak\hbox{-}\nobreak{}R & 0.55 & 512-unit GELU & 0 & 72.44 &
91.48 \\
XLM\nobreak\hbox{-}\nobreak{}R & 0.55 & Affine & −1, 0, +1 & 72.19 &
91.40 \\
XLM\nobreak\hbox{-}\nobreak{}R O4 & 0.55 & Affine & 0 & 76.90 & 91.53 \\
\end{longtable}
\else
\begin{longtable}[]{@{}
  >{\raggedright\arraybackslash}X
  >{\raggedleft\arraybackslash}X
  >{\raggedright\arraybackslash}X
  >{\raggedright\arraybackslash}X
  >{\raggedleft\arraybackslash}X
  >{\raggedleft\arraybackslash}X@{}}
\caption{Flash\nobreak\hbox{-}\nobreak{}Next, Gemma and
XLM\nobreak\hbox{-}\nobreak{}R heads on the 659 Silver-dev segments from
468 source documents. Scores pool counts with optional references and
zero \textbf{O} bias. Offsets are relative to the target token; 0 uses
only its own state. LLM rows use the annotation prompt unless marked
unprompted; XLM\nobreak\hbox{-}\nobreak{}R uses no prompt. Parameter
counts are nominal backbone sizes in billions, excluding the BIOES head.
Flash\nobreak\hbox{-}\nobreak{}Next has 125B main-model parameters (6B
active per token) plus 51B n-gram embeddings; its unused 4B multi-token
prediction module is excluded. Flash and Gemma heads use the same 32,821
training targets, 279-target selector and annotation prompt; the fit
ceiling is ten epochs of 250 batches of 4,096 tokens. LLM heads train in
about 15 minutes or less on frozen, already-computed encodings. The
first three XLM\nobreak\hbox{-}\nobreak{}R heads receive 1,000 steps on
40,303 training windows with the encoder frozen and the same selector.
O4 is the fully fine-tuned serving model.
}\label{tbl-flash-readouts}\tabularnewline
\toprule\noalign{}
\raggedright
(Prompted) LLM / encoder
 & \raggedleft
Params (B)
 & \raggedright
Head
 & \raggedright
Input state offsets
 & \raggedleft
Typed span F1 (\%)
 & \raggedleft
Character F1 (\%)
 \\
\midrule\noalign{}
\endfirsthead
\toprule\noalign{}
\raggedright
(Prompted) LLM / encoder
 & \raggedleft
Params (B)
 & \raggedright
Head
 & \raggedright
Input state offsets
 & \raggedleft
Typed span F1 (\%)
 & \raggedleft
Character F1 (\%)
 \\
\midrule\noalign{}
\endhead
\bottomrule\noalign{}
\endlastfoot
Flash-Next unprompted & 176 & Affine & 0 & 39.14 & 78.59 \\
Qwen3.8-Flash-Next & 176 & Affine & 0 & 44.56 & 80.64 \\
Qwen3.8-Flash-Next & 176 & Affine & −1, 0, +1 & 53.00 & 82.96 \\
Qwen3.8-Flash-Next & 176 & 512-unit GELU & 0 & 53.80 & 85.25 \\
Qwen3.8-Flash-Next & 176 & 512-unit GELU & −1, 0, +1 & 59.54 & 86.61 \\
Gemma\nobreak\hbox{-}\nobreak{}4-31B base & 30.7 & 512-unit GELU & −1,
0, +1 & 62.41 & 87.07 \\
Gemma\nobreak\hbox{-}\nobreak{}4-31B IT & 30.7 & 512-unit GELU & −1, 0,
+1 & 61.12 & 86.16 \\
XLM\nobreak\hbox{-}\nobreak{}R & 0.55 & Affine & 0 & 72.36 & 91.47 \\
XLM\nobreak\hbox{-}\nobreak{}R & 0.55 & 512-unit GELU & 0 & 72.44 &
91.48 \\
XLM\nobreak\hbox{-}\nobreak{}R & 0.55 & Affine & −1, 0, +1 & 72.19 &
91.40 \\
XLM\nobreak\hbox{-}\nobreak{}R O4 & 0.55 & Affine & 0 & 76.90 & 91.53 \\
\end{longtable}
\fi

Under the matched prompt and head recipe, Gemma base improves exact
typed-span F1 over Flash\nobreak\hbox{-}\nobreak{}Next by 2.86 points
(95\% interval {[}0.41, 5.32{]}; character redaction +0.46 {[}−.72,
1.68{]}). Instruction tuning, thought to improve performance of prompted
tasks, did not: the IT head changes typed F1 by −1.29 points {[}−3.67,
+1.10{]} and character F1 by −0.92 {[}−2.10, +0.31{]}. The frozen LLM
heads use feature standardization followed by two affine layers with
GELU, without LayerNorm or dropout (adding LayerNorm and dropout 0.1
gave no resolved improvement: typed F1 changed by −0.23 points for
Flash\nobreak\hbox{-}\nobreak{}Next and −0.30 for Gemma base).

On a 96-GB RTX PRO 6000 Blackwell, Flash\nobreak\hbox{-}\nobreak{}Next
representation capture for these 659 segments took 455 seconds: 29
labeled input tokens/s over 13,193 tokens. Requests ran one at a time
with instructions and neighboring context; the elapsed time includes
preparation, representation transfer and cache writes, and excludes
server model loading and head training. This is about 95 times slower
than the XLM\nobreak\hbox{-}\nobreak{}R reference of 2,744 tokens/s
(Table~\ref{tbl-serving-cost}), with different hardware and batching.

Gemma\nobreak\hbox{-}\nobreak{}4-31B base and IT each took 142 seconds
to encode the same 659 targets at 93 labeled tokens/s on this GPU. Both
use the same task instructions and neighboring context, bfloat16
inference and cached instruction prefixes. Timing includes prefix
prefill, forward passes, representation transfer and cache writes,
excluding model loading. Gemma uses Transformers while Flash uses vLLM.

Using \emph{only} the previous or next token state scores 30.9\% and
22.7\% exact typed span F1 with an affine head, below the current-state
control. Together with an audit of token alignment, this favors
combining complementary states rather than correcting a uniform
one-token offset.

These comparisons reuse development data and condition on selected heads
and one representation cache.

\begin{figure*}[tp]

\centering{

\ifXeTeX
\includegraphics[width=1\linewidth,height=6in,alt={Two precision--recall panels compare exact typed spans and redacted characters. O4 has the highest observed F1 in both panels. Four frozen LLM heads and three prompted annotators are shown. Marker shapes and line styles distinguish systems in grayscale.}]{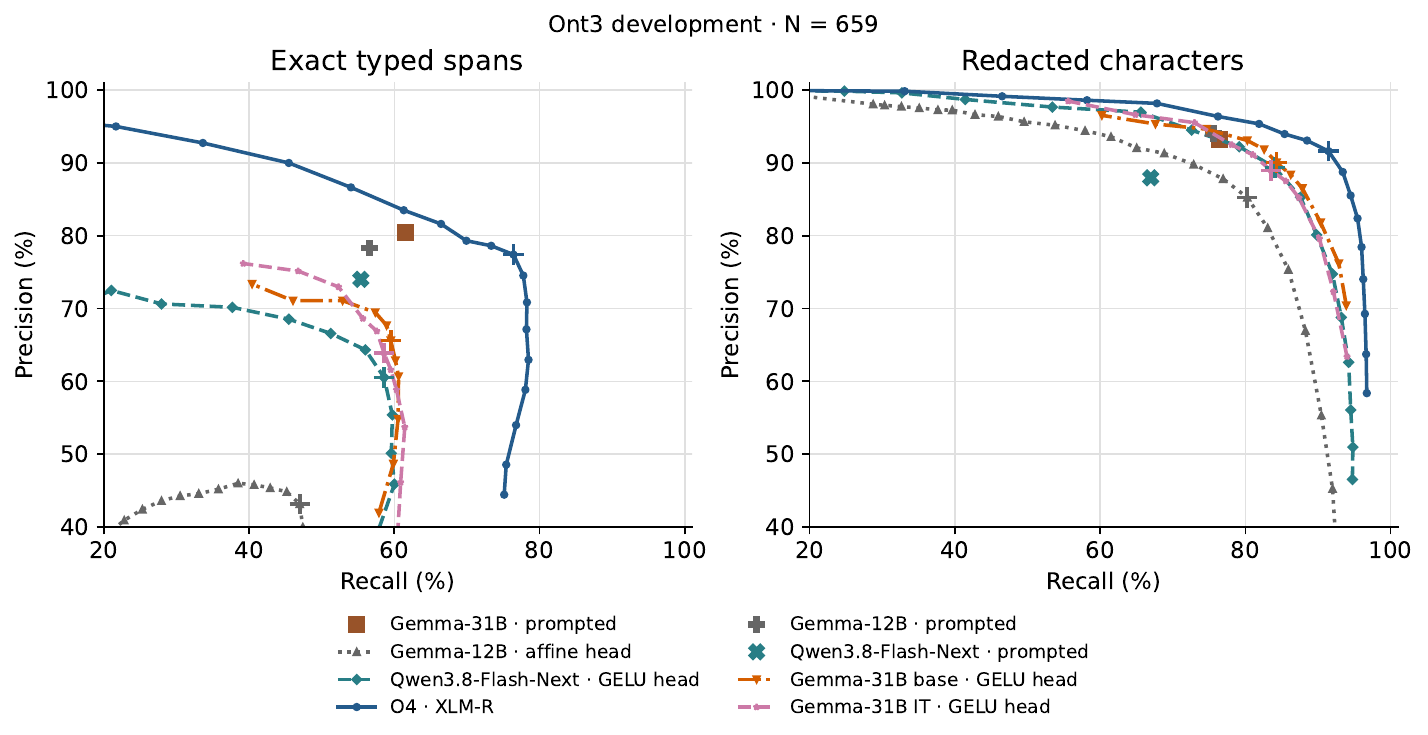}
\else
\includegraphics[alt={Two precision--recall panels compare exact typed spans and redacted characters. O4 has the highest observed F1 in both panels. Four frozen LLM heads and three prompted annotators are shown. Marker shapes and line styles distinguish systems in grayscale.}]{figures/o4-local-llm.pdf}
\fi

}

\caption{\label{fig-local-llm-comparison}Precision--recall comparison on
the same 659 Silver-dev inputs across 35 languages; axes start at 20\%
recall and 40\% precision. Curves vary the \textbf{O} bias before
constrained BIOES decoding; thin plus signs mark zero bias. Standalone
points mark prompted annotation, whose format failures count as empty
predictions without encoder fallback. Current corrected references and
optional-reference scoring apply throughout; counts are pooled across
inputs. Gemma\nobreak\hbox{-}\nobreak{}4-12B uses layer-32
representations conditioned by a short task prompt.
Flash\nobreak\hbox{-}\nobreak{}Next and both
Gemma\nobreak\hbox{-}\nobreak{}4-31B variants use matched 512-unit GELU
heads over final-layer states at token offsets {[}−1, 0, +1{]}, with the
same training targets, annotation prompt and fitting schedule. O4 uses
isolated input, mapped-gold replay and character-boundary refinement.
Name-kind subspans are not scored.}

\end{figure*}

Figure~\ref{fig-local-llm-comparison} compares four frozen-LLM BIOES
heads with the three prompted annotators of
Section~\ref{sec-local-annotators} and our selected O4 model. The frozen
LLM heads recover redacted characters more readily than exact typed
spans; O4 achieves higher F1 on both measures in this comparison.

\subsection{Ontology inventories and
refinements}\label{sec-ontology-structure}

\begin{figure*}[tp]

\centering{

\ifXeTeX
\includegraphics[width=1\linewidth,height=6in,alt={The full Ont1 label forest contains overlapping name components and person roles, detailed location and identifier branches, and many source-specific distinctions.}]{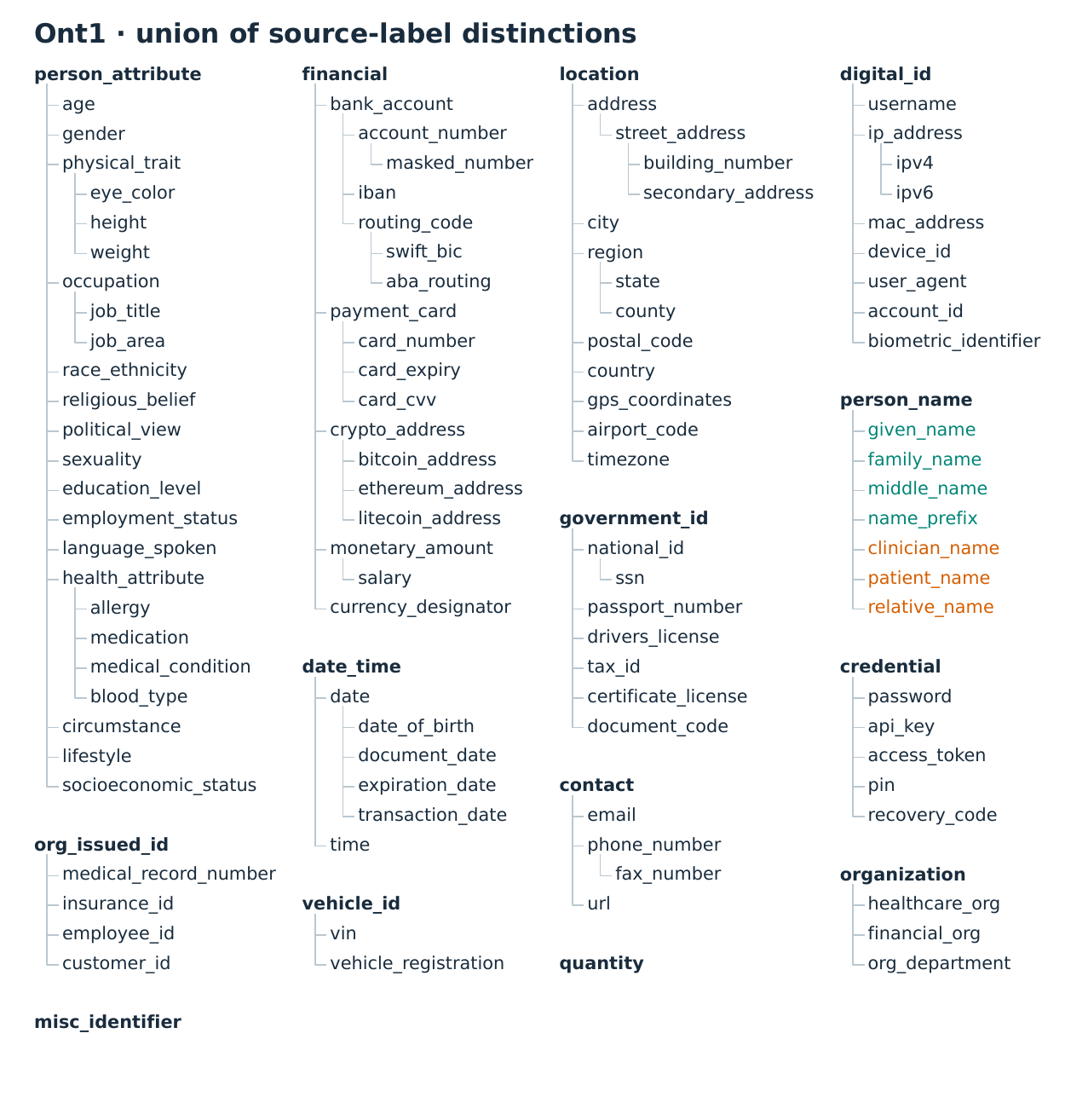}
\else
\includegraphics[alt={The full Ont1 label forest contains overlapping name components and person roles, detailed location and identifier branches, and many source-specific distinctions.}]{figures/ont1-union-inventory-v1.pdf}
\fi

}

\caption{\label{fig-ont1-union}Ont1's union inventory. Branches show the
recorded parent relations, not mutually exclusive subclasses. Under
person\_name, teal name components and orange person roles mix different
annotation axes. The government\_id branch likewise retains the finer
distinctions contributed by source corpora.}

\end{figure*}%

Figure~\ref{fig-name-refinements} shows name components and roles as
separate refinements of primary spans.

Ont3 keeps all 29 Ont2 primary labels and adds only
\texttt{person\_\allowbreak{}reference} and
\texttt{organization\_\allowbreak{}reference}.
Figure~\ref{fig-ont-inventory} shows the shared inventory and marks the
two additions. Its 12 semantic families group the primary labels.

\begin{figure*}[tp]

\centering{

\ifXeTeX
\includegraphics[width=1\linewidth,height=6in,alt={BIOES identifies primary spans; token-level heads predict components and binary attributes; averaging token logits gives one category for a whole span.}]{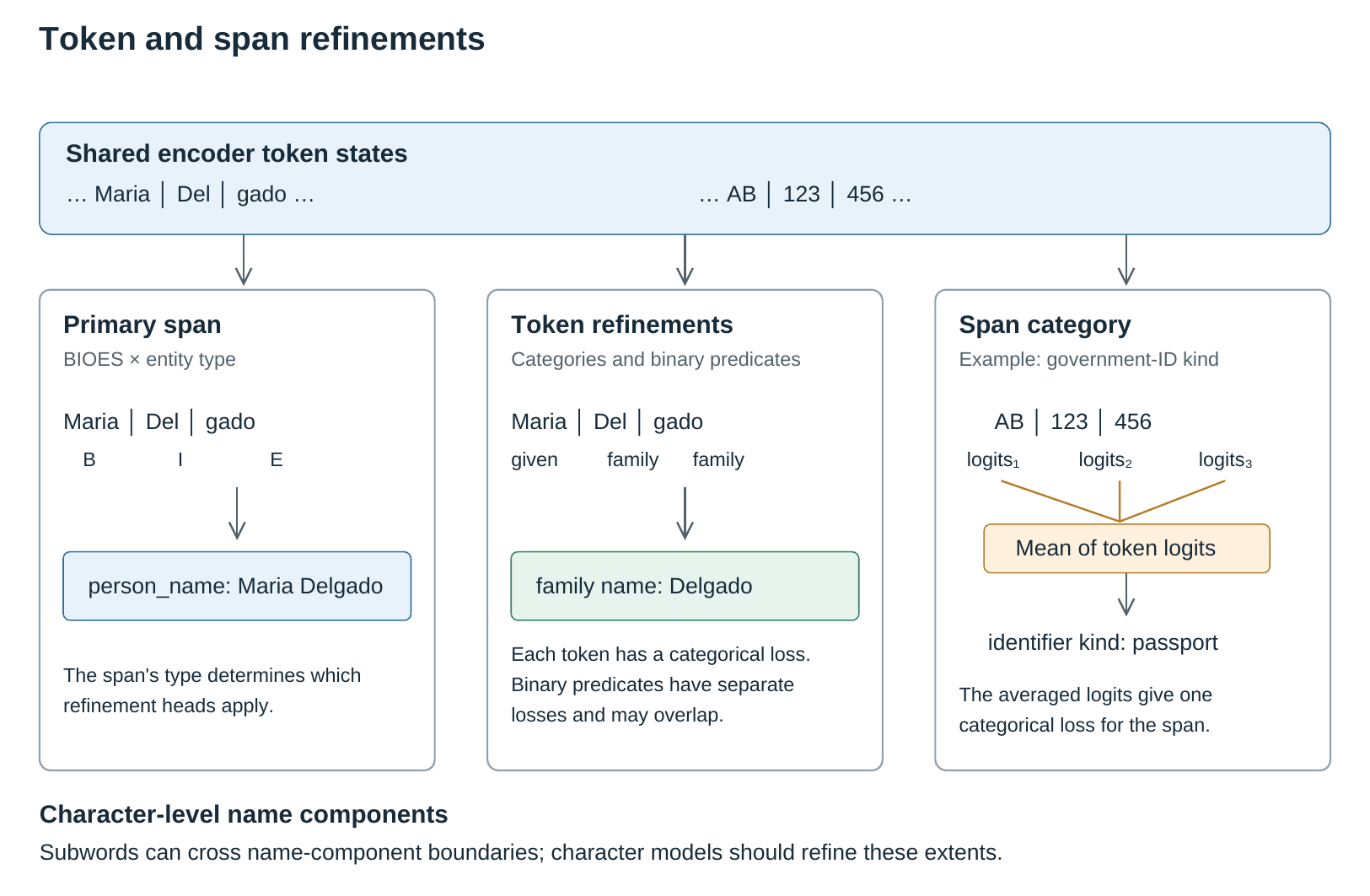}
\else
\includegraphics[alt={BIOES identifies primary spans; token-level heads predict components and binary attributes; averaging token logits gives one category for a whole span.}]{figures/refinement-heads-v1.pdf}
\fi

}

\caption{\label{fig-ont-primary}Refinement heads read the same encoder
states as the BIOES head. Token categories identify name components;
independent binary predicates allow overlapping attributes. A whole-span
category, such as identifier kind, is predicted from the mean token
logits. Refinements do not affect the reported model comparisons.}

\end{figure*}%

Ont3 also extends the refinements attached to primary spans to name
components, identifier kinds, place level, jurisdiction, currency and
person roles; Ont2 already had family-name and care-provider attributes.
Although the two inventories share their primary labels, their
annotation instructions differ: Ont3 explicitly distinguishes names from
grounded references, including where organization determiners fall
relative to the span. We have no independent gold for the component
refinements.

Subwords can cross name-component boundaries, particularly in Asian
languages: one subword may touch two components without enclosing
either. Character models should refine these extents, just as
character-level repair addresses primary-span boundaries
(Section~\ref{sec-boundary-refiner-details}). Our separate name
postprocessor combines a compact character model's given-name and
surname scores with language-specific ordering rules. This
discriminative use differs from generating replacement names
(Section~\ref{sec-character-surface-generation}).

A model moving from one inventory to the next can keep a separate head
for each over one shared encoder and shift loss weight between the heads
gradually; Section~\ref{sec-ontology-transition} describes the
Ont1-to-Ont2 transition.

\subsection{Evaluation populations}\label{sec-evaluation-details}

Gold-7 contains 1,283 distinct test segments in seven languages; Silver
contains 1,201 frontier-annotated development segments across 35
languages, including 12 segments without an attributable sentence
language. Its seven-language subset contains 302 segments. Silver
combines general text and rare-type-targeted text. Silver-dev references
come from two annotation passes using GPT-5.6-Luna, GPT-5.6-Terra and
GPT-5.6-Sol, keeping exact agreements and adjudicating disputes with
GPT-5.6-Sol, followed by manual revision; Silver-test references come
from a single GPT-6-Luna pass.

The current comparisons use O4 in isolation and retain O3 with its
original neighboring-sentence policy: both neighbors where reserved,
otherwise only the preceding sentence. O1 and O2 are the frozen
predecessor checkpoints. GL4 was selected on Silver-dev (659 rows).
Published GLiNER2, GL, receives its supported-category and title
allowances (Section~\ref{sec-title-scoring}). Recipes and scoring
allowances differ between systems, so the curves compare whole systems.

\begin{figure*}[tp]

\centering{

\ifXeTeX
\includegraphics[width=1\linewidth,height=6in,alt={Exact-region comparison for Gold-7 and Silver, with O4 leading the measured aggregate curves.}]{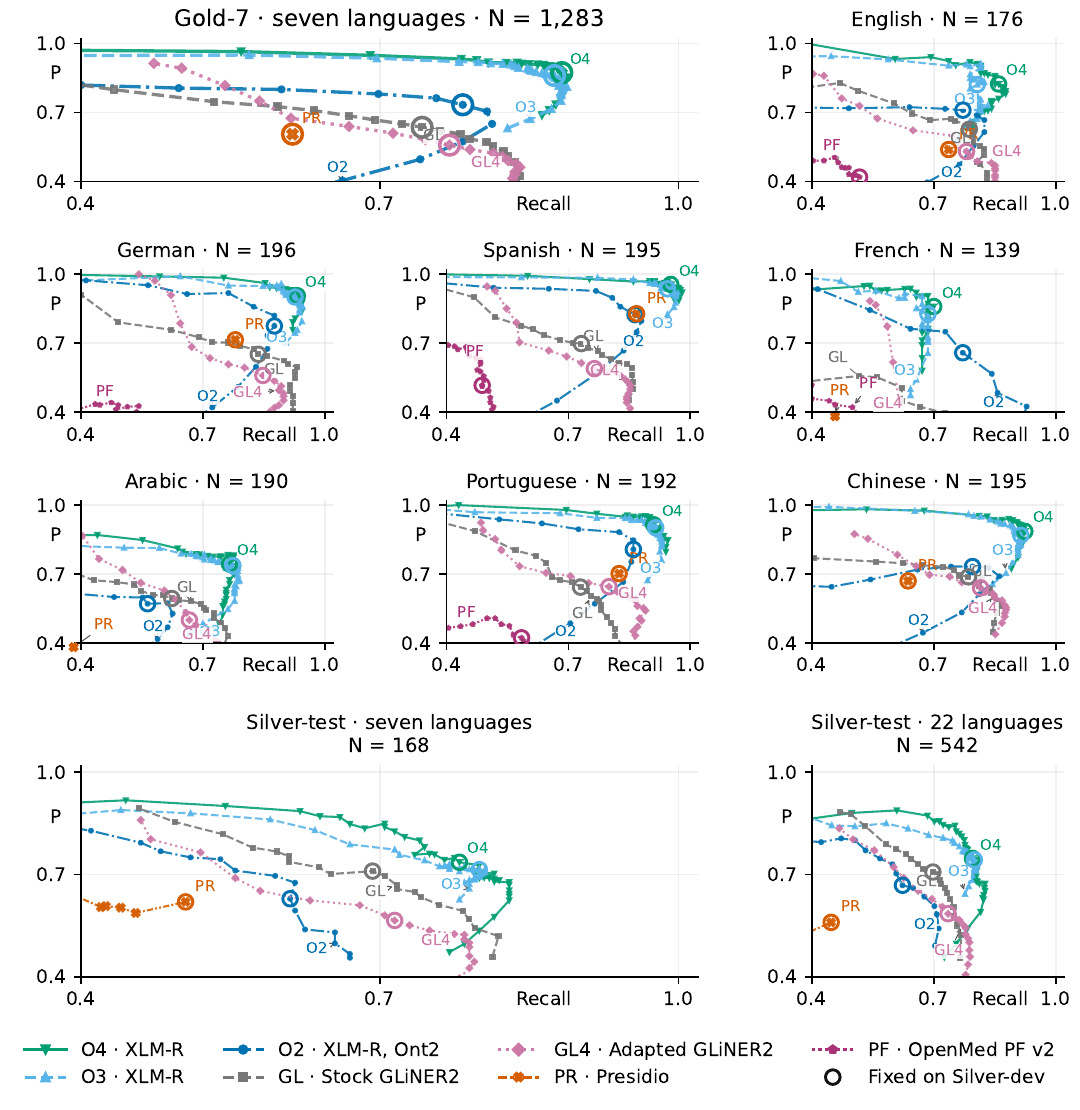}
\else
\includegraphics[alt={Exact-region comparison for Gold-7 and Silver, with O4 leading the measured aggregate curves.}]{figures/o4-human-ont3-overlap100-tr-v1.pdf}
\fi

}

\caption{\label{fig-priority9-exact}The Gold-7 and Silver-test panels of
Figure~\ref{fig-final20-priority9-overlap}, requiring identical
redaction-region boundaries. GL has an unfair advantage: categories its
labels cannot express are excluded from its scoring obligations, and it
may omit job titles. Honorifics are optional; titles may be inside name
spans or separate. All models receive corpus-coverage masks
(Section~\ref{sec-title-scoring}); Figure~\ref{fig-gliner-categories}
compares O4 and GL on identical supported categories.}

\end{figure*}%

\begin{figure*}[tp]

\centering{

\ifXeTeX
\includegraphics[width=1\linewidth,height=6in,alt={Aggregate Silver precision--recall curves for O1 through O4, published GLiNER2 and GL4, Presidio and OpenMed Privacy Filter.}]{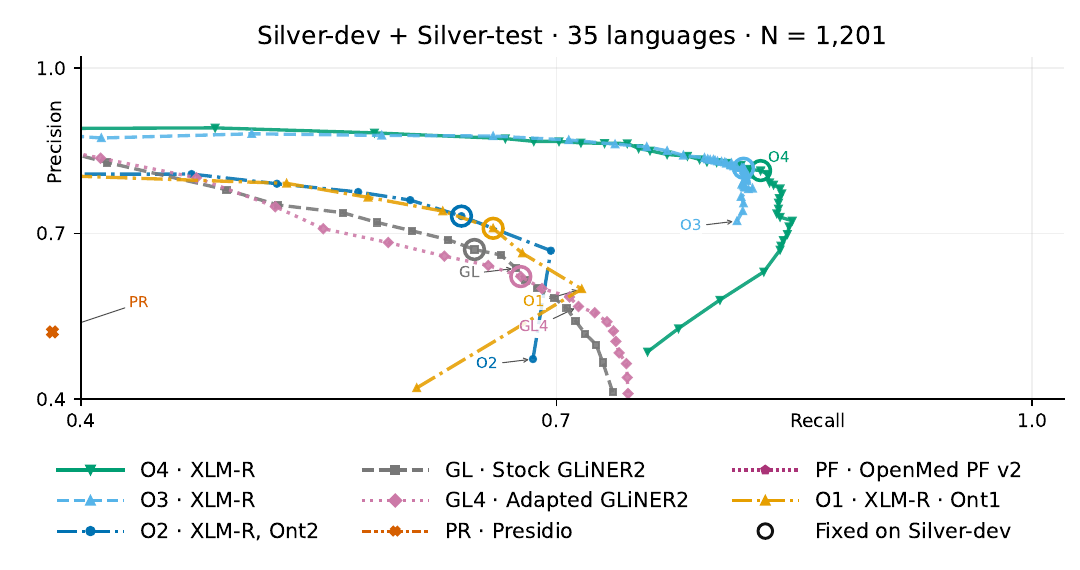}
\else
\includegraphics[alt={Aggregate Silver precision--recall curves for O1 through O4, published GLiNER2 and GL4, Presidio and OpenMed Privacy Filter.}]{figures/o4-ont3-all-overlap80-tr-v1.pdf}
\fi

}

\caption{\label{fig-ont3-priority9-overlap}Silver comparison on 1,201
segments, including the O1 predecessor omitted from the main figure.
Regions match at 80\% symmetric overlap. Rings mark each system's
operating point, fixed on Silver-dev. GL has an unfair advantage:
categories its labels cannot express are excluded from its scoring
obligations, and it may omit job titles. Honorifics are optional; titles
may be inside name spans or separate. All models receive corpus-coverage
masks (Section~\ref{sec-title-scoring});
Figure~\ref{fig-gliner-categories} compares O4 and GL on identical
supported categories.}

\end{figure*}%

Lines connect measured settings in threshold order, including backward
bends. Changing the outside-tag bias can replace spans or move their
boundaries; thresholding fixed GLiNER2 spans can split a merged
redaction region. A stricter setting may therefore improve both
precision and recall. Long terminal backward connectors are omitted,
with measured endpoints kept. Both axes start at 0.4; off-scale Presidio
points are marked at the axis edge.

Silver-dev, the 659-row selection subset, has been repeatedly inspected.
Language assignments were reviewed against a saved sentence-language
detector, and annotations were corrected after error inspection. Exact
and NFC/whitespace checks found no matches against 18 recovered training
sets, but training ancestry and semantic overlap were not fully
certified. On Gold-7, corpus test membership does not make the
repeatedly scored subset a fresh selection holdout. Wojood demographic
predictions outside annotated demographic spans are unscored for every
system because that corpus does not exhaustively cover the category;
overlapping predictions keep their full boundaries.

\subsection{Earlier evaluation sets}\label{sec-evaluation-sets}

These collections preceded the Gold-7 and Silver populations of the main
comparison; appendix technique comparisons still report on them. Fresh6
is an independently authored 426-document diagnostic suite spanning
Chinese, Japanese, Korean, Spanish, German, and French. Fresh7 adds 61
English documents. Fresh20 combines that frozen seven-language suite
with 780 later documents---60 in each of thirteen added languages---for
1,267 documents across twenty languages. Scoped MultiGraSCCo
\citep{baroud2026-multigrascco} contains 630 synthetic clinical
document-language instances, 63 in each of ten languages.

The twenty-language Final20 collection contains 7,000 records but only
3,212 distinct texts across twenty languages. Repeated texts also cross
the boundary between its calibration and evaluation records. We
therefore do not use it for the main precision--recall comparison or for
document-count-based uncertainty and dominance claims.

Language-weighted scores on these sets use importance weights reflecting
expected use. The nine priority languages are those weighted ×2 or ×4.
The same weights guided how annotation effort was spread across the 35
languages, with a minimum of about 0.5\% of annotated examples per
language as the first priority. Because training started heavy in the
gold-annotated languages, the realized mixtures do not match these
weights exactly (Section~\ref{sec-training-mixtures}). Codes follow the
dataset's language tags; \texttt{fil} denotes Filipino. The pooled
Gold-7 and Silver comparisons sum counts and use no weights.

\ACLwidetabletrue\ACLcontinuedtablefalse

\ifXeTeX
\begin{longtable}[]{@{}
  >{\raggedright\arraybackslash}p{(\linewidth - 4\tabcolsep) * \real{0.1429}}
  >{\raggedleft\arraybackslash}p{(\linewidth - 4\tabcolsep) * \real{0.1429}}
  >{\raggedright\arraybackslash}p{(\linewidth - 4\tabcolsep) * \real{0.7143}}@{}}
\caption{Language weights in historical evaluation sets. The nine
priority languages have weight ×2 or ×4. Main Silver and Gold-7
comparisons use pooled counts without language
weights.}\label{tbl-language-weights}\tabularnewline
\toprule\noalign{}
\begin{minipage}[b]{\linewidth}\raggedright
Coverage
\end{minipage} & \begin{minipage}[b]{\linewidth}\raggedleft
Importance
\end{minipage} & \begin{minipage}[b]{\linewidth}\raggedright
Language codes
\end{minipage} \\
\midrule\noalign{}
\endfirsthead
\toprule\noalign{}
\begin{minipage}[b]{\linewidth}\raggedright
Coverage
\end{minipage} & \begin{minipage}[b]{\linewidth}\raggedleft
Importance
\end{minipage} & \begin{minipage}[b]{\linewidth}\raggedright
Language codes
\end{minipage} \\
\midrule\noalign{}
\endhead
\bottomrule\noalign{}
\endlastfoot
Original 20 & ×4 & en \\
& ×2 & ar, de, es, fr, ko, pt, vi, zh \\
& ×1 & cs, hi, id, it, ja, nl, pl, ru, sv, tr, uk \\
+15 → 35 & ×1 & bn, da, el, fa, fi, fil, he, hr, ms, no, ro, ta, te, th,
ur \\
\end{longtable}
\else
\begin{longtable}[]{@{}
  >{\raggedright\arraybackslash}X
  >{\raggedleft\arraybackslash}X
  >{\raggedright\arraybackslash}X@{}}
\caption{Language weights in historical evaluation sets. The nine
priority languages have weight ×2 or ×4. Main Silver and Gold-7
comparisons use pooled counts without language
weights.}\label{tbl-language-weights}\tabularnewline
\toprule\noalign{}
\raggedright
Coverage
 & \raggedleft
Importance
 & \raggedright
Language codes
 \\
\midrule\noalign{}
\endfirsthead
\toprule\noalign{}
\raggedright
Coverage
 & \raggedleft
Importance
 & \raggedright
Language codes
 \\
\midrule\noalign{}
\endhead
\bottomrule\noalign{}
\endlastfoot
Original 20 & ×4 & en \\
& ×2 & ar, de, es, fr, ko, pt, vi, zh \\
& ×1 & cs, hi, id, it, ja, nl, pl, ru, sv, tr, uk \\
+15 → 35 & ×1 & bn, da, el, fa, fi, fil, he, hr, ms, no, ro, ta, te, th,
ur \\
\end{longtable}
\fi

\subsection{Span, character, and ontology views}\label{sec-span-views}

Spans use half-open Unicode codepoint offsets into the supplied text,
after any declared preprocessing. Changing Unicode normalization
requires remapping the offsets; each model card specifies the
normalization and offset contract. Exact span matching requires
identical boundaries. To tolerate small boundary misses, overlap scoring
also accepts a prediction when its intersection with the gold span
covers at least 80\% of each span. Among eligible pairs, we choose the
largest set of matches in which neither a prediction nor a gold span
appears twice. This prevents duplicate or fragmented predictions from
earning repeated credit for one entity.

Fine-grained scoring retains the most specific jointly expressible
label. Each comparison applies its declared mapping symmetrically and
reports excluded labels.

Character precision is the fraction of redacted Unicode character
positions covered by gold spans; recall is the fraction of gold
character positions redacted. F1 is their harmonic mean. We score the
final spans after boundary repair, counting each character once.
Character scoring gives proportional credit to partial coverage; span
scoring tests entity localization.

The span examples in Section~\ref{sec-published-filter-gains} use
configurations selected for 90\% character recall in the twenty-language
Final20 comparison (Section~\ref{sec-evaluation-sets}). Their displayed
spans ignore type; the examples were chosen for legibility.

Reference spans, the \texttt{person} and \texttt{organization}
\texttt{\_\allowbreak{}reference} types in Ont3 (such as ``the suspect''
in Figure~\ref{fig-luna-result}), are optional for redaction: leaving
referential wording visible can preserve readable text after identifying
names are removed. We exclude reference targets and predictions, and
neutralize an exact-boundary \texttt{person\_\allowbreak{}name}
prediction on \texttt{person\_\allowbreak{}reference}, or
\texttt{organization} on \texttt{organization\_\allowbreak{}reference}.
Missing an optional reference incurs no false negative. Wrong entity
families and boundary errors remain errors; neutral spans are also
removed before character scoring. This avoids penalizing low-agreement
distinctions or corpora that intentionally use the base label.
\textbf{\texttt{+\_\allowbreak{}reference}} instead requires the
specific reference label and is a fair task comparison only for models
trained to that Ont3 distinction.

\subsubsection{Titles and expressible
categories}\label{sec-title-scoring}

We accommodate title conventions before redaction-region matching. An
annotated honorific beside or inside a person name is neutral:
predicting ``Dr.~Maya Chen'', ``Maya Chen'', or separate spans for
``Dr.'' and ``Maya Chen'' receives the same credit for the name.
Predicting only ``Dr.'' does not recover the name. The scorer removes
the known title extent from both sides before matching the remaining
name; it does not forgive other boundary errors.

Job titles, offices and professions are distinct from honorifics. In
``President Maya Chen'', ``President'' is a demographic attribute. A
model capable of that category must cover the title, but may include it
in the name span or emit it separately: title coverage is scored as its
own region. For a model unable to express demographic attributes,
leaving the title visible or redacting it is free. Thus the allowance is
not that all titles are optional for every model. Known title extents
come from corpus fine labels, separate Ont3 title spans and a reviewed
neutral-honorific sidecar; the evaluator does not guess arbitrary title
boundaries from a prediction.

Published GLiNER2 is prompted with its 42 PII labels plus organization.
Gold categories outside their reviewed mappings are removed from its
task; its fine Ont3-tuned variants are evaluated on the full inventory.
This gives the published model a less demanding target, so its aggregate
score cannot by itself measure the effect of adaptation. All systems
also receive the source-corpus coverage masks: an unannotated type is
not automatically a false positive. For partially covered categories, a
prediction overlapping an annotated example keeps its full extent, so
coverage masking cannot erase its boundary error. Untyped region
matching further accepts different internal type assignments when they
imply the same redaction.

\subsection{Published comparison systems}\label{sec-incumbents}

\subsubsection{Presidio configuration}\label{sec-presidio-configuration}

We run Microsoft Presidio 2.2.364 from the data-privacy-stack
multilingual fork, with spaCy 3.8.16 and model packages 3.8.0. Fourteen
languages use their language-specific large spaCy model; the others use
\texttt{xx\_\allowbreak{}ent\_\allowbreak{}wiki\_\allowbreak{}sm}. We
keep Presidio's rule recognizers, run at score threshold zero, and sweep
the threshold by filtering the saved recognizer scores, which in this
fork matches rerunning at each threshold. The spaCy entity recognizer
reports a constant 0.85, so the sweep moves only pattern-recognizer
detections: Presidio's Gold-7 curve is a single point, and its
Silver-dev-fixed threshold of 0.2 removes one Silver-dev false positive.
Presidio reads Gold-7's Chinese rows without the corpus's
word-separating spaces, as its Chinese pipeline expects
(Section~\ref{sec-operating-points}). Filipino and language-free inputs
run under engine codes \texttt{tl} and \texttt{xx}, respectively, but
are still scored under their original language assignments.

\subsubsection{Shared-category
comparisons}\label{sec-shared-category-comparisons}

\begin{figure*}[tp]

\centering{

\ifXeTeX
\includegraphics[width=1\linewidth,height=6in,alt={Four character precision--recall panels compare O4 and Privacy Filter on supported categories, across all languages and English only.}]{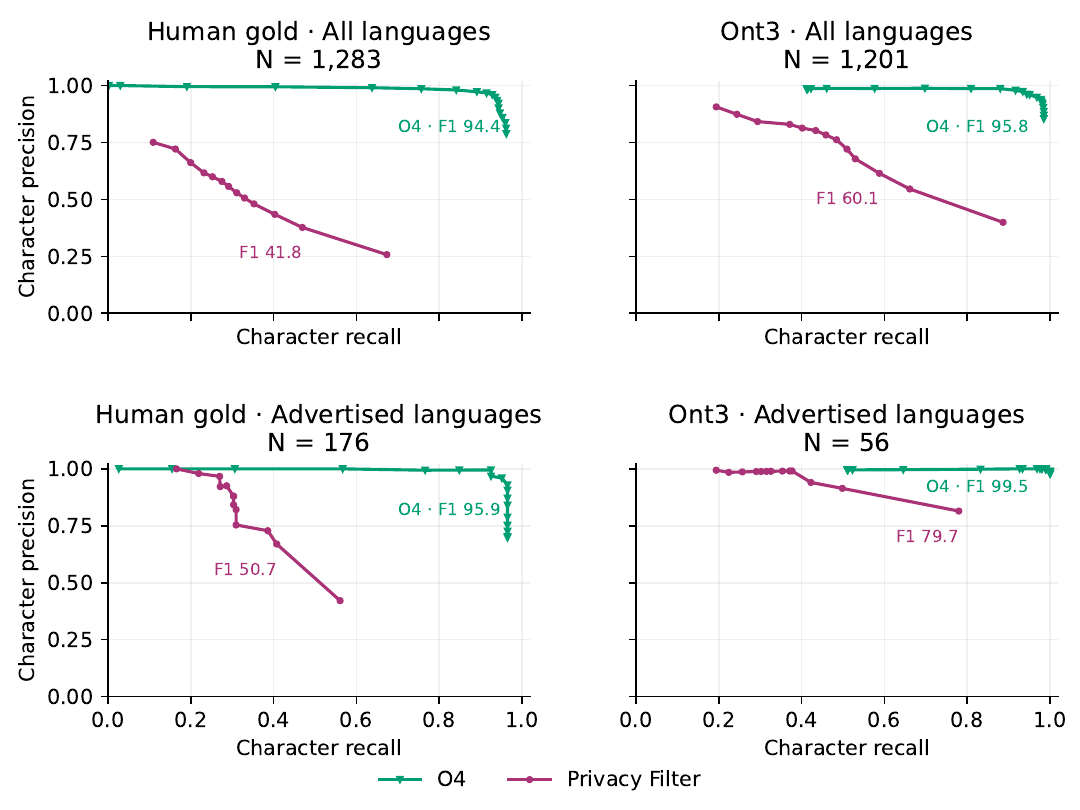}
\else
\includegraphics[alt={Four character precision--recall panels compare O4 and Privacy Filter on supported categories, across all languages and English only.}]{figures/o4-opf-representable.pdf}
\fi

}

\caption{\label{fig-privacy-filter-categories}O4 and the original OpenAI
Privacy Filter (PF) on a reduced task that scores only the entity
categories covered by PF's eight labels. Organizations, generic
locations and every other category outside PF's labels are removed from
the task, gold and predictions alike, preserving many-to-many
alternatives admitted by coarse gold labels, without retraining either
model; characters inside removed gold regions are not scored. Curves
sweep each model's decision threshold; labels give each curve's best
character F1. Columns are Gold-7 and Silver; the top row uses all
languages (1,283 and 1,201 segments), the bottom English only (176 and
56), the primary language named by PF's model card. Both models use the
same subset in each panel.}

\end{figure*}

The original OpenAI Privacy Filter (PF) predicts eight entity labels and
covers neither organizations nor general locations by design.
Figure~\ref{fig-privacy-filter-categories} therefore scores PF against
O4, our selected Ont3-trained XLM\nobreak\hbox{-}\nobreak{}R model, only
on the categories PF can express.

\begin{figure*}[tp]

\centering{

\ifXeTeX
\includegraphics[width=1\linewidth,height=6in,alt={Four O4 and GLiNER2 precision--recall panels on identical category obligations, across all languages and GLiNER2\textquotesingle s advertised languages. Silver-test excludes checkpoint-selection inputs.}]{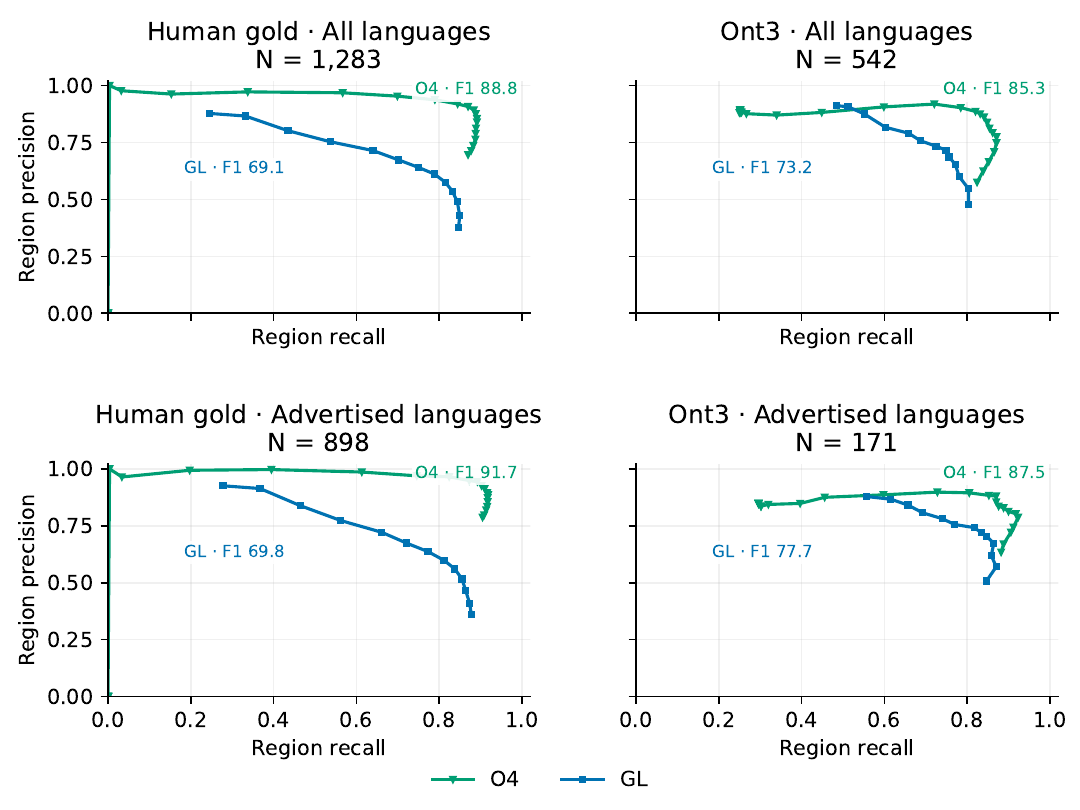}
\else
\includegraphics[alt={Four O4 and GLiNER2 precision--recall panels on identical category obligations, across all languages and GLiNER2\textquotesingle s advertised languages. Silver-test excludes checkpoint-selection inputs.}]{figures/o4-gliner-representable-nonselection.pdf}
\fi

}

\caption{\label{fig-gliner-categories}O4 and published GLiNER2 (GL) with
identical category obligations. Both models are scored only on
categories expressible by GL's 42 requested labels plus organization,
with the same title allowances and corpus-coverage masks. O4 predictions
outside that inventory are removed as well, preserving many-to-many
alternatives admitted by coarse gold labels. Unlike the main comparison,
GL receives no exclusive category exemption. Untyped regions match at
80\% symmetric overlap; labels give best measured redaction F1. Columns
are Gold-7 and Silver. The Silver panels use only Silver-test, the 542
segments not used to select O4 or GL4. The top row uses all languages
(1,283 and 542 segments); the bottom uses GL's advertised English,
French, Spanish, German, Italian, Portuguese and Dutch (898 and 171).
Both models use the same subset in each panel.}

\end{figure*}%

\subsubsection{GLiNER2 adaptation
conditions}\label{sec-gliner2-adaptation}

GL4 starts from the published multilingual PII checkpoint, with Ont3
training spans snapped outward to its word tokenizer. Its training
mixture includes the expanded frontier annotations and mapped gold,
using O4's 50\%-mapped-gold sampling recipe (Table~\ref{tbl-o4-data};
Section~\ref{sec-o4-annotation-recipe}). From 50,000 weighted draws,
processor validation admits 32,204 windows and 70,269 positive word
spans. All 35 languages meet the 0.75\% sampling floor. Evaluation
boundaries are unchanged. Mapped-gold targets use the mapping's
single-label fallback; rows with unknown coverage receive positive-only
tasks. This differs from the encoder's many-to-many loss.

Each Ont3 type receives a familiar prompt name where possible.
Twenty-one new atomic label tokens use weighted averages of donor
subtoken embeddings (weight three for the preferred donor, one for each
alternative); ten names retain their tokenization and embeddings.
Predictions map back to Ont3. The initial model below includes this
transfer, so its comparison with later checkpoints holds the target
inventory and scoring obligations fixed.

Each training example names the types present in its window and eight
sampled absent types. GL4 shuffles their order per example. Our first
adaptation kept present types first, and its checkpoints learned
position as a cue: late in training they favored whichever types the
inference prompt listed first.

We train for 6,000 updates, batch eight with four-step accumulation and
encoder/task peak rates \(10^{-5}\)/\(2\times10^{-5}\). The cosine
schedule warms up over 1,000 steps and decays by step 2,000, then
restarts with fresh optimizer state, 500 warmup steps and decay over the
remaining 4,000. Training takes 29 minutes 46 seconds on an RTX PRO 6000
Blackwell, including a checkpoint every 100 steps. Validation loss on
249 representable selector windows would choose step 4,000; we instead
compare 13 stages on the 659 Silver-dev rows at confidence 0.5, by the
encoder's language-weighted criterion restricted to Silver-dev: 40\%
exact regions and 60\% exact typed spans, pooling weighted counts before
F1. Step 2,500 scores 55.52, ahead of step 2,000 (54.76). The first
adaptation trained for 2,000 updates; its selection also weighted 23,251
gold examples at 70\% and chose step 600.

\ACLwidetabletrue\ACLcontinuedtablefalse

\ifXeTeX
\begin{longtable}[]{@{}
  >{\raggedright\arraybackslash}p{(\linewidth - 8\tabcolsep) * \real{0.1579}}
  >{\raggedleft\arraybackslash}p{(\linewidth - 8\tabcolsep) * \real{0.2105}}
  >{\raggedleft\arraybackslash}p{(\linewidth - 8\tabcolsep) * \real{0.2105}}
  >{\raggedleft\arraybackslash}p{(\linewidth - 8\tabcolsep) * \real{0.2105}}
  >{\raggedleft\arraybackslash}p{(\linewidth - 8\tabcolsep) * \real{0.2105}}@{}}
\caption{Exact-region F1 (\%) at confidence 0.5 on the paper's 1,283
Gold-7 segments and the 659-segment Silver-dev selection subset, for the
first adaptation (unshuffled type order, 2,000 updates) and GL4
(shuffled), from the same transferred initialization and data. Without
shuffling, quality collapses after step 600; with it, Silver-dev peaks
at step 2,500 while Gold-7 declines slowly. Gold-7 does not choose GL4's
checkpoint. }\label{tbl-gl4-trajectory}\tabularnewline
\toprule\noalign{}
\begin{minipage}[b]{\linewidth}\raggedright
Updates
\end{minipage} & \begin{minipage}[b]{\linewidth}\raggedleft
Unshuffled Gold-7
\end{minipage} & \begin{minipage}[b]{\linewidth}\raggedleft
Unshuffled Silver-dev
\end{minipage} & \begin{minipage}[b]{\linewidth}\raggedleft
GL4 Gold-7
\end{minipage} & \begin{minipage}[b]{\linewidth}\raggedleft
GL4 Silver-dev
\end{minipage} \\
\midrule\noalign{}
\endfirsthead
\toprule\noalign{}
\begin{minipage}[b]{\linewidth}\raggedright
Updates
\end{minipage} & \begin{minipage}[b]{\linewidth}\raggedleft
Unshuffled Gold-7
\end{minipage} & \begin{minipage}[b]{\linewidth}\raggedleft
Unshuffled Silver-dev
\end{minipage} & \begin{minipage}[b]{\linewidth}\raggedleft
GL4 Gold-7
\end{minipage} & \begin{minipage}[b]{\linewidth}\raggedleft
GL4 Silver-dev
\end{minipage} \\
\midrule\noalign{}
\endhead
\bottomrule\noalign{}
\endlastfoot
0 (transferred initialization) & 57.44 & 50.09 & 57.44 & 50.09 \\
100 & 63.73 & 55.43 & 63.95 & 55.17 \\
200 & 64.72 & 57.46 & 64.25 & 56.69 \\
300 & 64.52 & 56.52 & 63.82 & 55.37 \\
400 & 62.98 & 57.66 & 62.97 & 56.97 \\
600 (selected, unshuffled) & 67.24 & 55.42 & 63.27 & 58.76 \\
\end{longtable}
\else
\begin{longtable}[]{@{}
  >{\raggedright\arraybackslash}X
  >{\raggedleft\arraybackslash}X
  >{\raggedleft\arraybackslash}X
  >{\raggedleft\arraybackslash}X
  >{\raggedleft\arraybackslash}X@{}}
\caption{Exact-region F1 (\%) at confidence 0.5 on the paper's 1,283
Gold-7 segments and the 659-segment Silver-dev selection subset, for the
first adaptation (unshuffled type order, 2,000 updates) and GL4
(shuffled), from the same transferred initialization and data. Without
shuffling, quality collapses after step 600; with it, Silver-dev peaks
at step 2,500 while Gold-7 declines slowly. Gold-7 does not choose GL4's
checkpoint. }\label{tbl-gl4-trajectory}\tabularnewline
\toprule\noalign{}
\raggedright
Updates
 & \raggedleft
Unshuffled Gold-7
 & \raggedleft
Unshuffled Silver-dev
 & \raggedleft
GL4 Gold-7
 & \raggedleft
GL4 Silver-dev
 \\
\midrule\noalign{}
\endfirsthead
\toprule\noalign{}
\raggedright
Updates
 & \raggedleft
Unshuffled Gold-7
 & \raggedleft
Unshuffled Silver-dev
 & \raggedleft
GL4 Gold-7
 & \raggedleft
GL4 Silver-dev
 \\
\midrule\noalign{}
\endhead
\bottomrule\noalign{}
\endlastfoot
0 (transferred initialization) & 57.44 & 50.09 & 57.44 & 50.09 \\
100 & 63.73 & 55.43 & 63.95 & 55.17 \\
200 & 64.72 & 57.46 & 64.25 & 56.69 \\
300 & 64.52 & 56.52 & 63.82 & 55.37 \\
400 & 62.98 & 57.66 & 62.97 & 56.97 \\
600 (selected, unshuffled) & 67.24 & 55.42 & 63.27 & 58.76 \\
\end{longtable}
\fi

\ACLwidetabletrue\ACLcontinuedtabletrue

\ifXeTeX
\begin{longtable}[]{@{}
  >{\raggedright\arraybackslash}p{(\linewidth - 8\tabcolsep) * \real{0.1579}}
  >{\raggedleft\arraybackslash}p{(\linewidth - 8\tabcolsep) * \real{0.2105}}
  >{\raggedleft\arraybackslash}p{(\linewidth - 8\tabcolsep) * \real{0.2105}}
  >{\raggedleft\arraybackslash}p{(\linewidth - 8\tabcolsep) * \real{0.2105}}
  >{\raggedleft\arraybackslash}p{(\linewidth - 8\tabcolsep) * \real{0.2105}}@{}}
\caption{Exact-region F1 (\%) at confidence 0.5 on the paper's 1,283 Gold-7 segments and the 659-segment Silver-dev selection subset, for the first adaptation (unshuffled type order, 2,000 updates) and GL4 (shuffled), from the same transferred initialization and data. Without shuffling, quality collapses after step 600; with it, Silver-dev peaks at step 2,500 while Gold-7 declines slowly. Gold-7 does not choose GL4's checkpoint.  (continued)}\tabularnewline
\toprule\noalign{}
\begin{minipage}[b]{\linewidth}\raggedright
Updates
\end{minipage} & \begin{minipage}[b]{\linewidth}\raggedleft
Unshuffled Gold-7
\end{minipage} & \begin{minipage}[b]{\linewidth}\raggedleft
Unshuffled Silver-dev
\end{minipage} & \begin{minipage}[b]{\linewidth}\raggedleft
GL4 Gold-7
\end{minipage} & \begin{minipage}[b]{\linewidth}\raggedleft
GL4 Silver-dev
\end{minipage} \\
\midrule\noalign{}
\endfirsthead
\toprule\noalign{}
\begin{minipage}[b]{\linewidth}\raggedright
Updates
\end{minipage} & \begin{minipage}[b]{\linewidth}\raggedleft
Unshuffled Gold-7
\end{minipage} & \begin{minipage}[b]{\linewidth}\raggedleft
Unshuffled Silver-dev
\end{minipage} & \begin{minipage}[b]{\linewidth}\raggedleft
GL4 Gold-7
\end{minipage} & \begin{minipage}[b]{\linewidth}\raggedleft
GL4 Silver-dev
\end{minipage} \\
\midrule\noalign{}
\endhead
\bottomrule\noalign{}
\endlastfoot
1,000 & 60.67 & 50.35 & 60.93 & 58.03 \\
2,000 & 41.19 & 33.93 & 60.31 & 58.59 \\
2,500 (selected, GL4) & --- & --- & 62.79 & 58.93 \\
3,000 & --- & --- & 59.39 & 57.16 \\
4,000 & --- & --- & 60.34 & 57.33 \\
5,000 & --- & --- & 59.39 & 54.33 \\
\end{longtable}
\else
\begin{longtable}[]{@{}
  >{\raggedright\arraybackslash}X
  >{\raggedleft\arraybackslash}X
  >{\raggedleft\arraybackslash}X
  >{\raggedleft\arraybackslash}X
  >{\raggedleft\arraybackslash}X@{}}
\caption{Exact-region F1 (\%) at confidence 0.5 on the paper's 1,283 Gold-7 segments and the 659-segment Silver-dev selection subset, for the first adaptation (unshuffled type order, 2,000 updates) and GL4 (shuffled), from the same transferred initialization and data. Without shuffling, quality collapses after step 600; with it, Silver-dev peaks at step 2,500 while Gold-7 declines slowly. Gold-7 does not choose GL4's checkpoint.  (continued)}\tabularnewline
\toprule\noalign{}
\raggedright
Updates
 & \raggedleft
Unshuffled Gold-7
 & \raggedleft
Unshuffled Silver-dev
 & \raggedleft
GL4 Gold-7
 & \raggedleft
GL4 Silver-dev
 \\
\midrule\noalign{}
\endfirsthead
\toprule\noalign{}
\raggedright
Updates
 & \raggedleft
Unshuffled Gold-7
 & \raggedleft
Unshuffled Silver-dev
 & \raggedleft
GL4 Gold-7
 & \raggedleft
GL4 Silver-dev
 \\
\midrule\noalign{}
\endhead
\bottomrule\noalign{}
\endlastfoot
1,000 & 60.67 & 50.35 & 60.93 & 58.03 \\
2,000 & 41.19 & 33.93 & 60.31 & 58.59 \\
2,500 (selected, GL4) & --- & --- & 62.79 & 58.93 \\
3,000 & --- & --- & 59.39 & 57.16 \\
4,000 & --- & --- & 60.34 & 57.33 \\
5,000 & --- & --- & 59.39 & 54.33 \\
\end{longtable}
\fi

\ACLwidetabletrue\ACLcontinuedtabletrue

\ifXeTeX
\begin{longtable}[]{@{}
  >{\raggedright\arraybackslash}p{(\linewidth - 8\tabcolsep) * \real{0.1579}}
  >{\raggedleft\arraybackslash}p{(\linewidth - 8\tabcolsep) * \real{0.2105}}
  >{\raggedleft\arraybackslash}p{(\linewidth - 8\tabcolsep) * \real{0.2105}}
  >{\raggedleft\arraybackslash}p{(\linewidth - 8\tabcolsep) * \real{0.2105}}
  >{\raggedleft\arraybackslash}p{(\linewidth - 8\tabcolsep) * \real{0.2105}}@{}}
\caption{Exact-region F1 (\%) at confidence 0.5 on the paper's 1,283 Gold-7 segments and the 659-segment Silver-dev selection subset, for the first adaptation (unshuffled type order, 2,000 updates) and GL4 (shuffled), from the same transferred initialization and data. Without shuffling, quality collapses after step 600; with it, Silver-dev peaks at step 2,500 while Gold-7 declines slowly. Gold-7 does not choose GL4's checkpoint.  (continued)}\tabularnewline
\toprule\noalign{}
\begin{minipage}[b]{\linewidth}\raggedright
Updates
\end{minipage} & \begin{minipage}[b]{\linewidth}\raggedleft
Unshuffled Gold-7
\end{minipage} & \begin{minipage}[b]{\linewidth}\raggedleft
Unshuffled Silver-dev
\end{minipage} & \begin{minipage}[b]{\linewidth}\raggedleft
GL4 Gold-7
\end{minipage} & \begin{minipage}[b]{\linewidth}\raggedleft
GL4 Silver-dev
\end{minipage} \\
\midrule\noalign{}
\endfirsthead
\toprule\noalign{}
\begin{minipage}[b]{\linewidth}\raggedright
Updates
\end{minipage} & \begin{minipage}[b]{\linewidth}\raggedleft
Unshuffled Gold-7
\end{minipage} & \begin{minipage}[b]{\linewidth}\raggedleft
Unshuffled Silver-dev
\end{minipage} & \begin{minipage}[b]{\linewidth}\raggedleft
GL4 Gold-7
\end{minipage} & \begin{minipage}[b]{\linewidth}\raggedleft
GL4 Silver-dev
\end{minipage} \\
\midrule\noalign{}
\endhead
\bottomrule\noalign{}
\endlastfoot
6,000 & --- & --- & 59.94 & 53.91 \\
\end{longtable}
\else
\begin{longtable}[]{@{}
  >{\raggedright\arraybackslash}X
  >{\raggedleft\arraybackslash}X
  >{\raggedleft\arraybackslash}X
  >{\raggedleft\arraybackslash}X
  >{\raggedleft\arraybackslash}X@{}}
\caption{Exact-region F1 (\%) at confidence 0.5 on the paper's 1,283 Gold-7 segments and the 659-segment Silver-dev selection subset, for the first adaptation (unshuffled type order, 2,000 updates) and GL4 (shuffled), from the same transferred initialization and data. Without shuffling, quality collapses after step 600; with it, Silver-dev peaks at step 2,500 while Gold-7 declines slowly. Gold-7 does not choose GL4's checkpoint.  (continued)}\tabularnewline
\toprule\noalign{}
\raggedright
Updates
 & \raggedleft
Unshuffled Gold-7
 & \raggedleft
Unshuffled Silver-dev
 & \raggedleft
GL4 Gold-7
 & \raggedleft
GL4 Silver-dev
 \\
\midrule\noalign{}
\endfirsthead
\toprule\noalign{}
\raggedright
Updates
 & \raggedleft
Unshuffled Gold-7
 & \raggedleft
Unshuffled Silver-dev
 & \raggedleft
GL4 Gold-7
 & \raggedleft
GL4 Silver-dev
 \\
\midrule\noalign{}
\endhead
\bottomrule\noalign{}
\endlastfoot
6,000 & --- & --- & 59.94 & 53.91 \\
\end{longtable}
\fi

The unshuffled collapse includes a dropped URL and location names
retyped as administrative areas, the first type listed. We have not
isolated whether replay of the original GLiNER2 mixture would prevent
GL4's remaining Gold-7 decline: that mixture is not publicly available.
At their Silver-dev operating points GL4 scores 65.1 versus 66.5
redaction F1 on Gold-7, whose rows helped select the first (−1.5
{[}−3.3, 0.3{]}), and 66.9 versus 64.5 on Silver-test. Over all of
Silver its redaction F1 is 2.6 points higher {[}0.9, 4.3{]} and its
exact typed F1 3.8 points higher {[}2.3, 5.3{]}; pooled over Gold-7 and
Silver, neither is reliably better (+0.9 {[}−0.4, +2.1{]}).

Before shuffling, we also repeated the trajectory with a loss accepting
any mapped label for each gold span and masking negatives for types with
unknown coverage. The data, initialization and schedule were unchanged;
training took 11 minutes 25 seconds. This variant selected step 300 and
scored 58.74 on the first adaptation's selection criterion, versus 59.32
for the first adaptation. On all 1,201 Silver inputs, exact-boundary
redaction F1 rose from 57.47 to 59.12 (paired 95\% interval for the
gain: {[}0.01, 3.34{]} points), while exact typed F1 fell from 53.48 to
52.48. Gold-7 exact-boundary redaction F1 rose from 67.24 to 68.19, with
an interval spanning zero. GL4 keeps the higher-scoring fallback
objective.

On the 542 Silver-test segments, not used to select either checkpoint,
O4 scores 86.8 fine-type character F1 versus GL4's 75.5, and 89.3
character F1 versus 80.0. The gains are 11.3 {[}8.6, 14.0{]} and 9.4
{[}6.7, 12.2{]} points (paired document-bootstrap 95\% intervals).
Restricting to GL's seven advertised languages leaves 171 segments:
fine-type character F1 is 85.1 versus 75.2, and character F1 is 87.9
versus 79.8. These comparisons keep all primary types under the paper's
optional-reference policy, with no model-specific category exemptions,
at each system's Silver-dev operating point (O4 bias zero, GL4
confidence 0.65).

\subsubsection{Gains over published privacy
filters}\label{sec-published-filter-gains}

We measured the span-overlap F1 gains of our twenty-language Ont2
XLM\nobreak\hbox{-}\nobreak{}R encoder (Section~\ref{sec-ont2-transfer})
over five published privacy-filter configurations at fixed operating
points on Fresh20, a 1,267-document evaluation set in twenty languages
(Section~\ref{sec-evaluation-sets}). Each row scores the encoder and one
baseline only on the languages that baseline declares it supports, so
the evaluation subset differs from row to row. Span matching ignores
entity type, and gains are in percentage points.

\ACLwidetabletrue\ACLcontinuedtablefalse

\ifXeTeX
\begin{longtable}[]{@{}
  >{\raggedright\arraybackslash}p{(\linewidth - 8\tabcolsep) * \real{0.1579}}
  >{\raggedleft\arraybackslash}p{(\linewidth - 8\tabcolsep) * \real{0.2105}}
  >{\raggedleft\arraybackslash}p{(\linewidth - 8\tabcolsep) * \real{0.2105}}
  >{\raggedleft\arraybackslash}p{(\linewidth - 8\tabcolsep) * \real{0.2105}}
  >{\raggedleft\arraybackslash}p{(\linewidth - 8\tabcolsep) * \real{0.2105}}@{}}
\caption{Ont2 encoder gains in untyped span-overlap F1 at fixed
operating points on Fresh20. Each row uses only the baseline's supported
languages; differing subsets prevent ranking the baselines by these
gains. Intervals use paired document
resampling.}\label{tbl-published-filter-gains}\tabularnewline
\toprule\noalign{}
\begin{minipage}[b]{\linewidth}\raggedright
Baseline configuration
\end{minipage} & \begin{minipage}[b]{\linewidth}\raggedleft
Languages
\end{minipage} & \begin{minipage}[b]{\linewidth}\raggedleft
Documents
\end{minipage} & \begin{minipage}[b]{\linewidth}\raggedleft
Our model's F1 gain over baseline (points)
\end{minipage} & \begin{minipage}[b]{\linewidth}\raggedleft
95\% interval (points)
\end{minipage} \\
\midrule\noalign{}
\endfirsthead
\toprule\noalign{}
\begin{minipage}[b]{\linewidth}\raggedright
Baseline configuration
\end{minipage} & \begin{minipage}[b]{\linewidth}\raggedleft
Languages
\end{minipage} & \begin{minipage}[b]{\linewidth}\raggedleft
Documents
\end{minipage} & \begin{minipage}[b]{\linewidth}\raggedleft
Our model's F1 gain over baseline (points)
\end{minipage} & \begin{minipage}[b]{\linewidth}\raggedleft
95\% interval (points)
\end{minipage} \\
\midrule\noalign{}
\endhead
\bottomrule\noalign{}
\endlastfoot
GLiNER2 & 7 & 421 & +21.78 & {[}+19.14, +24.42{]} \\
OpenAI Privacy Filter & 1 & 61 & +47.31 & {[}+42.37, +52.25{]} \\
Multilingual OpenMed & 14 & 907 & +40.55 & {[}+37.63, +43.46{]} \\
OpenMed with additional task fine-tuning & 20 & 1,267 & +48.70 &
{[}+46.08, +51.31{]} \\
OpenMed Nemotron & 9 & 541 & +43.76 & {[}+40.18, +47.35{]} \\
\end{longtable}
\else
\begin{longtable}[]{@{}
  >{\raggedright\arraybackslash}X
  >{\raggedleft\arraybackslash}X
  >{\raggedleft\arraybackslash}X
  >{\raggedleft\arraybackslash}X
  >{\raggedleft\arraybackslash}X@{}}
\caption{Ont2 encoder gains in untyped span-overlap F1 at fixed
operating points on Fresh20. Each row uses only the baseline's supported
languages; differing subsets prevent ranking the baselines by these
gains. Intervals use paired document
resampling.}\label{tbl-published-filter-gains}\tabularnewline
\toprule\noalign{}
\raggedright
Baseline configuration
 & \raggedleft
Languages
 & \raggedleft
Documents
 & \raggedleft
Our model's F1 gain over baseline (points)
 & \raggedleft
95\% interval (points)
 \\
\midrule\noalign{}
\endfirsthead
\toprule\noalign{}
\raggedright
Baseline configuration
 & \raggedleft
Languages
 & \raggedleft
Documents
 & \raggedleft
Our model's F1 gain over baseline (points)
 & \raggedleft
95\% interval (points)
 \\
\midrule\noalign{}
\endhead
\bottomrule\noalign{}
\endlastfoot
GLiNER2 & 7 & 421 & +21.78 & {[}+19.14, +24.42{]} \\
OpenAI Privacy Filter & 1 & 61 & +47.31 & {[}+42.37, +52.25{]} \\
Multilingual OpenMed & 14 & 907 & +40.55 & {[}+37.63, +43.46{]} \\
OpenMed with additional task fine-tuning & 20 & 1,267 & +48.70 &
{[}+46.08, +51.31{]} \\
OpenMed Nemotron & 9 & 541 & +43.76 & {[}+40.18, +47.35{]} \\
\end{longtable}
\fi

The OpenAI Privacy Filter row covers only 61 English documents and is
exploratory. The OpenMed model with additional task fine-tuning is a
local fine-tune evaluated on all 1,267 documents.

The encoder uses its coarse--fine-aware constrained decoder with zero
outside- label bias. Scores compare fixed operating points rather than
optimized precision--recall frontiers. Documents are weighted by
language: English four, the other eight of the nine priority languages
(Section~\ref{sec-evaluation-sets}) two, and the remaining languages
one, with weights renormalized over each row's subset.

The 95\% intervals use 10,000 paired document-bootstrap resamples.
Within each row's comparison, a single-step maximum absolute
standardized bootstrap error adjusts for all reported metric cells; the
five rows are not jointly adjusted as one comparison family.

\begin{figure*}[tp]

\centering{

\ifXeTeX
\includegraphics[width=1\linewidth,height=6in,alt={Three aligned text lines with gold spans above and both models below. GLiNER2 splits the name and extends a financial span across all three lines. Gold and encoder disagree on SSN versus national ID.}]{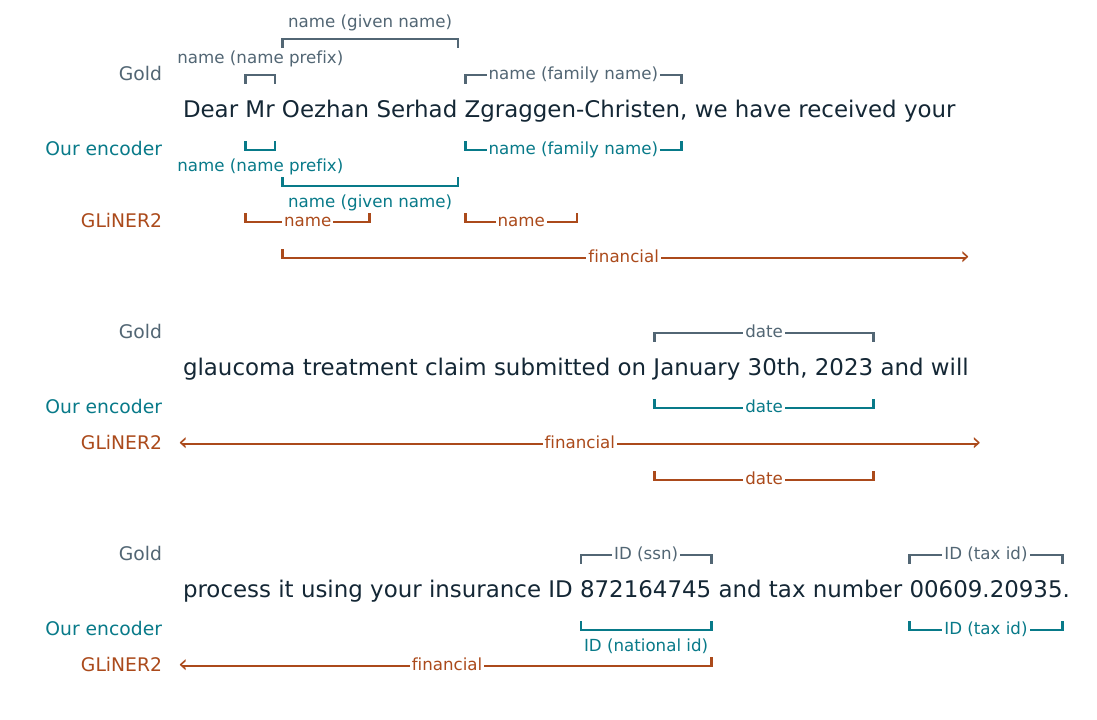}
\else
\includegraphics[alt={Three aligned text lines with gold spans above and both models below. GLiNER2 splits the name and extends a financial span across all three lines. Gold and encoder disagree on SSN versus national ID.}]{figures/span-example-en-v1.pdf}
\fi

}

\caption{\label{fig-span-example-en}\textbf{English.} GLiNER2's
financial span extends from the name through the insurance number,
including intervening prose. The encoder matches the gold boundaries;
its finer identifier type differs.}

\end{figure*}

\begin{figure*}[tp]

\centering{

\ifXeTeX
\includegraphics[width=1\linewidth,height=6in,alt={Two aligned text lines show a clinician name, street, postal code, city, country and email. Gold and our encoder label Inés Contreras; GLiNER2 labels only Contreras.}]{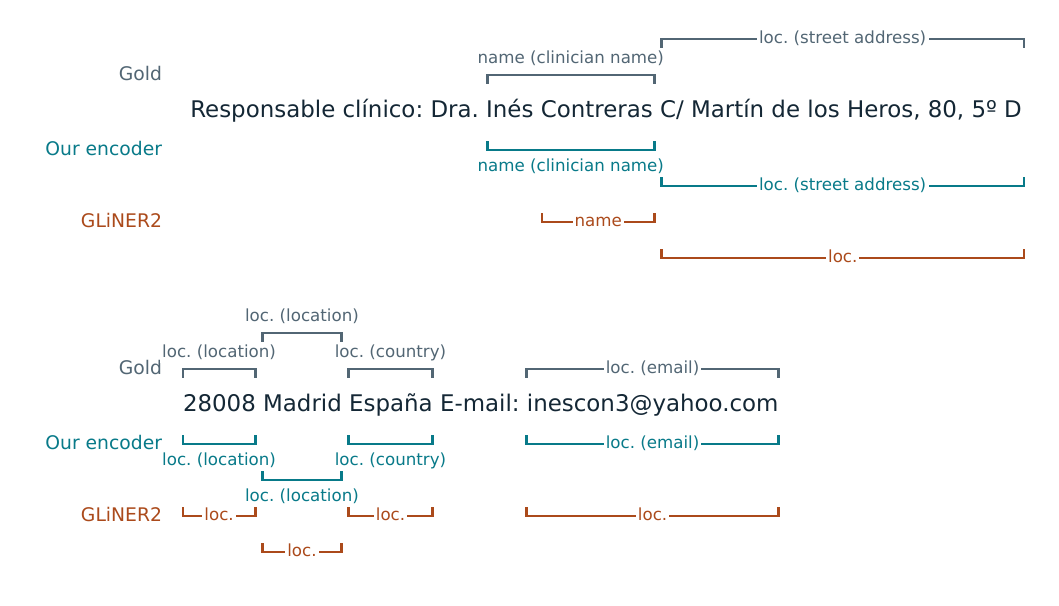}
\else
\includegraphics[alt={Two aligned text lines show a clinician name, street, postal code, city, country and email. Gold and our encoder label Inés Contreras; GLiNER2 labels only Contreras.}]{figures/span-example-es-v1.pdf}
\fi

}

\caption{\label{fig-span-example-es}\textbf{Spanish.} In this short
clinical contact block, GLiNER2 labels only the surname; the encoder
includes the given name as well. Both recover the address fields and
email.}

\end{figure*}

Figure~\ref{fig-span-example-en} and Figure~\ref{fig-span-example-es}
compare GLiNER2 and our encoder on one English and one Spanish input.
Gold brackets sit above the text and model brackets below, with the
finest labels in parentheses; GLiNER2 was prompted with broad groups
only. Arrowheads continue a span across lines. \texttt{ID} denotes
unique identifiers; \texttt{loc.} groups locations and contact details
(operating points and matching: Section~\ref{sec-span-views}).

These two selected examples come from OpenPII 1M (Ai4Privacy / Ai Suisse
SA) and MEDDOCAN (synthetic clinical corpus), respectively; both are CC
BY 4.0 \citep{ai4privacy2026-openpii, marimon2019-meddocan}. The OpenPII
example is not a member of the current Ai4Privacy-excluded benchmark.

\subsection{Operating-point calibration}\label{sec-operating-points}

After checkpoint selection, XLM\nobreak\hbox{-}\nobreak{}R caches token
offsets, the winning non-\textbf{O} label, and the
\textbf{O}-minus-winning-label logit (\textbf{O} is the BIOES outside
tag); calibration then searches a global \textbf{O} bias under three
disclosed decoders without another encoder forward pass. GLiNER2
analogously caches candidate spans and searches confidence thresholds.
The hashed operating manifest froze 11 XLM\nobreak\hbox{-}\nobreak{}R
and 32 GLiNER2 configurations before the Final20 comparison;
Section~\ref{sec-evaluation-sets} describes that twenty-language
collection and its prior use. Calibration used separate records, but not
necessarily distinct texts, and fixed all language, label and
reporting-metric settings. Neither sweep requires training updates.

The main comparison (Section~\ref{sec-priority9-comparison}) fixes each
system's operating point on Silver-dev alone. We score every grid
setting on Silver-dev and resample its source documents 10,000 times; a
setting is near-optimal when the 95\% interval of its F1 difference from
the best setting includes zero. The selected point is the median setting
of the contiguous near-optimal range around the best. Grids are quarter
steps of \textbf{O} bias from −4 to 4 plus integer steps beyond for O3,
O4 and OpenMed Privacy Filter; O2 keeps its saved integer grid, because
rerunning it did not reproduce its saved predictions exactly. GLiNER2
confidence uses steps of 0.05 and Presidio score steps of 0.025.
Relative to the single best setting, the rule leaves O4's point
unchanged and moves other systems' aggregate Gold-7 and Silver-test F1
by at most 1.2 points.

Each system reads Chinese and Japanese in the form it handles best, and
its spans are mapped back onto the unchanged scored text. Gold-7's
Chinese test text is the corpus's words joined by spaces: O2--O4,
GLiNER2 and GL4 read it as given, while Presidio (whose Chinese spaCy
pipeline segments natural text itself), the OpenMed Privacy Filter and
the frozen LLM heads (Section~\ref{sec-frozen-llm-heads}) read it with
the spaces removed, raising Gold-7 redaction F1 from 57.3 to 61.4 for
Presidio, from 34.1 to 37.0 for the Privacy Filter and from 50.9 to 56.5
for the Gemma head. On Silver's natural Chinese and Japanese, GLiNER2
and GL4 use GLiNER2's documented character-level word splitter, because
its default splitter treats a run of Chinese characters as one word;
this changes their Silver F1 by at most 1.1 points and does not change
GL4's selected checkpoint (Section~\ref{sec-gliner2-adaptation}). The
original OpenAI Privacy Filter in Appendix
Figure~\ref{fig-privacy-filter-categories} keeps the spaced form, on
which it does better.

Even with fully annotated data, reducing the loss weight on known
background tokens from 1 to 0.75 improved Ont2 span F1 in all four
matched training-seed pairs after 250 continuation steps per model. On
the 1,267-document Fresh20 set, importance-weighted language-macro
span-overlap F1 rose by a mean of 0.23 percentage points across seeds.
For the same comparisons, mean recall rose by 0.58 points and precision
fell by 0.11 points, consistent with an operating-point shift.

\subsection{Training and model
controls}\label{technique-attribution-recipes}

The following recipes specify the controls behind the main text's data,
decoder and encoder comparisons.

\subsubsection{Transition between labeling
tasks}\label{sec-ontology-transition}

To introduce the Ont2 label inventory while retaining supervision in
Ont1's label space, we kept the trained encoder and its existing affine
head (one linear map from each token state to label scores), then added
an Ont2 head. A reviewed label projection seeds the new head from the
old classifier. Both heads read the same encoder states. Examples
annotated in a head's own label space use ordinary cross-entropy;
examples from the other task use the many-to-many mapping, minimizing
\(-\log\sum_{k\in A(y)}p(k)\) over the labels compatible with annotation
\(y\). Thus a coarse annotation can supervise a finer task without
inventing a specific fine label.

We blend the separately normalized head losses as
\(L(t)=\lambda(t)L_{\mathrm{old}}+[1-\lambda(t)]L_{\mathrm{new}}\). The
evaluated transition linearly reduced \(\lambda\) from one to zero over
1,500 updates. Once it reaches zero, the old head is retired. Old-task
examples can still train the new head through the mapping. For continued
support of both labeling tasks, the same construction instead retains
both heads and nonzero loss weights. Separate output spaces preserve
task-specific distinctions while sharing encoder computation and
parameters; the desired balance is set by their supervision weights.

\subsubsection{Training mixtures}\label{sec-training-mixtures}

Table~\ref{tbl-ont2-ont3-methods} lists the methods active in O2, O3 and
O4 in the main comparison (Figure~\ref{fig-final20-priority9-overlap}).
Each column is a complete training recipe; no individual change is
ablated at matched exposure. O4 retains O3's architecture and label
inventory while changing the training data, sampling and input recipe
(Table~\ref{tbl-o4-data}; Section~\ref{sec-o4-annotation-recipe}).

\ACLwidetabletrue\ACLcontinuedtablefalse

\ifXeTeX
\begin{longtable}[]{@{}
  >{\raggedright\arraybackslash}p{(\linewidth - 6\tabcolsep) * \real{0.2100}}
  >{\raggedright\arraybackslash}p{(\linewidth - 6\tabcolsep) * \real{0.2300}}
  >{\raggedright\arraybackslash}p{(\linewidth - 6\tabcolsep) * \real{0.2500}}
  >{\raggedright\arraybackslash}p{(\linewidth - 6\tabcolsep) * \real{0.3100}}@{}}
\caption{Active methods in O2, O3 and O4. O2 uses Ont2; O3 and O4 use
Ont3. O4's additional annotations, cleanup and sampling are detailed in
Section~\ref{sec-o4-annotation-recipe}. Boundary refinement is described
in Section~\ref{sec-boundary-refiner-details}; the main curves score
primary redaction spans and ignore name-kind subspans.
}\label{tbl-ont2-ont3-methods}\tabularnewline
\toprule\noalign{}
\begin{minipage}[b]{\linewidth}\raggedright
Method
\end{minipage} & \begin{minipage}[b]{\linewidth}\raggedright
O2
\end{minipage} & \begin{minipage}[b]{\linewidth}\raggedright
O3
\end{minipage} & \begin{minipage}[b]{\linewidth}\raggedright
O4
\end{minipage} \\
\midrule\noalign{}
\endfirsthead
\toprule\noalign{}
\begin{minipage}[b]{\linewidth}\raggedright
Method
\end{minipage} & \begin{minipage}[b]{\linewidth}\raggedright
O2
\end{minipage} & \begin{minipage}[b]{\linewidth}\raggedright
O3
\end{minipage} & \begin{minipage}[b]{\linewidth}\raggedright
O4
\end{minipage} \\
\midrule\noalign{}
\endhead
\bottomrule\noalign{}
\endlastfoot
Starting encoder & Previously task-trained encoder, transferred to Ont2
& Upstream pretrained XLM\nobreak\hbox{-}\nobreak{}R-large & Ont3
encoder's isolated-input branch \\
Language coverage & 20 languages & 35 languages & 35 languages \\
Supervision & Mapped annotations, translated data and Ont2
complete-label calibration mixture & Ont3 mixture plus 40\% draws from
four mapped-gold corpora & 49,812 additional frontier-annotated texts;
earlier labels cleaned; prompts clarify occupation/job titles as
\texttt{demographic\_\allowbreak{}attribute}, outside names. 50\%
mapped-gold draws. \\
Primary label inventory & 29 types & Same 29 types plus two optional
reference types & Same as O3 \\
Additional training objectives & Binary entity-presence loss, weight 2 &
Subclass and type-conditioned predicate losses, each weight 1 & Same as
O3 \\
\textbf{O} loss weight & 1 on complete annotations; 0 on partial
annotations & 0.75 on supervised \textbf{O} tokens; corpus-unannotated
types masked & Same as O3 \\
Language-specific output bias & None & Learned bias per language added
to every output score & Same as O3 \\
Neighboring-sentence input & Target sentence only & Training: equal
isolated, previous-only and both-neighbor draws. Inference: both
neighbors, within 512 tokens & Target sentence only, in training and
inference \\
Masked language modeling & Off & Off & Off \\
Character-boundary refinement in the main curves & ±1-codepoint endpoint
ranker, 35 languages & Same ranker & Same ranker \\
\end{longtable}
\else
\begin{longtable}[]{@{}
  >{\raggedright\arraybackslash}X
  >{\raggedright\arraybackslash}X
  >{\raggedright\arraybackslash}X
  >{\raggedright\arraybackslash}X@{}}
\caption{Active methods in O2, O3 and O4. O2 uses Ont2; O3 and O4 use
Ont3. O4's additional annotations, cleanup and sampling are detailed in
Section~\ref{sec-o4-annotation-recipe}. Boundary refinement is described
in Section~\ref{sec-boundary-refiner-details}; the main curves score
primary redaction spans and ignore name-kind subspans.
}\label{tbl-ont2-ont3-methods}\tabularnewline
\toprule\noalign{}
\raggedright
Method
 & \raggedright
O2
 & \raggedright
O3
 & \raggedright
O4
 \\
\midrule\noalign{}
\endfirsthead
\toprule\noalign{}
\raggedright
Method
 & \raggedright
O2
 & \raggedright
O3
 & \raggedright
O4
 \\
\midrule\noalign{}
\endhead
\bottomrule\noalign{}
\endlastfoot
Starting encoder & Previously task-trained encoder, transferred to Ont2
& Upstream pretrained XLM\nobreak\hbox{-}\nobreak{}R-large & Ont3
encoder's isolated-input branch \\
Language coverage & 20 languages & 35 languages & 35 languages \\
Supervision & Mapped annotations, translated data and Ont2
complete-label calibration mixture & Ont3 mixture plus 40\% draws from
four mapped-gold corpora & 49,812 additional frontier-annotated texts;
earlier labels cleaned; prompts clarify occupation/job titles as
\texttt{demographic\_\allowbreak{}attribute}, outside names. 50\%
mapped-gold draws. \\
Primary label inventory & 29 types & Same 29 types plus two optional
reference types & Same as O3 \\
Additional training objectives & Binary entity-presence loss, weight 2 &
Subclass and type-conditioned predicate losses, each weight 1 & Same as
O3 \\
\textbf{O} loss weight & 1 on complete annotations; 0 on partial
annotations & 0.75 on supervised \textbf{O} tokens; corpus-unannotated
types masked & Same as O3 \\
Language-specific output bias & None & Learned bias per language added
to every output score & Same as O3 \\
Neighboring-sentence input & Target sentence only & Training: equal
isolated, previous-only and both-neighbor draws. Inference: both
neighbors, within 512 tokens & Target sentence only, in training and
inference \\
Masked language modeling & Off & Off & Off \\
Character-boundary refinement in the main curves & ±1-codepoint endpoint
ranker, 35 languages & Same ranker & Same ranker \\
\end{longtable}
\fi

The recorded samplers also differ substantially in language
concentration: the nine priority languages (weighted ×2 or ×4 in
Section~\ref{sec-evaluation-sets}) receive 75.1\% of O2's expected
sampled-example mass and 55.2\% of O3's; the seven Gold-7 languages
receive 71.7\% and 53.6\%. English alone receives 40.8\% and 19.6\%.
These are sampling probabilities for the selected stage alone. O2's
selected stage includes substantial translated supervision and MEDDOCAN,
MultiGraSCCo and TAB data. Relative to the matched Ont3 mixture without
added gold, mapped-gold replay (the 40\% draws above) increases the
seven gold languages' expected share from 37.9\% to 53.6\%; domain
exposure and language allocation therefore both change. Excluding
overlapping evaluation inputs does not remove similarities in domain,
text formatting or annotation conventions between train and test.

We also explored ordinary-density text from FineWeb and FineWeb2 to
broaden training beyond entity-dense examples. Its effect on
entity/background exposure depends on sampling and annotation coverage.
Related changes to background supervision and operating points are
described in Section~\ref{sec-operating-points}.

\subsubsection{BIOES heads and decoding}\label{bioes-heads-and-decoding}

Documents are divided into verbatim, sentence-friendly windows without
cutting a labeled character span. Tokenizer offset maps align gold spans
to overlapping subword tokens; special tokens are ignored. Single-token
spans receive S; longer spans receive B, zero or more I labels, and E,
each paired with its entity type.

\paragraph{Decoder contrast conditions}\label{sec-decoder-contrast}

\begin{figure*}[tp]

\centering{

\ifXeTeX
\includegraphics[width=1\linewidth,height=6in,alt={Aligned token and score rows for St Mary Hospital called Ana. Fine-type decoding chooses outside for the hospital name; class-aware decoding chooses beginning, inside and end of one organization. Ana is a single-token person span.}]{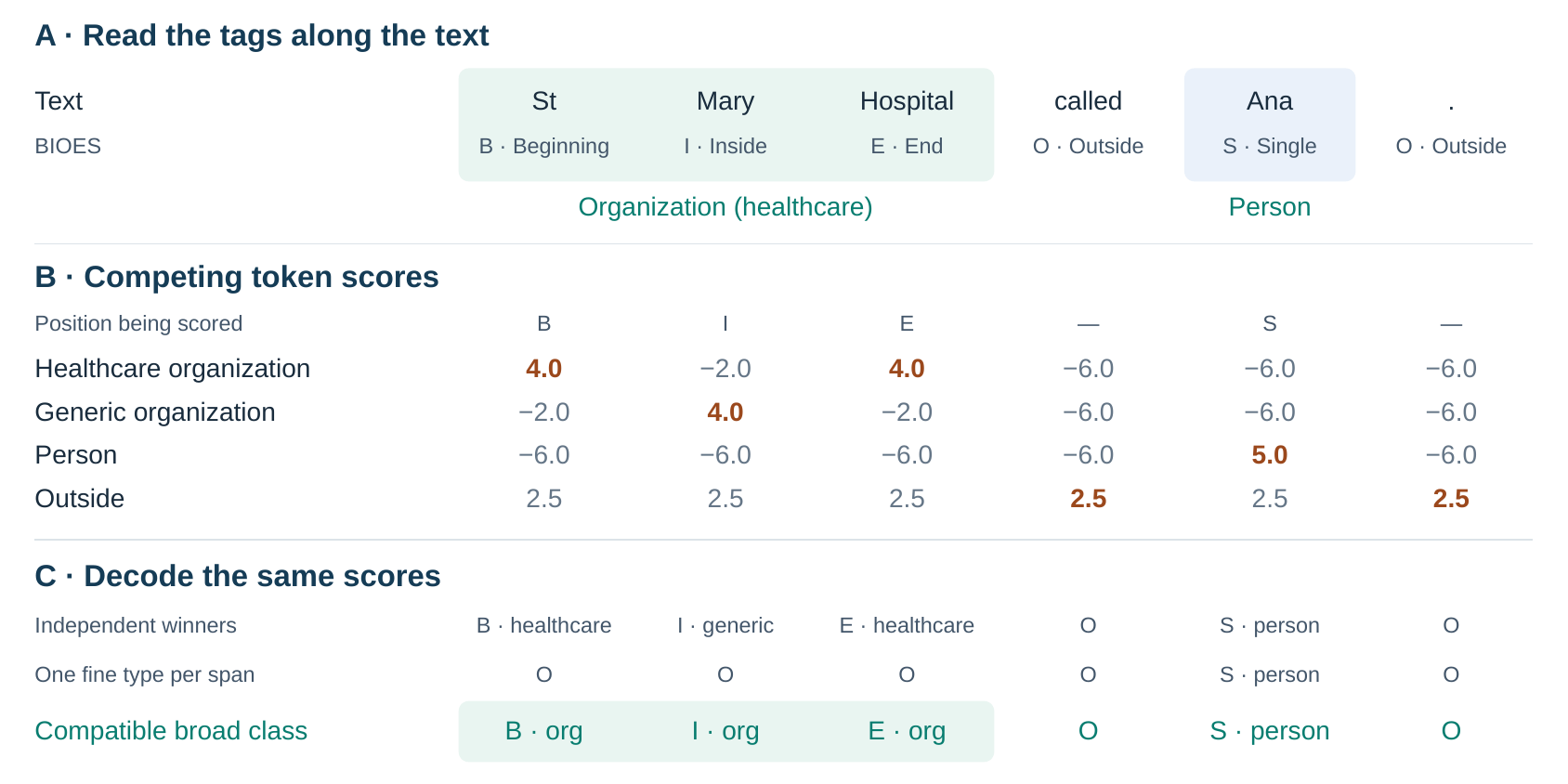}
\else
\includegraphics[alt={Aligned token and score rows for St Mary Hospital called Ana. Fine-type decoding chooses outside for the hospital name; class-aware decoding chooses beginning, inside and end of one organization. Ana is a single-token person span.}]{figures/class-aware-bioes-v1.pdf}
\fi

}

\caption{\label{fig-class-aware-bioes}BIOES labels mark a span's
beginning, interior, end or single token; O marks text outside spans. In
this constructed example, independent winners switch fine type inside
``St Mary Hospital.'' Requiring one fine type loses the name;
class-aware decoding recovers one organization span. Bold numbers are
the independent winners. Scores are invented logits, not probabilities
or measured model outputs; words stand in for tokenizer tokens.}

\end{figure*}

To compare decoders without changing encoder scores, we reuse full
logits from one XLM\nobreak\hbox{-}\nobreak{}R checkpoint (step 2,000)
trained before the Ont2 and Ont3 inventories existed. The diagnostic set
is Fresh7 (Section~\ref{sec-evaluation-sets}): 61 English, 60 each
German, Spanish, French, Japanese and Korean, and 126 Chinese documents.
All methods share logits, gold, reporting classes, token offsets, window
deduplication and a zero \textbf{O} bias, the inference-time offset
added to the outside-label score before decoding.
Figure~\ref{fig-class-aware-bioes} illustrates the class-aware decoder
on a constructed example. Matching-type Viterbi uses legal fine-type
transitions. The greedy heuristic it replaced merges compatible
continuations and retains the opening type. Both coarse-class Viterbi
variants use the 20-class compatibility map and assign the final fine
type by character-weighted token votes; only their state aggregation
differs. The soft variant uses normalized log-mean-exp at temperature
0.5.

Section~\ref{sec-transport-deployment} reports a training comparison
under the greedy deployment decoder, including its tradeoff between span
detection and fine typing.

In the example, combining four organization types at temperature 1 gives
\(\log[(e^4+e^{-2}+e^{-6}+e^{-6})/4]\approx2.62\) at each name token,
above the outside score of 2.5. The two unshown organization types and
all unshown boundary scores are −6. The class-aware path keeps the
alternating fine-type preferences within one organization span, then
labels it healthcare organization: ``St'' and ``Hospital'' contribute
ten characters of support, versus four from ``Mary'' for generic
organization.

\ACLwidetabletrue\ACLcontinuedtablefalse

\ifXeTeX
\begin{longtable}[]{@{}
  >{\raggedright\arraybackslash}p{(\linewidth - 6\tabcolsep) * \real{0.4500}}
  >{\raggedleft\arraybackslash}p{(\linewidth - 6\tabcolsep) * \real{0.2000}}
  >{\raggedleft\arraybackslash}p{(\linewidth - 6\tabcolsep) * \real{0.2000}}
  >{\raggedleft\arraybackslash}p{(\linewidth - 6\tabcolsep) * \real{0.1500}}@{}}
\caption{Decoder contrast on the 487 Fresh7 development documents in
seven languages, with the same XLM\nobreak\hbox{-}\nobreak{}R checkpoint
logits and zero \textbf{O} bias. Decode times exclude encoder
inference.}\label{tbl-decoder-contrast}\tabularnewline
\toprule\noalign{}
\begin{minipage}[b]{\linewidth}\raggedright
Decoder
\end{minipage} & \begin{minipage}[b]{\linewidth}\raggedleft
Typed exact F1 (\%)
\end{minipage} & \begin{minipage}[b]{\linewidth}\raggedleft
Typed overlap F1 (\%)
\end{minipage} & \begin{minipage}[b]{\linewidth}\raggedleft
CPU decode time (s)
\end{minipage} \\
\midrule\noalign{}
\endfirsthead
\toprule\noalign{}
\begin{minipage}[b]{\linewidth}\raggedright
Decoder
\end{minipage} & \begin{minipage}[b]{\linewidth}\raggedleft
Typed exact F1 (\%)
\end{minipage} & \begin{minipage}[b]{\linewidth}\raggedleft
Typed overlap F1 (\%)
\end{minipage} & \begin{minipage}[b]{\linewidth}\raggedleft
CPU decode time (s)
\end{minipage} \\
\midrule\noalign{}
\endhead
\bottomrule\noalign{}
\endlastfoot
Matching-type Viterbi & 73.77 & 78.65 & 2.41 \\
Coarse-compatible greedy; first label & 70.51 & 75.61 & 0.32 \\
Coarse-class Viterbi; max scores & 74.00 & 78.91 & 1.54 \\
Coarse-class Viterbi; soft scores, T = 0.5 & 74.19 & 79.18 & 2.33 \\
Score-adjusted Viterbi; α = 0.95, T = 0.5 & 74.19 & 79.17 & 2.53 \\
\end{longtable}
\else
\begin{longtable}[]{@{}
  >{\raggedright\arraybackslash}X
  >{\raggedleft\arraybackslash}X
  >{\raggedleft\arraybackslash}X
  >{\raggedleft\arraybackslash}X@{}}
\caption{Decoder contrast on the 487 Fresh7 development documents in
seven languages, with the same XLM\nobreak\hbox{-}\nobreak{}R checkpoint
logits and zero \textbf{O} bias. Decode times exclude encoder
inference.}\label{tbl-decoder-contrast}\tabularnewline
\toprule\noalign{}
\raggedright
Decoder
 & \raggedleft
Typed exact F1 (\%)
 & \raggedleft
Typed overlap F1 (\%)
 & \raggedleft
CPU decode time (s)
 \\
\midrule\noalign{}
\endfirsthead
\toprule\noalign{}
\raggedright
Decoder
 & \raggedleft
Typed exact F1 (\%)
 & \raggedleft
Typed overlap F1 (\%)
 & \raggedleft
CPU decode time (s)
 \\
\midrule\noalign{}
\endhead
\bottomrule\noalign{}
\endlastfoot
Matching-type Viterbi & 73.77 & 78.65 & 2.41 \\
Coarse-compatible greedy; first label & 70.51 & 75.61 & 0.32 \\
Coarse-class Viterbi; max scores & 74.00 & 78.91 & 1.54 \\
Coarse-class Viterbi; soft scores, T = 0.5 & 74.19 & 79.18 & 2.33 \\
Score-adjusted Viterbi; α = 0.95, T = 0.5 & 74.19 & 79.17 & 2.53 \\
\end{longtable}
\fi

Table~\ref{tbl-decoder-contrast} reuses development data. Typed spans
must satisfy the 80\% mutual-coverage rule. CPU timings are single-pass
observations; a corresponding Ont3 contrast has not been measured.
Temperature 0.5 had been inspected on these documents before this run,
which adds no tuning and no paired intervals.

\begin{figure*}[tp]

\centering{

\ifXeTeX
\includegraphics[width=1\linewidth,height=6in,alt={Two curves show exact and overlap typed F1 increasing toward the coarse-class reference as sibling interpolation approaches one.}]{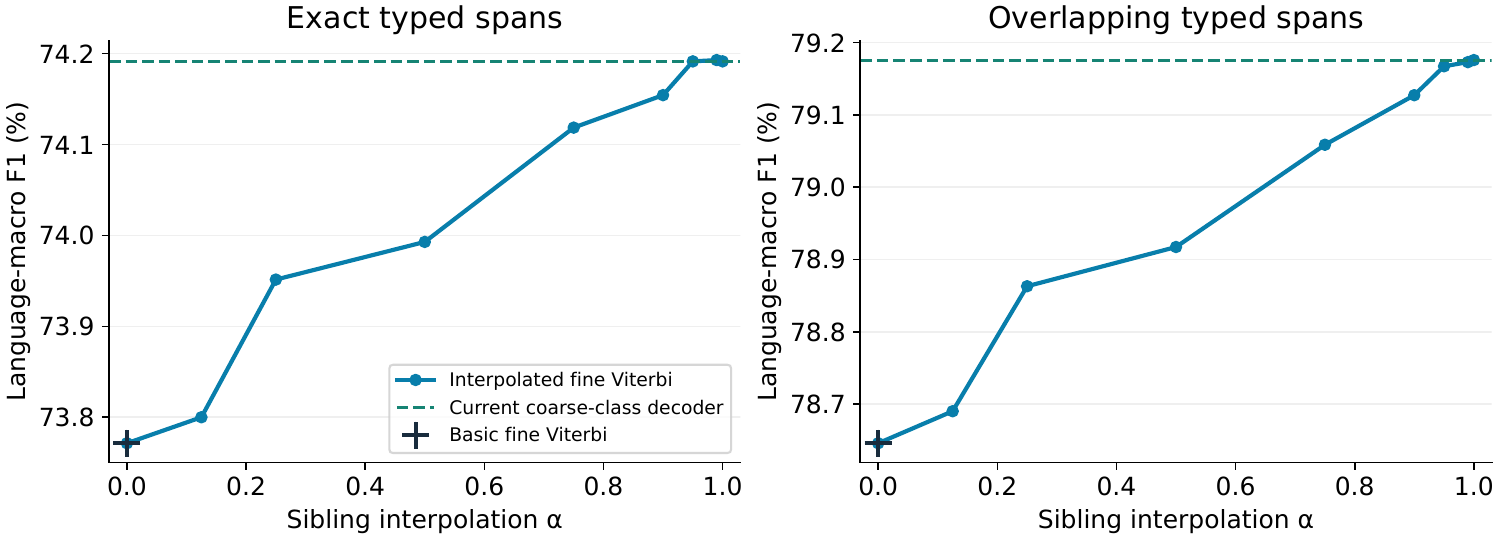}
\else
\includegraphics[alt={Two curves show exact and overlap typed F1 increasing toward the coarse-class reference as sibling interpolation approaches one.}]{figures/class-aware-bioes-backoff-v1.pdf}
\fi

}

\caption{\label{fig-bioes-backoff}Exact fine-type Viterbi approaches the
soft coarse-class decoder as its scores back off toward the normalized
sibling aggregate. Points are the nine tested interpolation weights;
dashed lines show the coarse-class decoder. F1 uses 20 shared reporting
classes on the same 487 reused development documents. Fine-label quality
is not measured by these curves.}

\end{figure*}

An alternative adjusts the label scores before ordinary Viterbi decoding
(Figure~\ref{fig-bioes-backoff}). At each token, we combine scores for
related labels---for example, beginning of organization and beginning of
healthcare organization. We move each label's score toward that shared
score by a fraction \(\alpha\):

\[
z'_{t,b,f}=(1-\alpha)z_{t,b,f}+\alpha G_{t,b,C(f)},
\qquad 0\leq\alpha\leq1.
\]

Here \(C(f)\) is the compatible fine-type group containing \(f\), and
\(G\) is that group's aggregate score. We repeat this separately for
inside, end, and single-token labels; outside scores stay unchanged.
Viterbi then selects the highest-scoring valid sequence, requiring the
same fine type throughout each span. On the same Fresh7 documents, 95\%
interpolation scores 79.17 typed overlap F1, versus 79.18 for the soft
coarse-class decoder (Table~\ref{tbl-decoder-contrast}).

The group contains fine labels with the same BIOES position and
compatible coarse class; \textbf{O} is a singleton and remains
unchanged. We fix \(T=0.5\), zero \textbf{O} bias, and all preceding
evaluation conditions. The tested weights are 0, .125, .25, .5, .75, .9,
.95, .99 and 1. Dividing by group size removes the explicit
subclass-count bonus. The convex combination preserves invariance to a
common additive shift in a token's logits.

At zero backoff, predictions exactly reproduce fine-type Viterbi. At
complete backoff, every member receives its group's score, so a
fine-state lattice optimizes the same coarse boundary objective: all 487
documents then have the coarse-class decoder's spans and coarse labels,
and fine types within a class tie. Interior weights retain fine-score
preferences. For fixed boundaries and any weight below one, the shared
coarse term cancels when comparing sibling fine types, so their ranking
follows summed original logits, whereas the coarse-class decoder uses
character-weighted votes over token winners. At weight .95, the two
choose different fine labels on 14 of 4,359 spans with identical
boundaries and coarse class (0.32\%). For example, ``Klinikum Sankt
Marien'' receives generic organization under the fine-compatible path
and healthcare organization under the vote. At complete backoff, 2,769
of 4,362 fine labels differ because the fine-score preference has been
removed.

\paragraph{Head size and alternatives}\label{sec-head-alternatives}

For the XLM\nobreak\hbox{-}\nobreak{}R-large O1 model, the head is
dropout followed by one affine map from the final 1,024-dimensional
token state to 349 logits: \textbf{O} plus four BIOES boundary labels
for each of 87 entity types. The head has 357,725 trainable parameters.
To compare alternative heads, we fixed the encoder, a 192,000-window
training subset, the validation procedure, example order, and seed.
After correcting the task-head initialization seed, we tested seven
alternatives: four layer concatenations, one nonlinear head and two
rank-constrained affine heads. Peak validation F1 was:

\ACLwidetabletrue\ACLcontinuedtablefalse

\ifXeTeX
\begin{longtable}[]{@{}
  >{\raggedright\arraybackslash}p{(\linewidth - 8\tabcolsep) * \real{0.2900}}
  >{\raggedleft\arraybackslash}p{(\linewidth - 8\tabcolsep) * \real{0.1200}}
  >{\raggedleft\arraybackslash}p{(\linewidth - 8\tabcolsep) * \real{0.1600}}
  >{\raggedleft\arraybackslash}p{(\linewidth - 8\tabcolsep) * \real{0.2400}}
  >{\raggedright\arraybackslash}p{(\linewidth - 8\tabcolsep) * \real{0.1900}}@{}}
\caption{XLM\nobreak\hbox{-}\nobreak{}R head and layer-feature screens
on a fixed 192,000-window training subset. Entries report peak
validation F1, except where an early stopping step is shown. The initial
middle-layer gain did not repeat.
}\label{tbl-head-alternatives}\tabularnewline
\toprule\noalign{}
\begin{minipage}[b]{\linewidth}\raggedright
Token features and head
\end{minipage} & \begin{minipage}[b]{\linewidth}\raggedleft
Head params
\end{minipage} & \begin{minipage}[b]{\linewidth}\raggedleft
Initial screen F1 (\%)
\end{minipage} & \begin{minipage}[b]{\linewidth}\raggedleft
Repeat screen F1 (\%)
\end{minipage} & \begin{minipage}[b]{\linewidth}\raggedright
Interpretation
\end{minipage} \\
\midrule\noalign{}
\endfirsthead
\toprule\noalign{}
\begin{minipage}[b]{\linewidth}\raggedright
Token features and head
\end{minipage} & \begin{minipage}[b]{\linewidth}\raggedleft
Head params
\end{minipage} & \begin{minipage}[b]{\linewidth}\raggedleft
Initial screen F1 (\%)
\end{minipage} & \begin{minipage}[b]{\linewidth}\raggedleft
Repeat screen F1 (\%)
\end{minipage} & \begin{minipage}[b]{\linewidth}\raggedright
Interpretation
\end{minipage} \\
\midrule\noalign{}
\endhead
\bottomrule\noalign{}
\endlastfoot
final layer {[}24{]}, affine & 357,725 & 82.20 & 82.14 & stable
reference \\
middle + final {[}12,24{]}, affine & 715,101 & 83.82 & 81.86 & initial
gain does not repeat \\
layers {[}23,24{]}, affine & 715,101 & --- & 73.73 step 1,250 &
early-screen loss \\
layers {[}8,24{]}, affine & 715,101 & --- & 71.67 step 1,250 &
early-screen loss \\
embedding + final {[}0,24{]}, affine & 715,101 & --- & 69.16 step 1,250
& early-screen loss \\
final layer, 1,024-unit GELU & 1,409,373 & 69.21 step 1,250 & --- &
early-screen loss \\
\end{longtable}
\else
\begin{longtable}[]{@{}
  >{\raggedright\arraybackslash}X
  >{\raggedleft\arraybackslash}X
  >{\raggedleft\arraybackslash}X
  >{\raggedleft\arraybackslash}X
  >{\raggedright\arraybackslash}X@{}}
\caption{XLM\nobreak\hbox{-}\nobreak{}R head and layer-feature screens
on a fixed 192,000-window training subset. Entries report peak
validation F1, except where an early stopping step is shown. The initial
middle-layer gain did not repeat.
}\label{tbl-head-alternatives}\tabularnewline
\toprule\noalign{}
\raggedright
Token features and head
 & \raggedleft
Head params
 & \raggedleft
Initial screen F1 (\%)
 & \raggedleft
Repeat screen F1 (\%)
 & \raggedright
Interpretation
 \\
\midrule\noalign{}
\endfirsthead
\toprule\noalign{}
\raggedright
Token features and head
 & \raggedleft
Head params
 & \raggedleft
Initial screen F1 (\%)
 & \raggedleft
Repeat screen F1 (\%)
 & \raggedright
Interpretation
 \\
\midrule\noalign{}
\endhead
\bottomrule\noalign{}
\endlastfoot
final layer {[}24{]}, affine & 357,725 & 82.20 & 82.14 & stable
reference \\
middle + final {[}12,24{]}, affine & 715,101 & 83.82 & 81.86 & initial
gain does not repeat \\
layers {[}23,24{]}, affine & 715,101 & --- & 73.73 step 1,250 &
early-screen loss \\
layers {[}8,24{]}, affine & 715,101 & --- & 71.67 step 1,250 &
early-screen loss \\
embedding + final {[}0,24{]}, affine & 715,101 & --- & 69.16 step 1,250
& early-screen loss \\
final layer, 1,024-unit GELU & 1,409,373 & 69.21 step 1,250 & --- &
early-screen loss \\
\end{longtable}
\fi

Five nonlinear configurations screened before the seed correction are
excluded from this count and support only coarse pruning. Some corrected
alternatives were stopped early, as indicated by their reported steps.

Concatenating middle and final layers initially gained 1.6 F1 points,
then lost 0.3 points on repetition. It also peaked below final-only
features on a separate native-gold screen (74.9 versus 76.1 F1).
Rank-constrained affine heads reached 65.4 and 57.0 F1 and were dropped
from this search.

A second frozen-feature test concatenated the final token states of
XLM\nobreak\hbox{-}\nobreak{}R and mmBERT, aligned by character
position, and trained only an affine head. On 480 validation rows after
fitting 1,280 training rows, both seeds lost fine-label F1 relative to
XLM\nobreak\hbox{-}\nobreak{}R alone. The recorded differences are
diagnostic point estimates.

We also tested sequence-model alternatives in less controlled
comparisons. An mmBERT package replacing its stock nonlinear head with
affine emissions and a first-order BIOES conditional random field fell
from 94.4/72.2 to 92.0/54.5 F1 on Fresh6/MultiGraSCCo (a six-language
diagnostic suite and a synthetic clinical set;
Section~\ref{sec-evaluation-sets}) and from 36.9 to 0.96 documents per
second. A zero-parameter legal-BIOES projection improved
familiar-distribution token diagnostics but reduced frozen Fresh6
overlap F1 by 1.9 points (interval {[}−2.38, −1.43{]} points). We
therefore retained the final-state affine head. The full search record
is in the
\href{../../frontier.md\#classifier-head-and-encoder-depth-contrast}{classifier-head
and encoder-depth contrast} and
\href{../../frontier.md\#what-the-tested-mmbert-crf-wasand-why-it-is-dropped}{CRF
analysis}.

\subsubsection{Pretrained-encoder selection}\label{sec-encoder-screen}

To compare encoder families on common training data, we use only 1,854
MEDDOCAN and TAB training documents and select checkpoints without using
results on the evaluation sets, the six-language Fresh6 diagnostic suite
and synthetic clinical MultiGraSCCo (Section~\ref{sec-evaluation-sets}).
XLM\nobreak\hbox{-}\nobreak{}R-large and LaBSE each use a separately
trained final-state affine head. mmBERT uses its stock two-affine
GELU/LayerNorm head, so its row changes the encoder and head together.

\ACLwidetabletrue\ACLcontinuedtablefalse

\ifXeTeX
\begin{longtable}[]{@{}
  >{\raggedright\arraybackslash}p{(\linewidth - 6\tabcolsep) * \real{0.2143}}
  >{\raggedleft\arraybackslash}p{(\linewidth - 6\tabcolsep) * \real{0.2857}}
  >{\raggedleft\arraybackslash}p{(\linewidth - 6\tabcolsep) * \real{0.2857}}
  >{\raggedright\arraybackslash}p{(\linewidth - 6\tabcolsep) * \real{0.2143}}@{}}
\caption{Encoder systems trained on the same 1,854 MEDDOCAN/TAB
documents and evaluated on Fresh6 and MultiGraSCCo.
XLM\nobreak\hbox{-}\nobreak{}R and LaBSE each use an affine head; mmBERT
uses its stock nonlinear head. Brackets give 95\% intervals.
}\label{tbl-encoder-screen}\tabularnewline
\toprule\noalign{}
\begin{minipage}[b]{\linewidth}\raggedright
Encoder and task head
\end{minipage} & \begin{minipage}[b]{\linewidth}\raggedleft
Fresh6 F1 (\%) {[}95\% interval{]}
\end{minipage} & \begin{minipage}[b]{\linewidth}\raggedleft
MultiGraSCCo F1 (\%) {[}95\% interval{]}
\end{minipage} & \begin{minipage}[b]{\linewidth}\raggedright
Disposition
\end{minipage} \\
\midrule\noalign{}
\endfirsthead
\toprule\noalign{}
\begin{minipage}[b]{\linewidth}\raggedright
Encoder and task head
\end{minipage} & \begin{minipage}[b]{\linewidth}\raggedleft
Fresh6 F1 (\%) {[}95\% interval{]}
\end{minipage} & \begin{minipage}[b]{\linewidth}\raggedleft
MultiGraSCCo F1 (\%) {[}95\% interval{]}
\end{minipage} & \begin{minipage}[b]{\linewidth}\raggedright
Disposition
\end{minipage} \\
\midrule\noalign{}
\endhead
\bottomrule\noalign{}
\endlastfoot
XLM\nobreak\hbox{-}\nobreak{}R-large, final affine & 92.23 {[}91.41,
92.98{]} & 75.07 {[}74.27, 75.90{]} & quality base selected \\
LaBSE, final affine & 87.88 {[}86.68, 88.94{]} & 72.53 {[}71.69,
73.38{]} & common-head comparison \\
mmBERT-base, stock head & 89.25 {[}88.27, 90.17{]} & 70.61 {[}69.77,
71.49{]} & faster challenger; different head \\
\end{longtable}
\else
\begin{longtable}[]{@{}
  >{\raggedright\arraybackslash}X
  >{\raggedleft\arraybackslash}X
  >{\raggedleft\arraybackslash}X
  >{\raggedright\arraybackslash}X@{}}
\caption{Encoder systems trained on the same 1,854 MEDDOCAN/TAB
documents and evaluated on Fresh6 and MultiGraSCCo.
XLM\nobreak\hbox{-}\nobreak{}R and LaBSE each use an affine head; mmBERT
uses its stock nonlinear head. Brackets give 95\% intervals.
}\label{tbl-encoder-screen}\tabularnewline
\toprule\noalign{}
\raggedright
Encoder and task head
 & \raggedleft
Fresh6 F1 (\%) {[}95\% interval{]}
 & \raggedleft
MultiGraSCCo F1 (\%) {[}95\% interval{]}
 & \raggedright
Disposition
 \\
\midrule\noalign{}
\endfirsthead
\toprule\noalign{}
\raggedright
Encoder and task head
 & \raggedleft
Fresh6 F1 (\%) {[}95\% interval{]}
 & \raggedleft
MultiGraSCCo F1 (\%) {[}95\% interval{]}
 & \raggedright
Disposition
 \\
\midrule\noalign{}
\endhead
\bottomrule\noalign{}
\endlastfoot
XLM\nobreak\hbox{-}\nobreak{}R-large, final affine & 92.23 {[}91.41,
92.98{]} & 75.07 {[}74.27, 75.90{]} & quality base selected \\
LaBSE, final affine & 87.88 {[}86.68, 88.94{]} & 72.53 {[}71.69,
73.38{]} & common-head comparison \\
mmBERT-base, stock head & 89.25 {[}88.27, 90.17{]} & 70.61 {[}69.77,
71.49{]} & faster challenger; different head \\
\end{longtable}
\fi

With independently trained affine heads, XLM\nobreak\hbox{-}\nobreak{}R
exceeds LaBSE by 4.4 Fresh6 F1 points and 2.5 MultiGraSCCo points. Other
encoder-family comparisons used different training mixtures; this table
is the common-data reference.

Using O4's full training data, we also compare mmBERT's stock 768-unit
GELU/LayerNorm head with an affine head, keeping the encoder, fresh
classifier initialization, dropout 0.1, sampling and learning rates
matched. Both use O4's boundary adjustment, name handling and regex
supplementation at evaluation. Checkpoints are selected on the same 279
development examples. The affine and stock-head fits stop after 7,500
and 4,000 updates, taking 69 and 43 minutes on a 96-GB GPU; their
selected checkpoints are at 5,500 and 2,000 updates.

\ACLwidetabletrue\ACLcontinuedtablefalse

\ifXeTeX
\begin{longtable}[]{@{}lrrr@{}}
\caption{Exact-boundary F1 (\%): redaction regions, and typed spans in
the last column, on 1,283 Gold-7 examples and 1,201 Silver development
and held-out examples. The mmBERT head comparison uses O4's full
training data and serving policy; O4 retains its own training lineage.
}\label{tbl-mmbert-o4}\tabularnewline
\toprule\noalign{}
Model and head & Gold-7 regions & Silver regions & Silver typed \\
\midrule\noalign{}
\endfirsthead
\toprule\noalign{}
Model and head & Gold-7 regions & Silver regions & Silver typed \\
\midrule\noalign{}
\endhead
\bottomrule\noalign{}
\endlastfoot
mmBERT-base, stock GELU & 84.89 & 76.86 & 72.32 \\
mmBERT-base, affine & 87.03 & 77.02 & 73.90 \\
XLM\nobreak\hbox{-}\nobreak{}R-large O4, affine & 87.80 & 80.20 &
76.30 \\
\end{longtable}
\else
\begin{longtable}[]{@{}lrrr@{}}
\caption{Exact-boundary F1 (\%): redaction regions, and typed spans in
the last column, on 1,283 Gold-7 examples and 1,201 Silver development
and held-out examples. The mmBERT head comparison uses O4's full
training data and serving policy; O4 retains its own training lineage.
}\label{tbl-mmbert-o4}\tabularnewline
\toprule\noalign{}
Model and head & Gold-7 regions & Silver regions & Silver typed \\
\midrule\noalign{}
\endfirsthead
\toprule\noalign{}
Model and head & Gold-7 regions & Silver regions & Silver typed \\
\midrule\noalign{}
\endhead
\bottomrule\noalign{}
\endlastfoot
mmBERT-base, stock GELU & 84.89 & 76.86 & 72.32 \\
mmBERT-base, affine & 87.03 & 77.02 & 73.90 \\
XLM\nobreak\hbox{-}\nobreak{}R-large O4, affine & 87.80 & 80.20 &
76.30 \\
\end{longtable}
\fi

The affine head gains 2.14 Gold-7 F1 points (paired document-bootstrap
95\% interval {[}0.95, 3.38{]}) and 1.59 Silver typed points {[}0.05,
3.08{]} over the stock head. Its Silver region change, +0.16 {[}−1.36,
+1.68{]}, is unresolved. It remains 2.40 typed points below O4 {[}−3.98,
−0.80{]}.

\subsubsection{Locale-aware names and dates}\label{sec-locale-repairs}

Our augmentation pipeline translated text containing typed entity
placeholders, then filled them with generated values, including names
from the \href{https://faker.readthedocs.io/}{Faker library}. For
\texttt{{[}GIVEN\_\allowbreak{}NAME{]}\ {[}FAMILY\_\allowbreak{}NAME{]}},
the filler drew from separate given-name and surname generators and
assigned each span its placeholder's label. Translation often preserved
this given-then-family order in Korean. The encoder trained on these
examples subsequently classified 65 of 79 Korean evaluation surnames as
given names. Only 3.7\% of training names were family-first, compared
with 87.8\% of evaluation names. Dates showed a similar mismatch: 85\%
of evaluation dates used the native \texttt{YYYY년\ M월\ D일} form,
which was absent from training. These mismatches motivated repairs to
name order and date formatting.

Related work generates PII training data for GLiNER2-PII
\citep{zaratiana2026-gliner2-pii} and multilingual evaluation data for
REDACT and DialogPII \citep{vats2026-redact, roller2026-dialogpii}.

For affected Chinese, Japanese and Korean examples derived from English,
we placed family names before given names, preserving each component's
label. We also rendered Korean dates in the native form while retaining
one-eighth in other formats.

Across three random seeds, matched training runs showed that name-order
repair improved span-overlap F1 averaged across languages by 1.4--1.9
points, with the largest gains in Japanese. Date-format repair improved
Korean span-overlap F1 by 4.4--7.5 points. All six paired 95\%
confidence intervals were entirely positive, on the 487 Fresh7
development documents in seven languages
(Section~\ref{sec-evaluation-sets}). The combined locale repairs
improved span-overlap F1 on the twenty-language Fresh20 set by 1.5
points and fine-label F1 by 1.7 points, both with entirely positive 95\%
intervals.

The name-order treatment changes only the realized order of separate
given-name and family-name values in audited source families whose
carrier text was English; the labels stay attached to their original
values. The Korean-date treatment changes parseable Korean date surfaces
to native \texttt{YYYY년\ M월\ D일} while a deterministic one-eighth
remains non-native. Each seed's treatment and control share the starting
checkpoint, learning rate, schedule, step count and all other data
composition. Each interval uses 10,000 paired document resamples.

\ACLwidetabletrue\ACLcontinuedtablefalse

\ifXeTeX
\begin{longtable}[]{@{}
  >{\raggedleft\arraybackslash}p{(\linewidth - 4\tabcolsep) * \real{0.3333}}
  >{\raggedleft\arraybackslash}p{(\linewidth - 4\tabcolsep) * \real{0.3333}}
  >{\raggedleft\arraybackslash}p{(\linewidth - 4\tabcolsep) * \real{0.3333}}@{}}
\caption{Name-order repair versus its matched control on 487 Fresh7
documents. Deltas are language-macro overlap-F1 points, with paired 95\%
intervals. }\label{tbl-name-order-repair}\tabularnewline
\toprule\noalign{}
\begin{minipage}[b]{\linewidth}\raggedleft
Seed
\end{minipage} & \begin{minipage}[b]{\linewidth}\raggedleft
Name repair: macro span Δ (points) {[}95\% interval{]}
\end{minipage} & \begin{minipage}[b]{\linewidth}\raggedleft
Name repair: macro fine Δ (points) {[}95\% interval{]}
\end{minipage} \\
\midrule\noalign{}
\endfirsthead
\toprule\noalign{}
\begin{minipage}[b]{\linewidth}\raggedleft
Seed
\end{minipage} & \begin{minipage}[b]{\linewidth}\raggedleft
Name repair: macro span Δ (points) {[}95\% interval{]}
\end{minipage} & \begin{minipage}[b]{\linewidth}\raggedleft
Name repair: macro fine Δ (points) {[}95\% interval{]}
\end{minipage} \\
\midrule\noalign{}
\endhead
\bottomrule\noalign{}
\endlastfoot
154 & +1.57 {[}+0.79, +2.38{]} & +2.47 {[}+1.78, +3.20{]} \\
155 & +1.44 {[}+0.64, +2.23{]} & +1.96 {[}+1.30, +2.62{]} \\
156 & +1.88 {[}+0.92, +2.84{]} & +2.64 {[}+1.86, +3.42{]} \\
\end{longtable}
\else
\begin{longtable}[]{@{}
  >{\raggedleft\arraybackslash}X
  >{\raggedleft\arraybackslash}X
  >{\raggedleft\arraybackslash}X@{}}
\caption{Name-order repair versus its matched control on 487 Fresh7
documents. Deltas are language-macro overlap-F1 points, with paired 95\%
intervals. }\label{tbl-name-order-repair}\tabularnewline
\toprule\noalign{}
\raggedleft
Seed
 & \raggedleft
Name repair: macro span Δ (points) {[}95\% interval{]}
 & \raggedleft
Name repair: macro fine Δ (points) {[}95\% interval{]}
 \\
\midrule\noalign{}
\endfirsthead
\toprule\noalign{}
\raggedleft
Seed
 & \raggedleft
Name repair: macro span Δ (points) {[}95\% interval{]}
 & \raggedleft
Name repair: macro fine Δ (points) {[}95\% interval{]}
 \\
\midrule\noalign{}
\endhead
\bottomrule\noalign{}
\endlastfoot
154 & +1.57 {[}+0.79, +2.38{]} & +2.47 {[}+1.78, +3.20{]} \\
155 & +1.44 {[}+0.64, +2.23{]} & +1.96 {[}+1.30, +2.62{]} \\
156 & +1.88 {[}+0.92, +2.84{]} & +2.64 {[}+1.86, +3.42{]} \\
\end{longtable}
\fi

\ACLwidetabletrue\ACLcontinuedtablefalse

\ifXeTeX
\begin{longtable}[]{@{}
  >{\raggedleft\arraybackslash}p{(\linewidth - 4\tabcolsep) * \real{0.3333}}
  >{\raggedleft\arraybackslash}p{(\linewidth - 4\tabcolsep) * \real{0.3333}}
  >{\raggedleft\arraybackslash}p{(\linewidth - 4\tabcolsep) * \real{0.3333}}@{}}
\caption{Korean date-format repair versus its matched control on Fresh7.
Deltas are span-overlap-F1 points for Korean and the seven-language
macro average, with paired 95\%
intervals.}\label{tbl-date-format-repair}\tabularnewline
\toprule\noalign{}
\begin{minipage}[b]{\linewidth}\raggedleft
Seed
\end{minipage} & \begin{minipage}[b]{\linewidth}\raggedleft
Date repair: Korean span Δ (points) {[}95\% interval{]}
\end{minipage} & \begin{minipage}[b]{\linewidth}\raggedleft
Date repair: macro span Δ (points) {[}95\% interval{]}
\end{minipage} \\
\midrule\noalign{}
\endfirsthead
\toprule\noalign{}
\begin{minipage}[b]{\linewidth}\raggedleft
Seed
\end{minipage} & \begin{minipage}[b]{\linewidth}\raggedleft
Date repair: Korean span Δ (points) {[}95\% interval{]}
\end{minipage} & \begin{minipage}[b]{\linewidth}\raggedleft
Date repair: macro span Δ (points) {[}95\% interval{]}
\end{minipage} \\
\midrule\noalign{}
\endhead
\bottomrule\noalign{}
\endlastfoot
154 & +5.08 {[}+2.64, +7.63{]} & +1.62 {[}+0.92, +2.33{]} \\
155 & +7.53 {[}+4.61, +10.63{]} & +2.12 {[}+1.29, +2.96{]} \\
156 & +4.38 {[}+1.06, +7.58{]} & +1.16 {[}+0.27, +2.04{]} \\
\end{longtable}
\else
\begin{longtable}[]{@{}
  >{\raggedleft\arraybackslash}X
  >{\raggedleft\arraybackslash}X
  >{\raggedleft\arraybackslash}X@{}}
\caption{Korean date-format repair versus its matched control on Fresh7.
Deltas are span-overlap-F1 points for Korean and the seven-language
macro average, with paired 95\%
intervals.}\label{tbl-date-format-repair}\tabularnewline
\toprule\noalign{}
\raggedleft
Seed
 & \raggedleft
Date repair: Korean span Δ (points) {[}95\% interval{]}
 & \raggedleft
Date repair: macro span Δ (points) {[}95\% interval{]}
 \\
\midrule\noalign{}
\endfirsthead
\toprule\noalign{}
\raggedleft
Seed
 & \raggedleft
Date repair: Korean span Δ (points) {[}95\% interval{]}
 & \raggedleft
Date repair: macro span Δ (points) {[}95\% interval{]}
 \\
\midrule\noalign{}
\endhead
\bottomrule\noalign{}
\endlastfoot
154 & +5.08 {[}+2.64, +7.63{]} & +1.62 {[}+0.92, +2.33{]} \\
155 & +7.53 {[}+4.61, +10.63{]} & +2.12 {[}+1.29, +2.96{]} \\
156 & +4.38 {[}+1.06, +7.58{]} & +1.16 {[}+0.27, +2.04{]} \\
\end{longtable}
\fi

Japanese supplies the largest and most stable name-order effect;
Korean's name-only response varies by seed. The date repair is
Korean-positive in all three seeds. These matched contrasts credit the
two convention repairs individually. The release-level comparison
credits the wider locale bundle: the release trained with it, minus the
preceding sibling release, changes Fresh20 span and fine-label overlap
F1 by 1.5 points (interval {[}0.36, 2.63{]}) and 1.7 points (interval
{[}0.62, 2.81{]}), with no supported language loss.

\subsubsection{Routed translated-data
transport}\label{sec-transport-controls}

To improve translated training data, we used TranslateGemma-27B to
translate text containing entity placeholders. Checks on the
placeholders and output structure determined whether to accept a
translation, retry it with Gemma\nobreak\hbox{-}\nobreak{}4-31B, or
reject it.

An entity placeholder replaces a labeled value with its type and a
stable identifier:
\texttt{{[}GIVEN\_\allowbreak{}NAME\_\allowbreak{}1{]}} stands for one
given-name slot. The translation can move that slot while its identifier
retains the source label. Filling the translated slot with a replacement
name gives both the new text and the location of its label.

\ACLwidetabletrue\ACLcontinuedtablefalse

\ifXeTeX
\begin{longtable}[]{@{}ll@{}}
\caption{Placeholder translation preserves the given-name annotation
through its stable slot identifier. The translated marker is
subsequently filled with an entity
value.}\label{tbl-placeholder-translation}\tabularnewline
\toprule\noalign{}
Step & Text \\
\midrule\noalign{}
\endfirsthead
\toprule\noalign{}
Step & Text \\
\midrule\noalign{}
\endhead
\bottomrule\noalign{}
\endlastfoot
Source field & \texttt{First\ Name:\ Kevin} \\
Replace the name with a placeholder &
\texttt{First\ Name:\ {[}GIVEN\_\allowbreak{}NAME\_\allowbreak{}1{]}} \\
TranslateGemma output &
\texttt{Nombre:\ {[}GIVEN\_\allowbreak{}NAME\_\allowbreak{}1{]}} \\
\end{longtable}
\else
\begin{longtable}[]{@{}ll@{}}
\caption{Placeholder translation preserves the given-name annotation
through its stable slot identifier. The translated marker is
subsequently filled with an entity
value.}\label{tbl-placeholder-translation}\tabularnewline
\toprule\noalign{}
Step & Text \\
\midrule\noalign{}
\endfirsthead
\toprule\noalign{}
Step & Text \\
\midrule\noalign{}
\endhead
\bottomrule\noalign{}
\endlastfoot
Source field & \texttt{First\ Name:\ Kevin} \\
Replace the name with a placeholder &
\texttt{First\ Name:\ {[}GIVEN\_\allowbreak{}NAME\_\allowbreak{}1{]}} \\
TranslateGemma output &
\texttt{Nombre:\ {[}GIVEN\_\allowbreak{}NAME\_\allowbreak{}1{]}} \\
\end{longtable}
\fi

Marker-preserving translation has been used to transfer span labels
across languages \citep{chen2023-easyproject}. An alternative generates
candidate target spans and ranks them by translation probability before
transferring the source label \citep{garciaferrero2023-tprojection}. We
used placeholders instead, because typed names such as
\texttt{{[}GIVEN\_\allowbreak{}NAME\_\allowbreak{}1{]}} seemed more
direct to interpret in LLM outputs; this was a design preference,
without a comparative test against candidate ranking.

To test the checked-translation route, we replace half of the existing
translated-family sampling mass for Arabic, Czech, Hindi, Indonesian,
Dutch, Portuguese, Swedish, and Vietnamese. TranslateGemma-27B produces
the primary translation; deterministic placeholder, schema, and
structure checks either admit it, route it to a
Gemma\nobreak\hbox{-}\nobreak{}4-31B fallback, or reject the row. The
resulting route supplies 24,338 weighted windows, or 9.507\% of the
unchanged 256,000-window logical epoch. Language totals, every
non-transport source, the 15\% native-gold share, and the 37.425\%
English share are held fixed.

For each architecture, we train the checked-translation and
unchanged-data conditions for the same number of optimizer updates from
a shared starting checkpoint. XLM\nobreak\hbox{-}\nobreak{}R is
evaluated after 125 updates and mmBERT after 250. The Fresh20
development set contains 1,267 documents in twenty languages; the
segmented synthetic clinical MultiGraSCCo evaluation contains 630
document-language instances (Section~\ref{sec-evaluation-sets}). Each
report uses 10,000 paired document resamples. The results are:

\ACLwidetabletrue\ACLcontinuedtablefalse

\ifXeTeX
\begin{longtable}[]{@{}
  >{\raggedright\arraybackslash}p{(\linewidth - 6\tabcolsep) * \real{0.2143}}
  >{\raggedright\arraybackslash}p{(\linewidth - 6\tabcolsep) * \real{0.2143}}
  >{\raggedleft\arraybackslash}p{(\linewidth - 6\tabcolsep) * \real{0.2857}}
  >{\raggedleft\arraybackslash}p{(\linewidth - 6\tabcolsep) * \real{0.2857}}@{}}
\caption{Checked translations versus unchanged training data, at matched
exposure within each encoder. Fresh20 has 1,267 documents; MultiGraSCCo
has 630 document-language instances. Deltas are F1 points with paired
95\% intervals. }\label{tbl-checked-translation}\tabularnewline
\toprule\noalign{}
\begin{minipage}[b]{\linewidth}\raggedright
Surface
\end{minipage} & \begin{minipage}[b]{\linewidth}\raggedright
View
\end{minipage} & \begin{minipage}[b]{\linewidth}\raggedleft
XLM\nobreak\hbox{-}\nobreak{}R Δ (F1 points) {[}95\% interval{]}
\end{minipage} & \begin{minipage}[b]{\linewidth}\raggedleft
mmBERT Δ (F1 points) {[}95\% interval{]}
\end{minipage} \\
\midrule\noalign{}
\endfirsthead
\toprule\noalign{}
\begin{minipage}[b]{\linewidth}\raggedright
Surface
\end{minipage} & \begin{minipage}[b]{\linewidth}\raggedright
View
\end{minipage} & \begin{minipage}[b]{\linewidth}\raggedleft
XLM\nobreak\hbox{-}\nobreak{}R Δ (F1 points) {[}95\% interval{]}
\end{minipage} & \begin{minipage}[b]{\linewidth}\raggedleft
mmBERT Δ (F1 points) {[}95\% interval{]}
\end{minipage} \\
\midrule\noalign{}
\endhead
\bottomrule\noalign{}
\endlastfoot
Fresh20 & span overlap & +0.59 {[}+0.13, +1.06{]} & +0.44 {[}+0.18,
+0.71{]} \\
Fresh20 & fine overlap & +0.20 {[}-0.08, +0.49{]} & +0.30 {[}+0.08,
+0.53{]} \\
MultiGraSCCo & span overlap & +0.15 {[}+0.04, +0.26{]} & +0.38 {[}+0.22,
+0.53{]} \\
MultiGraSCCo & fine overlap & +0.05 {[}-0.04, +0.13{]} & +0.44 {[}+0.28,
+0.61{]} \\
\end{longtable}
\else
\begin{longtable}[]{@{}
  >{\raggedright\arraybackslash}X
  >{\raggedright\arraybackslash}X
  >{\raggedleft\arraybackslash}X
  >{\raggedleft\arraybackslash}X@{}}
\caption{Checked translations versus unchanged training data, at matched
exposure within each encoder. Fresh20 has 1,267 documents; MultiGraSCCo
has 630 document-language instances. Deltas are F1 points with paired
95\% intervals. }\label{tbl-checked-translation}\tabularnewline
\toprule\noalign{}
\raggedright
Surface
 & \raggedright
View
 & \raggedleft
XLM\nobreak\hbox{-}\nobreak{}R Δ (F1 points) {[}95\% interval{]}
 & \raggedleft
mmBERT Δ (F1 points) {[}95\% interval{]}
 \\
\midrule\noalign{}
\endfirsthead
\toprule\noalign{}
\raggedright
Surface
 & \raggedright
View
 & \raggedleft
XLM\nobreak\hbox{-}\nobreak{}R Δ (F1 points) {[}95\% interval{]}
 & \raggedleft
mmBERT Δ (F1 points) {[}95\% interval{]}
 \\
\midrule\noalign{}
\endhead
\bottomrule\noalign{}
\endlastfoot
Fresh20 & span overlap & +0.59 {[}+0.13, +1.06{]} & +0.44 {[}+0.18,
+0.71{]} \\
Fresh20 & fine overlap & +0.20 {[}-0.08, +0.49{]} & +0.30 {[}+0.08,
+0.53{]} \\
MultiGraSCCo & span overlap & +0.15 {[}+0.04, +0.26{]} & +0.38 {[}+0.22,
+0.53{]} \\
MultiGraSCCo & fine overlap & +0.05 {[}-0.04, +0.13{]} & +0.44 {[}+0.28,
+0.61{]} \\
\end{longtable}
\fi

The checked-translation route improves span-overlap F1 for both encoders
on both evaluation sets. The comparison combines translation, fallback
and structural checks; the COMETKiwi quality filter rejected no
structurally clean primary output. XLM\nobreak\hbox{-}\nobreak{}R
nevertheless loses Korean F1 in both Fresh20 views, while mmBERT has no
supported per-language loss. Machine-readable reports preserve the
\href{../../evidence/xlmr-tg27-routed05-s1375-vs-control-s1375-fresh20-bootstrap-v1.json}{XLM\nobreak\hbox{-}\nobreak{}R
Fresh20},
\href{../../evidence/xlmr-tg27-routed05-s1375-vs-control-s1375-mgs10-bootstrap-v1.json}{XLM\nobreak\hbox{-}\nobreak{}R
MultiGraSCCo},
\href{../../evidence/mmbert-tg27-routed05-s1250-vs-control-s1250-fresh20-bootstrap-v1.json}{mmBERT
Fresh20}, and
\href{../../evidence/mmbert-tg27-routed05-s1250-vs-control-s1250-mgs10-bootstrap-v1.json}{mmBERT
MultiGraSCCo} comparisons.

\paragraph{Comparison using the deployment
decoder}\label{sec-transport-deployment}

A separate XLM\nobreak\hbox{-}\nobreak{}R comparison added the routed
translations together with identifier-label repairs and locale-format
changes; both models used the same greedy deployment decoder. It merges
inside/end-token continuations into an open span even when their fine
labels differ, then assigns the fine type with the greatest total
token-character coverage within that span. The outside-label
(\textbf{O}) logit bias is zero. The combined update improves untyped
span detection at a fine-label cost: the MultiGraSCCo fine-label loss
has a paired interval entirely below zero, whereas the Fresh20
fine-label interval includes zero. These results concern the bundled
update, not the translator alone. Changes below are language-macro
overlap-F1 percentage points, with 95\% paired intervals.

\ACLwidetabletrue\ACLcontinuedtablefalse

\ifXeTeX
\begin{longtable}[]{@{}
  >{\raggedright\arraybackslash}p{(\linewidth - 4\tabcolsep) * \real{0.2727}}
  >{\raggedleft\arraybackslash}p{(\linewidth - 4\tabcolsep) * \real{0.3636}}
  >{\raggedleft\arraybackslash}p{(\linewidth - 4\tabcolsep) * \real{0.3636}}@{}}
\caption{Routed translations, identifier-label repairs and locale
changes evaluated with the same deployment decoder. Entries are
language-macro overlap-F1 changes in points, with paired 95\% intervals;
untyped gains accompany fine-label
losses.}\label{tbl-transport-deployment}\tabularnewline
\toprule\noalign{}
\begin{minipage}[b]{\linewidth}\raggedright
Evaluation
\end{minipage} & \begin{minipage}[b]{\linewidth}\raggedleft
Untyped spans
\end{minipage} & \begin{minipage}[b]{\linewidth}\raggedleft
Fine labels
\end{minipage} \\
\midrule\noalign{}
\endfirsthead
\toprule\noalign{}
\begin{minipage}[b]{\linewidth}\raggedright
Evaluation
\end{minipage} & \begin{minipage}[b]{\linewidth}\raggedleft
Untyped spans
\end{minipage} & \begin{minipage}[b]{\linewidth}\raggedleft
Fine labels
\end{minipage} \\
\midrule\noalign{}
\endhead
\bottomrule\noalign{}
\endlastfoot
Fresh20 (twenty-language development set) & +1.36 {[}0.90, 1.82{]} &
−0.45 {[}−1.02, 0.10{]} \\
MultiGraSCCo & +0.46 {[}0.24, 0.68{]} & −0.90 {[}−1.19, −0.61{]} \\
\end{longtable}
\else
\begin{longtable}[]{@{}
  >{\raggedright\arraybackslash}X
  >{\raggedleft\arraybackslash}X
  >{\raggedleft\arraybackslash}X@{}}
\caption{Routed translations, identifier-label repairs and locale
changes evaluated with the same deployment decoder. Entries are
language-macro overlap-F1 changes in points, with paired 95\% intervals;
untyped gains accompany fine-label
losses.}\label{tbl-transport-deployment}\tabularnewline
\toprule\noalign{}
\raggedright
Evaluation
 & \raggedleft
Untyped spans
 & \raggedleft
Fine labels
 \\
\midrule\noalign{}
\endfirsthead
\toprule\noalign{}
\raggedright
Evaluation
 & \raggedleft
Untyped spans
 & \raggedleft
Fine labels
 \\
\midrule\noalign{}
\endhead
\bottomrule\noalign{}
\endlastfoot
Fresh20 (twenty-language development set) & +1.36 {[}0.90, 1.82{]} &
−0.45 {[}−1.02, 0.10{]} \\
MultiGraSCCo & +0.46 {[}0.24, 0.68{]} & −0.90 {[}−1.19, −0.61{]} \\
\end{longtable}
\fi

\paragraph{Translation placeholder
repair}\label{translation-placeholder-repair}

To retain entity labels through translation, we sent markers such as
\texttt{{[}GIVEN\_\allowbreak{}NAME\_\allowbreak{}1{]}} to
TranslateGemma: an uppercase, underscore-separated class name followed
by \texttt{\_\allowbreak{}N}, the numeric slot identifier expected to
survive translation. If translation changed the class name, we restored
the source type using the numeric ID. We required every ID to occur the
expected number of times and rejected translations containing leaked
instruction text. We also filtered translations with COMETKiwi, a
machine-translation quality-estimation (MTQE) model that scores a source
and its translation without a reference translation. We retried failures
with Gemma\nobreak\hbox{-}\nobreak{}4-31B under the same checks.

\ACLwidetabletrue\ACLcontinuedtablefalse

\ifXeTeX
\begin{longtable}[]{@{}
  >{\raggedright\arraybackslash}p{(\linewidth - 4\tabcolsep) * \real{0.2727}}
  >{\raggedleft\arraybackslash}p{(\linewidth - 4\tabcolsep) * \real{0.3636}}
  >{\raggedleft\arraybackslash}p{(\linewidth - 4\tabcolsep) * \real{0.3636}}@{}}
\caption{Translation-route acceptance after marker restoration and
automated checks across eight target languages. Acceptance measures
structural and quality-filter passes, not human-reviewed annotation
accuracy.}\label{tbl-transport-acceptance}\tabularnewline
\toprule\noalign{}
\begin{minipage}[b]{\linewidth}\raggedright
Stage
\end{minipage} & \begin{minipage}[b]{\linewidth}\raggedleft
Accepted at this stage
\end{minipage} & \begin{minipage}[b]{\linewidth}\raggedleft
Cumulative acceptance
\end{minipage} \\
\midrule\noalign{}
\endfirsthead
\toprule\noalign{}
\begin{minipage}[b]{\linewidth}\raggedright
Stage
\end{minipage} & \begin{minipage}[b]{\linewidth}\raggedleft
Accepted at this stage
\end{minipage} & \begin{minipage}[b]{\linewidth}\raggedleft
Cumulative acceptance
\end{minipage} \\
\midrule\noalign{}
\endhead
\bottomrule\noalign{}
\endlastfoot
TranslateGemma-27B, after marker repair & 68,309 / 69,480 (98.31\%) &
98.31\% \\
Gemma\nobreak\hbox{-}\nobreak{}4-31B retry of the 1,171 failures & 1,007
/ 1,171 (86.00\%) & \textbf{99.76\%} \\
\end{longtable}
\else
\begin{longtable}[]{@{}
  >{\raggedright\arraybackslash}X
  >{\raggedleft\arraybackslash}X
  >{\raggedleft\arraybackslash}X@{}}
\caption{Translation-route acceptance after marker restoration and
automated checks across eight target languages. Acceptance measures
structural and quality-filter passes, not human-reviewed annotation
accuracy.}\label{tbl-transport-acceptance}\tabularnewline
\toprule\noalign{}
\raggedright
Stage
 & \raggedleft
Accepted at this stage
 & \raggedleft
Cumulative acceptance
 \\
\midrule\noalign{}
\endfirsthead
\toprule\noalign{}
\raggedright
Stage
 & \raggedleft
Accepted at this stage
 & \raggedleft
Cumulative acceptance
 \\
\midrule\noalign{}
\endhead
\bottomrule\noalign{}
\endlastfoot
TranslateGemma-27B, after marker repair & 68,309 / 69,480 (98.31\%) &
98.31\% \\
Gemma\nobreak\hbox{-}\nobreak{}4-31B retry of the 1,171 failures & 1,007
/ 1,171 (86.00\%) & \textbf{99.76\%} \\
\end{longtable}
\fi

Only 164 inputs (0.24\%) remained rejected. Most retries followed
structural failures; 200 also or instead failed the MTQE filter. These
counts cover eight target languages and are measured after marker
restoration.

\paragraph{Entity preservation versus
localization}\label{entity-preservation-versus-localization}

We explored preserving aligned source-language entity values, replacing
them after translation, and asking the translation model to localize
them jointly with the surrounding text. Replacement could also happen
each time a training example was sampled, either keeping the original
value or drawing a new one from locale-aware generators or compatible
observed surfaces. This separated translation of the sentence from
variation in its entity values.

The preferred choice depended on whether replacement was needed. In the
matched Ont3 training comparison, keeping original surfaces outperformed
replacing organization names at sampling time with observed donors,
values of the same type taken from other documents. When filling a
placeholder, observed donors were more reliable than free character
generation in our development review. Thus we retained original surfaces
by default for that Ont3 recipe, with observed donors preferred when
replacement was required. Section~\ref{sec-character-surface-generation}
separates these comparisons and describes the tested character and
token-encoded context models.

\begin{figure*}[tp]

\centering{

\ifXeTeX
\includegraphics[width=1\linewidth,height=6in,alt={A translated Turkish marker can be identified by its stable numeric suffix; a space before a postal-code frame is rejected, and a valid frame can still contain the wrong kind of place.}]{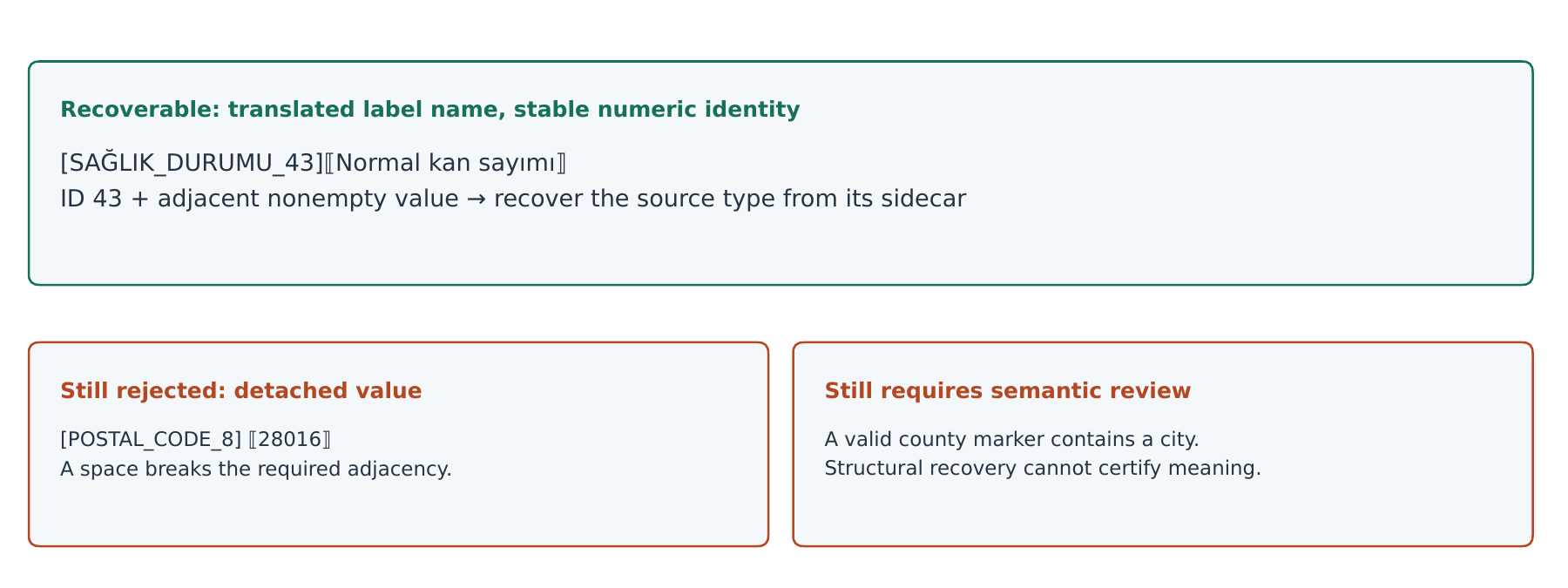}
\else
\includegraphics[alt={A translated Turkish marker can be identified by its stable numeric suffix; a space before a postal-code frame is rejected, and a valid frame can still contain the wrong kind of place.}]{figures/transport-recovery-boundaries-v1.pdf}
\fi

}

\caption{\label{fig-transport-recovery}Recovery boundaries in saved
transport output. The Turkish clinical example means ``normal blood
count'': its translated label name retains numeric ID 43 and an adjacent
value. A Unicode-aware validator retrospectively recovers 19 Turkish
rows in a 124-output, four-language TranslateGemma frame control.
Detached values remain rejected. The county/city example summarizes the
separate Gemma\nobreak\hbox{-}\nobreak{}4 pilot's semantic failure.}

\end{figure*}

TranslateGemma does not accept non-translation task instructions, so we
cannot instruct it to preserve a placeholder's type name. The numeric
suffix still identifies the source label: inspected examples such as the
Turkish marker in Figure~\ref{fig-transport-recovery} retain the
intended alignment. We restore the label from the source map, then
replace the entire marker with the entity value. The translated type
name never enters the encoder's training text. Missing IDs or lost value
boundaries still require rejection or retry.

To avoid wrong-language entity values in translated training data, we
piloted replacing bracketed values such as
\texttt{{[}GIVEN\_\allowbreak{}NAME\_\allowbreak{}1{]}⟦Kevin⟧} with
locally appropriate ones. We compared TranslateGemma's native
translation interface with an explicit Gemma\nobreak\hbox{-}\nobreak{}4
instruction, using 12B models. This pilot did not supply training data:
all 69,480 inputs in the reported training route used ID-only
placeholders followed by separate value generation.

\begin{figure*}[tp]

\centering{

\textbf{Gemma\nobreak\hbox{-}\nobreak{}4 prompt --- English to Spanish}

\begin{quote}
Translate and locally pseudonymize this English sentence into Spanish.
Every entity slot has the exact form
\texttt{{[}TYPE\_\allowbreak{}number{]}⟦source\ surface⟧}. Copy
\texttt{{[}TYPE\_\allowbreak{}number{]}} exactly once per occurrence,
then replace only its adjacent \texttt{⟦source\ surface⟧} with a
plausible Spanish surface of the same primary entity type. Every copied
slot must be followed immediately by one nonempty \texttt{⟦surface⟧}.
Output only the translated sentence.
\end{quote}

\textbf{Input:}
\texttt{-\ First\ Name:\ {[}GIVEN\_\allowbreak{}NAME\_\allowbreak{}1{]}⟦Kevin⟧}

\textbf{Output --- Gemma\nobreak\hbox{-}\nobreak{}4:}
\texttt{-\ Nombre:\ {[}GIVEN\_\allowbreak{}NAME\_\allowbreak{}1{]}⟦Carlos⟧}

}

\caption{\label{fig-gemma-pseudonymization}Instruction and example for
translating a typed entity slot while replacing its value with a locally
appropriate name; bracketed slot syntax is literal.}

\end{figure*}%

Gemma\nobreak\hbox{-}\nobreak{}4 replaced Kevin with Carlos and
preserved its label and boundaries. TranslateGemma returned
\texttt{-\ Nombre:\ {[}GIVEN\_\allowbreak{}NAME\_\allowbreak{}1{]}\ Kevin},
retaining the source name and losing the delimiters.

\begin{figure*}[tp]

\centering{

\ifXeTeX
\includegraphics[width=1\linewidth,height=6in,alt={Two source name fields, with failed TranslateGemma outputs beside structurally valid prompted Gemma‑4 replacements.}]{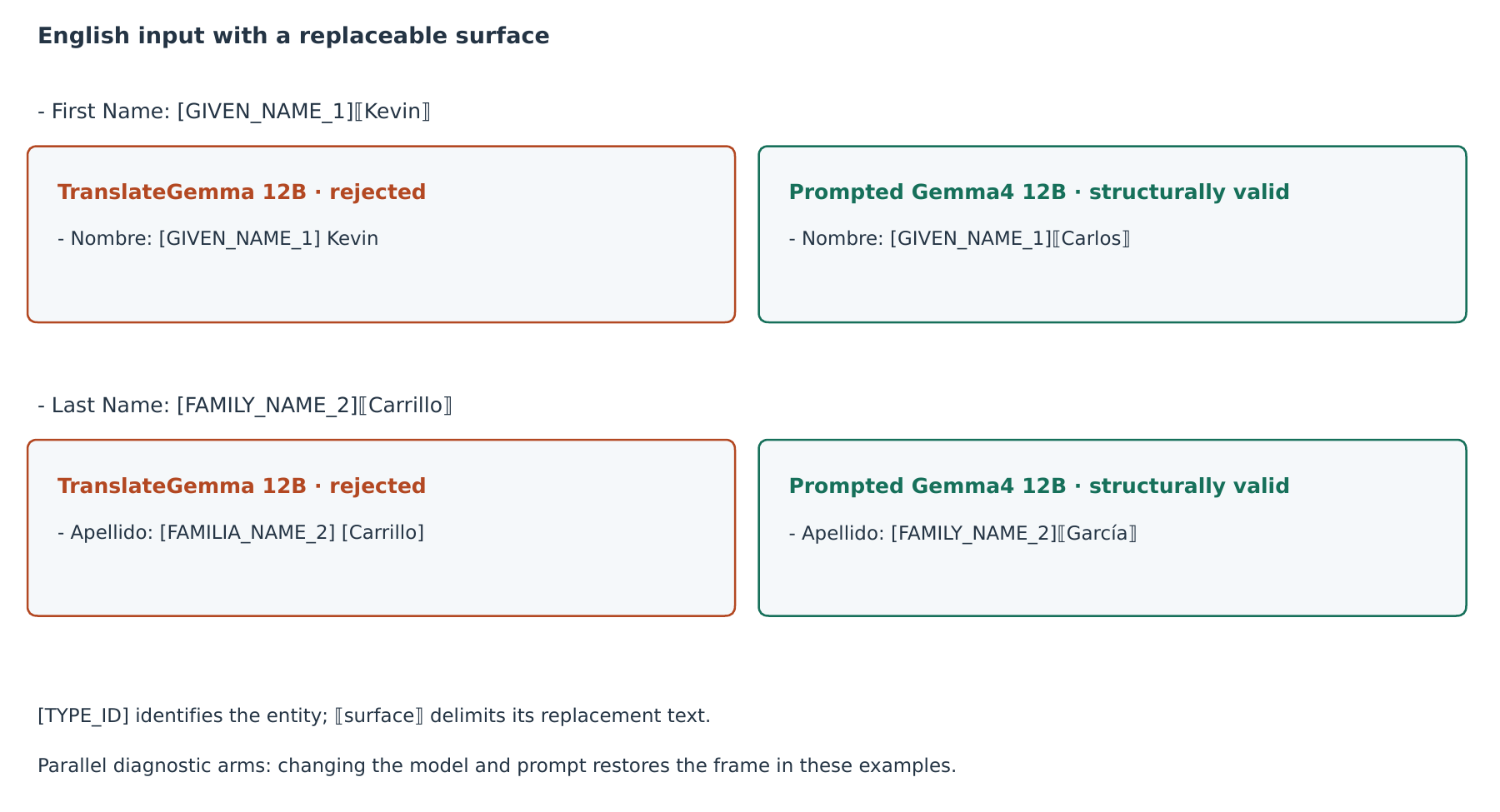}
\else
\includegraphics[alt={Two source name fields, with failed TranslateGemma outputs beside structurally valid prompted Gemma‑4 replacements.}]{figures/transport-model-examples-v1.pdf}
\fi

}

\caption{\label{fig-transport-examples}Literal paired Spanish outputs
from a nine-sentence, three-document pilot covering Spanish, Indonesian
and Japanese. TranslateGemma-12B loses the surface delimiters; prompted
Gemma\nobreak\hbox{-}\nobreak{}4-12B preserves them and supplies
replacement names.}

\end{figure*}%

The pilot's five entity-bearing sentences come from the first three
documents of an audit set covering Spanish, Indonesian and Japanese
(Figure~\ref{fig-transport-examples}). Gemma\nobreak\hbox{-}\nobreak{}4
preserved the format in all five but supplied city names for two county
slots; no output repairs were applied.

\ACLwidetabletrue\ACLcontinuedtablefalse

\ifXeTeX
\begin{longtable}[]{@{}lrr@{}}
\caption{Joint translation and entity replacement on five entity-bearing
sentences. Format requires intact markers and value delimiters;
replacement type also requires a value of the requested entity
type.}\label{tbl-replacement-pilot}\tabularnewline
\toprule\noalign{}
Model & Format passed & Format and replacement type passed \\
\midrule\noalign{}
\endfirsthead
\toprule\noalign{}
Model & Format passed & Format and replacement type passed \\
\midrule\noalign{}
\endhead
\bottomrule\noalign{}
\endlastfoot
TranslateGemma-12B & 0\% (0/5) & 0\% (0/5) \\
Prompted Gemma\nobreak\hbox{-}\nobreak{}4-12B & 100\% (5/5) & 60\%
(3/5) \\
\end{longtable}
\else
\begin{longtable}[]{@{}lrr@{}}
\caption{Joint translation and entity replacement on five entity-bearing
sentences. Format requires intact markers and value delimiters;
replacement type also requires a value of the requested entity
type.}\label{tbl-replacement-pilot}\tabularnewline
\toprule\noalign{}
Model & Format passed & Format and replacement type passed \\
\midrule\noalign{}
\endfirsthead
\toprule\noalign{}
Model & Format passed & Format and replacement type passed \\
\midrule\noalign{}
\endhead
\bottomrule\noalign{}
\endlastfoot
TranslateGemma-12B & 0\% (0/5) & 0\% (0/5) \\
Prompted Gemma\nobreak\hbox{-}\nobreak{}4-12B & 100\% (5/5) & 60\%
(3/5) \\
\end{longtable}
\fi

TranslateGemma preserved ordinary typed placeholders in all nine
sentences of the same test; the failures above concern the added value
delimiters.

Accepting non-ASCII marker names and recovering their labels by numeric
ID rescued 19 additional outputs in a separate clinical translation test
of 124 outputs, 31 each in Arabic, Hindi, Japanese and Turkish. All 19
recovered outputs were Turkish; missing or detached value delimiters
remained rejected.

\ACLwidetabletrue\ACLcontinuedtablefalse

\ifXeTeX
\begin{longtable}[]{@{}lrr@{}}
\caption{Structural acceptance in a 124-output, four-language
translation test. Recovering translated marker names by numeric ID
rescues 19 Turkish outputs; semantic correctness was not
measured.}\label{tbl-marker-recovery}\tabularnewline
\toprule\noalign{}
Model & Before repair & After numeric-ID recovery \\
\midrule\noalign{}
\endfirsthead
\toprule\noalign{}
Model & Before repair & After numeric-ID recovery \\
\midrule\noalign{}
\endhead
\bottomrule\noalign{}
\endlastfoot
TranslateGemma-12B & 25.0\% (31/124) & 40.3\% (50/124) \\
\end{longtable}
\else
\begin{longtable}[]{@{}lrr@{}}
\caption{Structural acceptance in a 124-output, four-language
translation test. Recovering translated marker names by numeric ID
rescues 19 Turkish outputs; semantic correctness was not
measured.}\label{tbl-marker-recovery}\tabularnewline
\toprule\noalign{}
Model & Before repair & After numeric-ID recovery \\
\midrule\noalign{}
\endfirsthead
\toprule\noalign{}
Model & Before repair & After numeric-ID recovery \\
\midrule\noalign{}
\endhead
\bottomrule\noalign{}
\endlastfoot
TranslateGemma-12B & 25.0\% (31/124) & 40.3\% (50/124) \\
\end{longtable}
\fi

We have not measured how many outputs pass semantic review after these
repairs, or how many failed translations
Gemma\nobreak\hbox{-}\nobreak{}4 can recover in this bracketed format.

\subsubsection{Positive-only teacher
supervision}\label{sec-teacher-omission-controls}

This comparison changes the loss attached to omissions in one fixed set
of 1,652 training rows labeled by a teacher annotator,
Gemma\nobreak\hbox{-}\nobreak{}4-31B-IT. Under complete supervision,
teacher omissions become trusted non-entity (\textbf{O}) targets with
loss weight \(\lambda=1\). Under positive-only supervision
(\(\lambda=0\)), teacher-emitted BIOES labels retain ordinary positive
loss and every other teacher-row token contributes no classification
loss. Fully labeled gold and other complete-label rows retain their
ordinary \textbf{O} loss. Fixed across conditions are the starting
checkpoint, a 10\% sampling share for teacher rows, a 15\% native-gold
share, 4:2:1 language weights (Section~\ref{sec-evaluation-sets}), a
1,000-step cosine schedule, and decoding at zero \textbf{O} bias.

Reducing the \textbf{O} loss weight to 0.75 on fully annotated data also
improved Ont2 F1, primarily by shifting the precision--recall operating
point toward recall (Section~\ref{sec-operating-points}). Here we vary
the weight of uncertain negatives from teacher omissions instead.

\ACLwidetabletrue\ACLcontinuedtablefalse

\ifXeTeX
\begin{longtable}[]{@{}
  >{\raggedright\arraybackslash}p{(\linewidth - 6\tabcolsep) * \real{0.2143}}
  >{\raggedleft\arraybackslash}p{(\linewidth - 6\tabcolsep) * \real{0.2857}}
  >{\raggedleft\arraybackslash}p{(\linewidth - 6\tabcolsep) * \real{0.2857}}
  >{\raggedright\arraybackslash}p{(\linewidth - 6\tabcolsep) * \real{0.2143}}@{}}
\caption{Ignoring unmarked Gemma teacher tokens versus treating them as
background. The contrast averages the two positive-only seeds 155 and
156 and subtracts the existing complete-supervision seed-155 baseline.
Results are on Fresh20, using 10,000 paired document resamples.
Intervals are per metric; win/loss means the interval excludes zero
after simultaneous adjustment across 276 comparisons, and unresolved
means it includes zero.
}\label{tbl-positive-only-controls}\tabularnewline
\toprule\noalign{}
\begin{minipage}[b]{\linewidth}\raggedright
Weighted view
\end{minipage} & \begin{minipage}[b]{\linewidth}\raggedleft
Mean delta (points)
\end{minipage} & \begin{minipage}[b]{\linewidth}\raggedleft
Paired 95\% interval (points)
\end{minipage} & \begin{minipage}[b]{\linewidth}\raggedright
Adjusted status
\end{minipage} \\
\midrule\noalign{}
\endfirsthead
\toprule\noalign{}
\begin{minipage}[b]{\linewidth}\raggedright
Weighted view
\end{minipage} & \begin{minipage}[b]{\linewidth}\raggedleft
Mean delta (points)
\end{minipage} & \begin{minipage}[b]{\linewidth}\raggedleft
Paired 95\% interval (points)
\end{minipage} & \begin{minipage}[b]{\linewidth}\raggedright
Adjusted status
\end{minipage} \\
\midrule\noalign{}
\endhead
\bottomrule\noalign{}
\endlastfoot
span overlap & +1.00 & {[}+0.73, +1.40{]} & win \\
character precision & -0.26 & {[}-0.38, -0.17{]} & loss \\
character recall & +2.80 & {[}+2.40, +3.20{]} & win \\
character F1 & +1.50 & {[}+1.20, +1.70{]} & win \\
fine overlap & +0.11 & {[}-0.20, +0.42{]} & unresolved \\
\end{longtable}
\else
\begin{longtable}[]{@{}
  >{\raggedright\arraybackslash}X
  >{\raggedleft\arraybackslash}X
  >{\raggedleft\arraybackslash}X
  >{\raggedright\arraybackslash}X@{}}
\caption{Ignoring unmarked Gemma teacher tokens versus treating them as
background. The contrast averages the two positive-only seeds 155 and
156 and subtracts the existing complete-supervision seed-155 baseline.
Results are on Fresh20, using 10,000 paired document resamples.
Intervals are per metric; win/loss means the interval excludes zero
after simultaneous adjustment across 276 comparisons, and unresolved
means it includes zero.
}\label{tbl-positive-only-controls}\tabularnewline
\toprule\noalign{}
\raggedright
Weighted view
 & \raggedleft
Mean delta (points)
 & \raggedleft
Paired 95\% interval (points)
 & \raggedright
Adjusted status
 \\
\midrule\noalign{}
\endfirsthead
\toprule\noalign{}
\raggedright
Weighted view
 & \raggedleft
Mean delta (points)
 & \raggedleft
Paired 95\% interval (points)
 & \raggedright
Adjusted status
 \\
\midrule\noalign{}
\endhead
\bottomrule\noalign{}
\endlastfoot
span overlap & +1.00 & {[}+0.73, +1.40{]} & win \\
character precision & -0.26 & {[}-0.38, -0.17{]} & loss \\
character recall & +2.80 & {[}+2.40, +3.20{]} & win \\
character F1 & +1.50 & {[}+1.20, +1.70{]} & win \\
fine overlap & +0.11 & {[}-0.20, +0.42{]} & unresolved \\
\end{longtable}
\fi

Seed 156 independently reproduces the direction: span overlap, character
recall, character F1, and precision change +1.1/+2.8/+1.5/−0.25
percentage points. Mechanism alternatives move in the opposite
direction. A retention term toward the starting checkpoint's posteriors,
at weight \texttt{.25}, gives back 0.25 span-overlap F1 points and 0.41
character-F1 points. Assigning omitted teacher-row tokens partial
\textbf{O} loss weights \(\lambda=0.05\) and \(\lambda=0.20\) changes
span overlap by −0.19 and −0.42 points, with simultaneous recall and
character-F1 losses at both weights. We therefore select \(\lambda=0\).

A separate comparison on the same 1,652 rows measures the teacher data
itself: Gemma\nobreak\hbox{-}\nobreak{}4-31B-IT annotations provided a
modest benefit over the encoder's own labels, raising fine-label F1 by
0.3 points with no clear character-F1 gain.

Relative annotation density provides a practical coverage proxy. We
compare saved first-pass annotations of 1,113 Ont3 development segments
in 35 languages against Luna, the annotator that labeled Ont3 training
data; this diagnostic is separate from the Ont2 training experiment
above. Gemma\nobreak\hbox{-}\nobreak{}4-31B emitted 82\% as many primary
spans as Luna and Qwen3.8-27B 61\%. The deficit was smaller when
reference-type (\texttt{\_\allowbreak{}reference}) tags were excluded.
These are span-count ratios, without adjustment for split or merged
names; they reflect omissions, false positives and span grouping.

\ACLwidetabletrue\ACLcontinuedtablefalse

\ifXeTeX
\begin{longtable}[]{@{}
  >{\raggedright\arraybackslash}p{(\linewidth - 4\tabcolsep) * \real{0.2727}}
  >{\raggedleft\arraybackslash}p{(\linewidth - 4\tabcolsep) * \real{0.3636}}
  >{\raggedleft\arraybackslash}p{(\linewidth - 4\tabcolsep) * \real{0.3636}}@{}}
\caption{First-pass span counts relative to Luna on the same 1,113 Ont3
development segments in 35 languages. Ratios measure annotation density,
not recall; span splitting and false positives also affect
them.}\label{tbl-teacher-density}\tabularnewline
\toprule\noalign{}
\begin{minipage}[b]{\linewidth}\raggedright
Annotator
\end{minipage} & \begin{minipage}[b]{\linewidth}\raggedleft
Span density relative to Luna
\end{minipage} & \begin{minipage}[b]{\linewidth}\raggedleft
Without \texttt{\_\allowbreak{}reference} tags
\end{minipage} \\
\midrule\noalign{}
\endfirsthead
\toprule\noalign{}
\begin{minipage}[b]{\linewidth}\raggedright
Annotator
\end{minipage} & \begin{minipage}[b]{\linewidth}\raggedleft
Span density relative to Luna
\end{minipage} & \begin{minipage}[b]{\linewidth}\raggedleft
Without \texttt{\_\allowbreak{}reference} tags
\end{minipage} \\
\midrule\noalign{}
\endhead
\bottomrule\noalign{}
\endlastfoot
Gemma\nobreak\hbox{-}\nobreak{}4-31B & 82\% & 90\% \\
Qwen3.8-27B & 61\% & 75\% \\
\end{longtable}
\else
\begin{longtable}[]{@{}
  >{\raggedright\arraybackslash}X
  >{\raggedleft\arraybackslash}X
  >{\raggedleft\arraybackslash}X@{}}
\caption{First-pass span counts relative to Luna on the same 1,113 Ont3
development segments in 35 languages. Ratios measure annotation density,
not recall; span splitting and false positives also affect
them.}\label{tbl-teacher-density}\tabularnewline
\toprule\noalign{}
\raggedright
Annotator
 & \raggedleft
Span density relative to Luna
 & \raggedleft
Without \texttt{\_\allowbreak{}reference} tags
 \\
\midrule\noalign{}
\endfirsthead
\toprule\noalign{}
\raggedright
Annotator
 & \raggedleft
Span density relative to Luna
 & \raggedleft
Without \texttt{\_\allowbreak{}reference} tags
 \\
\midrule\noalign{}
\endhead
\bottomrule\noalign{}
\endlastfoot
Gemma\nobreak\hbox{-}\nobreak{}4-31B & 82\% & 90\% \\
Qwen3.8-27B & 61\% & 75\% \\
\end{longtable}
\fi

\paragraph{Boundary conventions in teacher
evaluation}\label{sec-teacher-boundaries}

Boundary conventions can also mimic teacher omissions. For example,
Gemma\nobreak\hbox{-}\nobreak{}4-31B labeled ``Amara Bell'' as one
person span where the reference labeled ``Amara'' and ``Bell''
separately. On Fresh20's 61 English development documents, the same
Gemma\nobreak\hbox{-}\nobreak{}4-31B predictions score 78.0\% exact
typed-span F1, 97.2\% character F1 and 95.7\% character recall under a
shared permissive 12-family type mapping: much of the disagreement
concerned segmentation despite high entity-character coverage. Joining
compatible components in both prediction and reference, then requiring
80\% mutual span coverage, gives 96.8\% span F1. Joining permits gaps of
at most three Unicode characters and can combine distinct nearby
entities, so this permissive score can overstate how safely outer
boundaries could be shifted. Saved examples, counts and provenance are
in the
\href{../../evidence/gemma-qwen-recall-interpretation-correction-v1.md}{boundary
analysis}.

Our revised training objective records internal-segmentation ambiguity
by annotator, language and tag alongside the acceptable-label mapping. A
rule may cover all languages or have language-specific overrides. For a
declared name region, its first token permits B or S, its last permits E
or S, and interior tokens permit any BIOES entity prefix; a single-token
region permits only S. We sum probabilities over these alternatives and
the mapped acceptable types. Adjacent same-type annotations may join
across whitespace, but never across punctuation or an unannotated word.
\textbf{O} tokens retain their original supervision. This is a tokenwise
marginal objective; constrained decoding still enforces legal BIOES
sequences. Component and role annotations retain their original
carriers.

The implementation also reports a separate convention-aware exact score
that permits internal segmentation differences while requiring exact
outer boundaries. Previously reported scores and the training objectives
of completed runs are unchanged. The initial policy covers English
\texttt{person\_\allowbreak{}name} for the identified Gemma training
sources; it does not apply to every model or prompt sharing the Gemma
name. Ont3 training did use Gemma labels: one comparison of annotation
sources replaced the labels of 10,312 matched training segments with
Gemma\nobreak\hbox{-}\nobreak{}4-31B (FP8) annotations. That run used
exact BIOES targets. We have not yet measured the downstream effect of
the revised boundary objective.

\subsubsection{Mapped-gold replay and corpus
coverage}\label{sec-mapped-gold-replay}

Mapped-gold replay assigns 40\% of training draws to four
human-annotated corpora, with labels mapped into Ont3
(Table~\ref{tbl-ont2-ont3-methods}). After correcting the treatment of
unannotated demographic attributes, we continued training and selected
the final checkpoint using development results from both Silver-dev and
gold. A further adaptation to isolated and neighboring-sentence inputs
produced O3 (Section~\ref{sec-neighboring-context}). With
character-boundary refinement
(Section~\ref{sec-boundary-refiner-details}), its best historical
exact-boundary redaction F1 was 79.8\% on Silver-dev and 86.2\% on gold,
under that experiment's references and scoring policy.

The historical matched replay comparison reuses development data. The
same 279-input Ont3 selector chooses checkpoints. At zero \textbf{O}
bias, the replay model scores 84.01\% on gold and 75.65\% on Silver-dev.
Intervals condition on checkpoint and threshold selection. The 40\%
experiment includes entity-free gold inputs and masks only omitted
alternate-positive regions in nested annotation views. A 25\% experiment
using positive-bearing inputs scores 83.99\% and 76.18\%; the additional
0.5-point sum gain remains unresolved (95\% interval {[}-1.07, 2.01{]}).
Training took 35 minutes on one RTX PRO 6000 GPU. This continuation
starts from a separate checkpoint trained with \textbf{O} loss weight
0.75. The O3 curve in the main comparison
(Figure~\ref{fig-final20-priority9-overlap}) uses a continuation of it
that treats unannotated text as negative only for types the source
corpus annotates, described in Section~\ref{sec-ontology-comparison}.

Compatible labels can still leave mentions unannotated: reviewing
predictions on 400 training inputs from each corpus finds genuine
missing mentions alongside model errors. For example, Wojood occupation
and language labels map to demographic attributes but do not cover every
Ont3 personal descriptor. The runs above retain their mapping-derived
training coverage. Evaluation exempts unannotated Wojood demographics
for every model, as described in Section~\ref{sec-priority9-comparison}.

\subsubsection{Low-dose natural-text and MLM
replay}\label{sec-mlm-controls}

We mixed unlabeled natural text into privacy-label training and asked
the encoder to predict randomly hidden tokens, the masked-language-model
(MLM) objective. As with weight priors and frozen encoder layers, we
tried this objective to prevent forgetting during specialization for
redaction. Related work studies continued pretraining and preservation
of cross-lingual representations during task adaptation
\citep{gururangan2020-dont-stop, liu2021-preserving-crosslinguality}.
MLM is most attractive when acquiring in-domain annotations is
expensive; with those annotations available, these results do not
justify its additional cost.

Before adding natural-text teacher annotations, we tested
masked-language replay from a twenty-language redaction checkpoint with
a fully trainable encoder. Its supervised training data contained
306,439 windows: 256,000 from the existing mixture and 50,439 routed
translation windows. It contained no Gemma-labeled FineWeb training
windows. In the selected recipe, fifteen percent of sampled rows are
natural multilingual text carrying XLM\nobreak\hbox{-}\nobreak{}R's
original MLM objective at loss weight \texttt{.01}; the fixed original
MLM head supplies the readout while both tagging and MLM losses update
the shared encoder. The recipe uses learning rate \texttt{3e-6}, a
1,500-step cosine schedule, and the predeclared externally selected
checkpoint at step 250.

On the twenty-language Fresh20 development set, the selected recipe
raises language-weighted span-overlap F1 over its starting checkpoint by
1.2 points (paired 95\% interval {[}0.87, 1.54{]}). Four additional seed
runs reproduced the recipe's positive direction. On a fixed set of
masked tokens, prediction loss fell from 11.15 at the supervised
starting checkpoint to 9.28, compared with 1.21 for pretrained
XLM\nobreak\hbox{-}\nobreak{}R.

The recipe jointly changes natural-text replay, learning rate and the
MLM mixture. A matched zero-MLM control is missing, so attribution
remains at the recipe level.

A broader-data screen then added two teacher-labeled training sets,
varying their combined sampling share across 20\%, 30\% and 40\%,
alongside separate FineWeb MLM text. Those sets supplied 1,240 and 5,510
nonempty annotated windows across all 20 languages; partial-supervision
rows with no marked span were excluded. Together with 256,000 existing
windows, this supervised training data contained 262,750 windows. The
40\% setting assigned each teacher set 20\% of supervised sampling and
ran for the full trajectory. Its best saved checkpoint at the neutral
threshold was 0.4 span-overlap F1 points below the existing model on the
weighted twenty-language evaluation. Threshold calibration on a separate
200-document set left it 0.5 points below. Another checkpoint gained 0.6
character-recall points while losing 0.4 span-F1 points. The
\href{../../swept/pii-robust-buildout.md}{buildout record} preserves the
August 13 broad-replay comparison. These are combined-data/objective
results; no matched run isolated MLM's contribution.

In a timing benchmark at effective batch size 64, optimizer-step time
rose from 0.27 seconds without MLM to 0.38 seconds at 15\% replay
(+41.1\%). At 5\% replay, the increase was 20--23\%; at 1\%, it was
3.4\%. These costs depend on replay frequency: reducing only the MLM
loss weight does not avoid its forward pass and vocabulary loss.

To isolate MLM's contribution, we later compared it with continued
supervised training in an
\href{../../encoder-drift-prevention.md}{encoder-retention study}. At
learning rate \texttt{3e-5}, we tested 15\% MLM sampling with weights
\texttt{.001/.003/.005/.01/.03}, 5\% with \texttt{.003/.01}, and 1\%
with \texttt{.015}; none gives a resolved tagging gain, and neither do
lower-rate or additional-seed tests. Hidden-token prediction loss on
fixed masks improves in several arms, but broader language competence
was not measured. These results concern separate natural-text replay,
not MLM on the existing labeled text.

\subsection{Beyond mapped gold}\label{sec-ont2-transfer}

This section describes transfer from Ont1 supervision to Ont2 and
evaluates one Ont2 checkpoint. On fresh English legal text it left a
large gap: exact typed span F1 was 42.5\%, below the prespecified 85\%
target.

\subsubsection{Ont1-to-Ont2 transfer}\label{ont1-to-ont2-transfer}

We reused an Ont1-trained XLM\nobreak\hbox{-}\nobreak{}R-large encoder
and created an affine head with rows for the Ont2 tags, initialized from
the mapped Ont1 rows. As Ont2 annotations became available, training
continued to include Ont1-labeled examples and mapped gold. Reviewed
many-to-many label mappings let those older annotations supervise the
Ont2 head.

Earlier transition experiments gradually reduced the old head's loss
weight to zero (Section~\ref{sec-ontology-transition}). The selected
Ont2 lineage used zero old-head loss weight from its first stage: Ont1
annotations continued to contribute through their projected supervision
of the Ont2 head. Training for O3, the Ont3 predecessor, added more
varied annotated data, mapped-gold replay and mixed-context training
(Section~\ref{sec-ontology-comparison}).

\subsubsection{Domain shift to fresh court
text}\label{domain-shift-to-fresh-court-text}

\begin{figure*}[tp]

\centering{

\ifXeTeX
\includegraphics[width=1\linewidth,height=6in,alt={Ont2 adapted successor scores 90.2 on the older Final20 retention set, 86.4 on clinical-heavy development and 42.5 on fresh English legal text.}]{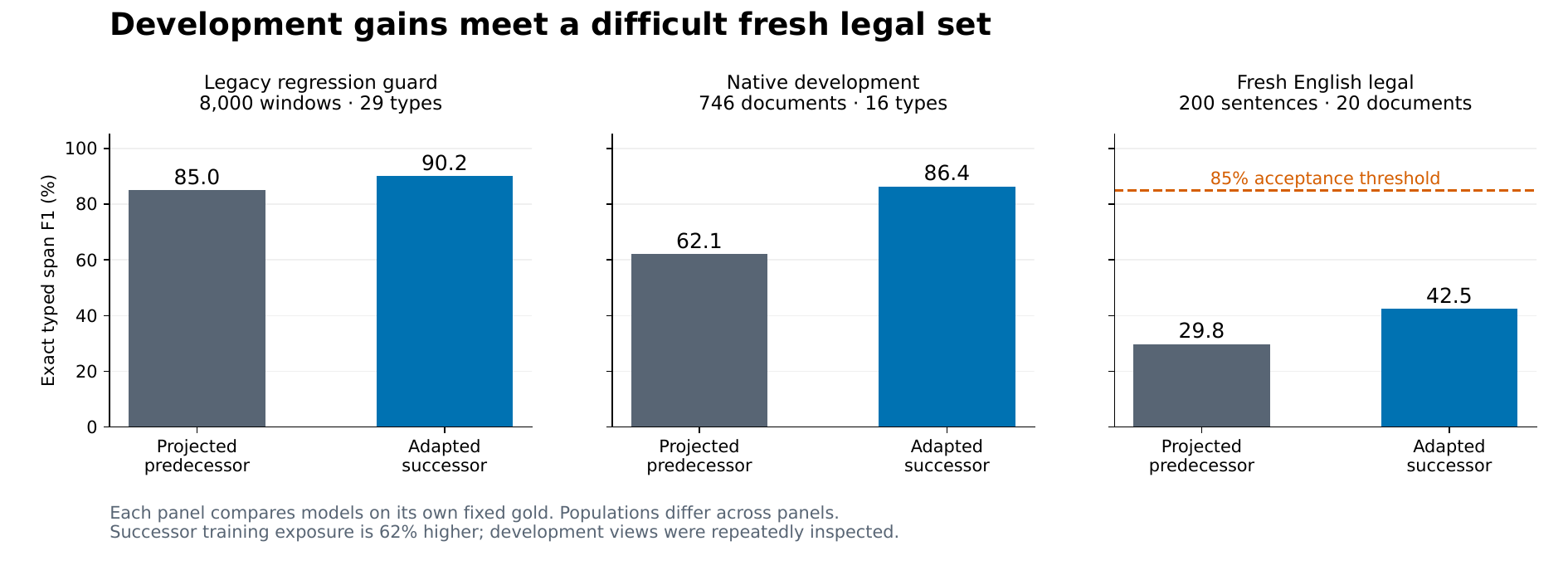}
\else
\includegraphics[alt={Ont2 adapted successor scores 90.2 on the older Final20 retention set, 86.4 on clinical-heavy development and 42.5 on fresh English legal text.}]{figures/ont2-development-confirmation-v1.pdf}
\fi

}

\caption{\label{fig-development-confirmation}Ont2 development accuracy
and fresh legal evaluation. Exact typed span F1 is measured separately
on each population. The clinical-heavy MEDDOCAN, MultiGraSCCo and TAB
development set covers 16 types; the older Final20 retention set covers
29. Fresh English HUDOC legal text reveals a large gap despite high
development scores.}

\end{figure*}

Ont2 reached 86.4\% exact typed-span F1 on 746 development documents
from MEDDOCAN, MultiGraSCCo and TAB, compared with 32.9\% for stock
GLiNER2 under the Ont2 projection. This encouraging result covered 16
types, with much of the data clinical; TAB contributed court-case text.
On new court documents from HUDOC, the European Court of Human Rights
case-law database, its accuracy fell to 42.5\% on 200 segments from 20
judgments (Figure~\ref{fig-development-confirmation}). Even new document
types from a court already represented in training exposed a large
remaining gap.

The training mixture supplemented those clinical and legal corpora with
translations and web text labeled by a 31-billion-parameter model. The
fresh HUDOC judgments were labeled directly in Ont2, whereas TAB's
labels reached Ont2 through a mapping. They were also newer judgments
sampled from every court section, unlike TAB's case collection. We
suspected annotation-inventory mismatch contributed most to the gap,
followed by document type and insufficient training data; these
explanations were not separately tested.

Checked translations and inexpensive local-model annotations had
improved our earlier recipe, but had not solved this harder legal task.
We therefore shifted effort toward broader, directly annotated text and
stronger frontier annotators, including Luna. Ont3 intake retained
quality-checked, low-temperature translations alongside
original-language annotations; subsequent expansion and mapped-gold
replay produced O3 and O4 (Section~\ref{sec-ontology-comparison};
Section~\ref{sec-o4-annotation-recipe}).

\subsubsection{Shared-label comparison on
Silver-dev}\label{shared-label-comparison-on-silver-dev}

In a precursor comparison separating evaluation difficulty from the
label inventory, we applied Ont2 and an Ont3 model trained before
mapped-gold replay was added to the same sentences of Silver-dev,
without neighboring context, and scored only their 29 shared primary
types. On the twenty languages supported by both models, Ont3 reaches
73.2\% exact typed span F1 versus Ont2's 52.5\%; character F1 is 91.6\%
versus 76.6\% (Table~\ref{tbl-ontology-comparison}). With the scored
labels held fixed, the principal change in difficulty is the more varied
evaluation text. The two additional reference labels form a separate
task only when a score explicitly requires them.

\ACLwidetabletrue\ACLcontinuedtablefalse

\ifXeTeX
\begin{longtable}[]{@{}
  >{\raggedright\arraybackslash}p{(\linewidth - 8\tabcolsep) * \real{0.1579}}
  >{\raggedleft\arraybackslash}p{(\linewidth - 8\tabcolsep) * \real{0.2105}}
  >{\raggedleft\arraybackslash}p{(\linewidth - 8\tabcolsep) * \real{0.2105}}
  >{\raggedleft\arraybackslash}p{(\linewidth - 8\tabcolsep) * \real{0.2105}}
  >{\raggedleft\arraybackslash}p{(\linewidth - 8\tabcolsep) * \real{0.2105}}@{}}
\caption{Ont2 and the pre-replay, sentence-only Ont3 model on identical
development text, with optional references. The shared-language
difference is +20.68 span-F1 points (95\% paired interval {[}+16.24,
+25.20{]}).}\label{tbl-ontology-comparison}\tabularnewline
\toprule\noalign{}
\begin{minipage}[b]{\linewidth}\raggedright
Model
\end{minipage} & \begin{minipage}[b]{\linewidth}\raggedleft
Languages
\end{minipage} & \begin{minipage}[b]{\linewidth}\raggedleft
Sentences
\end{minipage} & \begin{minipage}[b]{\linewidth}\raggedleft
Exact typed span F1 (\%)
\end{minipage} & \begin{minipage}[b]{\linewidth}\raggedleft
Character F1 (\%)
\end{minipage} \\
\midrule\noalign{}
\endfirsthead
\toprule\noalign{}
\begin{minipage}[b]{\linewidth}\raggedright
Model
\end{minipage} & \begin{minipage}[b]{\linewidth}\raggedleft
Languages
\end{minipage} & \begin{minipage}[b]{\linewidth}\raggedleft
Sentences
\end{minipage} & \begin{minipage}[b]{\linewidth}\raggedleft
Exact typed span F1 (\%)
\end{minipage} & \begin{minipage}[b]{\linewidth}\raggedleft
Character F1 (\%)
\end{minipage} \\
\midrule\noalign{}
\endhead
\bottomrule\noalign{}
\endlastfoot
Ont2 & Shared 20 & 374 & 52.52 & 76.63 \\
Ont3 & Shared 20 & 374 & 73.21 & 91.58 \\
Ont2 & 35 & 659 & 51.33 & 75.09 \\
Ont3 & 35 & 659 & 72.89 & 91.17 \\
\end{longtable}
\else
\begin{longtable}[]{@{}
  >{\raggedright\arraybackslash}X
  >{\raggedleft\arraybackslash}X
  >{\raggedleft\arraybackslash}X
  >{\raggedleft\arraybackslash}X
  >{\raggedleft\arraybackslash}X@{}}
\caption{Ont2 and the pre-replay, sentence-only Ont3 model on identical
development text, with optional references. The shared-language
difference is +20.68 span-F1 points (95\% paired interval {[}+16.24,
+25.20{]}).}\label{tbl-ontology-comparison}\tabularnewline
\toprule\noalign{}
\raggedright
Model
 & \raggedleft
Languages
 & \raggedleft
Sentences
 & \raggedleft
Exact typed span F1 (\%)
 & \raggedleft
Character F1 (\%)
 \\
\midrule\noalign{}
\endfirsthead
\toprule\noalign{}
\raggedright
Model
 & \raggedleft
Languages
 & \raggedleft
Sentences
 & \raggedleft
Exact typed span F1 (\%)
 & \raggedleft
Character F1 (\%)
 \\
\midrule\noalign{}
\endhead
\bottomrule\noalign{}
\endlastfoot
Ont2 & Shared 20 & 374 & 52.52 & 76.63 \\
Ont3 & Shared 20 & 374 & 73.21 & 91.58 \\
Ont2 & 35 & 659 & 51.33 & 75.09 \\
Ont3 & 35 & 659 & 72.89 & 91.17 \\
\end{longtable}
\fi

The 374-sentence subset contains 272 source documents; the complete set
contains 468. Scores pool counts across rows; intervals use 10,000
paired document resamples. Both encoders use the same sentence-only
input, legal BIOES decoding and adjacent-name joining, with no learned
character-boundary refiner or name/pattern postprocessors. This
comparison reuses development data. Optional references are scored under
the rule in Section~\ref{sec-span-views}.

\subsection{Character-boundary
refinement}\label{sec-boundary-refiner-details}

Training for the character-endpoint model of
Section~\ref{sec-character-boundaries}, which moves span endpoints to
correct encoder token boundaries, used TAB, MEDDOCAN and MultiGraSCCo,
with language caps yielding 23,270 endpoint-training groups. We selected
the boundary model on 601 documents sharing no source documents or exact
texts with training. We applied it to eleven trained languages (ar, de,
en, es, fa, fr, it, pl, ru, tr, uk); other languages retained the BIOES
endpoints.

The main-text curves (Section~\ref{sec-priority9-comparison}) use a
35-language continuation of this refiner. It keeps the endpoint features
and weights, adds seven span-pair features that score a candidate's
start and end jointly, and never trims letters or digits; the radius
stays ±1 character with the same movement penalty, and it remains a
small hashed linear model. It was trained on candidates around real
tagger proposals in 10,537 segments, which share no exact or
whitespace-normalized text with either evaluation population (gold and
Silver-dev). It was selected among fitted variants partly on the
659-segment Silver-dev; fitted ±2 variants never beat ±1. Relative to
the eleven-language refiner, the historical best redaction F1 at 80\%
overlap of O3 moves by +0.04 on gold and +2.3 over all 35 Ont3 languages
(+0.6 and +3.2 with exact boundaries).

An English-only fixed-development contrast kept the token logits and
legal BIOES path fixed and instead realized endpoints at Unicode word
edges covered by each token. This deterministic rule raised exact typed
F1 on the MEDDOCAN / MultiGraSCCo / TAB evaluation from 78.0\% to
84.7\%, near the character refiner's 85.2\%, but scored 84.9\% versus
88.0\% on the older Final20 windows. Exact-source-text bootstrap
supports the gain over unrefined BIOES on the former and the loss to the
refiner on the latter; the 0.49-point gap to the refiner on the former
is unresolved. Inspected rows show the word rule dropping terminal date
punctuation correctly while truncating punctuation-bearing identifiers
and names. The comparison favors the learned character module over a
fixed word rule; trained word pooling and multilingual word segmentation
were not compared.

\subsubsection{O4 contrast}\label{o4-contrast}

We reuse O4's token predictions at zero outside-label bias, removing
only endpoint adjustment. Both arms exclude optional reference outputs
and apply the same name and regex postprocessors. The refined arm
reproduces the main comparison's counts exactly.

\ACLwidetabletrue\ACLcontinuedtablefalse

\ifXeTeX
\begin{longtable}[]{@{}
  >{\raggedright\arraybackslash}p{(\linewidth - 8\tabcolsep) * \real{0.2500}}
  >{\raggedright\arraybackslash}p{(\linewidth - 8\tabcolsep) * \real{0.2100}}
  >{\raggedleft\arraybackslash}p{(\linewidth - 8\tabcolsep) * \real{0.1300}}
  >{\raggedleft\arraybackslash}p{(\linewidth - 8\tabcolsep) * \real{0.1300}}
  >{\raggedleft\arraybackslash}p{(\linewidth - 8\tabcolsep) * \real{0.2800}}@{}}
\caption{O4 without and with the ±1-character boundary-refinement model.
F1 and differences are percentage points; intervals use 10,000 paired
document resamples on reused development populations. The CPU replay and
scoring took 18 seconds. }\label{tbl-o4-boundary}\tabularnewline
\toprule\noalign{}
\begin{minipage}[b]{\linewidth}\raggedright
Population
\end{minipage} & \begin{minipage}[b]{\linewidth}\raggedright
Exact-match task
\end{minipage} & \begin{minipage}[b]{\linewidth}\raggedleft
Without model
\end{minipage} & \begin{minipage}[b]{\linewidth}\raggedleft
With model
\end{minipage} & \begin{minipage}[b]{\linewidth}\raggedleft
ΔF1 (95\% interval)
\end{minipage} \\
\midrule\noalign{}
\endfirsthead
\toprule\noalign{}
\begin{minipage}[b]{\linewidth}\raggedright
Population
\end{minipage} & \begin{minipage}[b]{\linewidth}\raggedright
Exact-match task
\end{minipage} & \begin{minipage}[b]{\linewidth}\raggedleft
Without model
\end{minipage} & \begin{minipage}[b]{\linewidth}\raggedleft
With model
\end{minipage} & \begin{minipage}[b]{\linewidth}\raggedleft
ΔF1 (95\% interval)
\end{minipage} \\
\midrule\noalign{}
\endhead
\bottomrule\noalign{}
\endlastfoot
Gold-7, 1,283 segments & Redaction regions & 87.87 & 87.80 & −0.07
{[}−0.23, 0.00{]} \\
Silver, 1,201 segments & Redaction regions & 77.71 & 80.20 & +2.49
{[}+1.58, +3.50{]} \\
Silver, 1,201 segments & Fine typed spans & 74.62 & 76.30 & +1.68
{[}+1.01, +2.41{]} \\
\end{longtable}
\else
\begin{longtable}[]{@{}
  >{\raggedright\arraybackslash}X
  >{\raggedright\arraybackslash}X
  >{\raggedleft\arraybackslash}X
  >{\raggedleft\arraybackslash}X
  >{\raggedleft\arraybackslash}X@{}}
\caption{O4 without and with the ±1-character boundary-refinement model.
F1 and differences are percentage points; intervals use 10,000 paired
document resamples on reused development populations. The CPU replay and
scoring took 18 seconds. }\label{tbl-o4-boundary}\tabularnewline
\toprule\noalign{}
\raggedright
Population
 & \raggedright
Exact-match task
 & \raggedleft
Without model
 & \raggedleft
With model
 & \raggedleft
ΔF1 (95\% interval)
 \\
\midrule\noalign{}
\endfirsthead
\toprule\noalign{}
\raggedright
Population
 & \raggedright
Exact-match task
 & \raggedleft
Without model
 & \raggedleft
With model
 & \raggedleft
ΔF1 (95\% interval)
 \\
\midrule\noalign{}
\endhead
\bottomrule\noalign{}
\endlastfoot
Gold-7, 1,283 segments & Redaction regions & 87.87 & 87.80 & −0.07
{[}−0.23, 0.00{]} \\
Silver, 1,201 segments & Redaction regions & 77.71 & 80.20 & +2.49
{[}+1.58, +3.50{]} \\
Silver, 1,201 segments & Fine typed spans & 74.62 & 76.30 & +1.68
{[}+1.01, +2.41{]} \\
\end{longtable}
\fi

Refinement changes two Gold-7 examples and 58 Silver examples.
Inspection finds both helpful punctuation removal from identifiers and
harmful endpoint changes, including a Korean occupation span extended
into a grammatical particle. The result supports a benefit on Silver,
not a general improvement across annotation conventions.

The original Ont2 audit found that 92\% of gold spans were reachable at
token boundaries. Exact typed F1 rose from 80.0 to 86.4 with refinement;
an oracle choosing among the same ±1 candidates reached 87.8. Tokens
averaged 3.4 characters, and 2,242 of 2,247 gold endpoints inside tokens
were one character from an edge. These historical diagnostics motivated
the small search radius.

\subsection{Ont3 context and data studies}\label{sec-ont3-historical}

The precursor Ont3 experiments below predate O4. Compare effects within
each matched study; the newer annotation-rich comparisons are identified
separately.

Sweeping the outside-label bias across 35 languages did not identify a
need to fit separate per-language operating points. This descriptive
check used only 15--22 sentences per language and required reference
labels (Section~\ref{sec-span-views}).

\subsubsection{Neighboring-sentence
context}\label{sec-neighboring-context}

Context experiments preserve the target sentence and fill the remaining
space in a 512-token window with preceding text, or with preceding and
following text sharing that space. Only target tokens receive tagging
loss.

Before the annotated training data grew, a frozen-encoder experiment
fitted heads on 16,634 sentences with matched additional exposure, about
2.96 million target tokens per arm. On 545 development sentences from
453 documents, excluding reference labels, exact typed F1 was 71.87 for
isolated input, 73.16 with the preceding sentence and 73.23 with both
neighbors. The apparent gains were +1.29 (paired 95\% interval {[}+0.07,
+2.53{]}) and +1.36 {[}+0.09, +2.68{]}; their difference was unresolved.
Unfreezing the top four blocks gave 72.9. These precursor results did
not require updating the encoder on contextual inputs.

A separate fully fine-tuned comparison on 659 development segments,
requiring non-bare reference labels, scored 70.10 for isolated training,
69.26 for preceding-context training and 69.82 for both-neighbor
training. Its context differences were unresolved. The different
reference policies prevent comparing absolute scores with the
frozen-encoder study.

For O3, a 2,000-update continuation mixed isolated, preceding-only and
both-neighbor inputs equally. Isolated-input Silver-dev redaction F1
improved by 1.1 points (interval {[}+0.01, +2.18{]}) with no detected
full-context regression. There was no matched additional-exposure
control, and another 2,000 updates did not resolve a further gain.

With the richer training data, the selected isolated-trained O4 model
loses 0.44 points on the weighted selection criterion when given the
preceding sentence (paired interval {[}−0.90, 0.00{]}); the two-seed
weight average loses 0.34 {[}−0.81, +0.12{]}. Other checked seeds
likewise give no consistent inference gain. A continuation mixing
contextual and isolated training loses 0.30 points under isolated
inference; a longer context-only continuation peaks at an unresolved
+0.16 before declining. These bounded comparisons support isolated input
for O4.

\subsubsection{Annotated training-data
size}\label{annotated-training-data-size}

\begin{figure*}[tp]

\centering{

\ifXeTeX
\includegraphics[width=1\linewidth,height=6in,alt={Required-reference F1 increases from 67.31 with 11,746 annotated windows to 69.01 with 23,183 and 69.26 with 47,641.}]{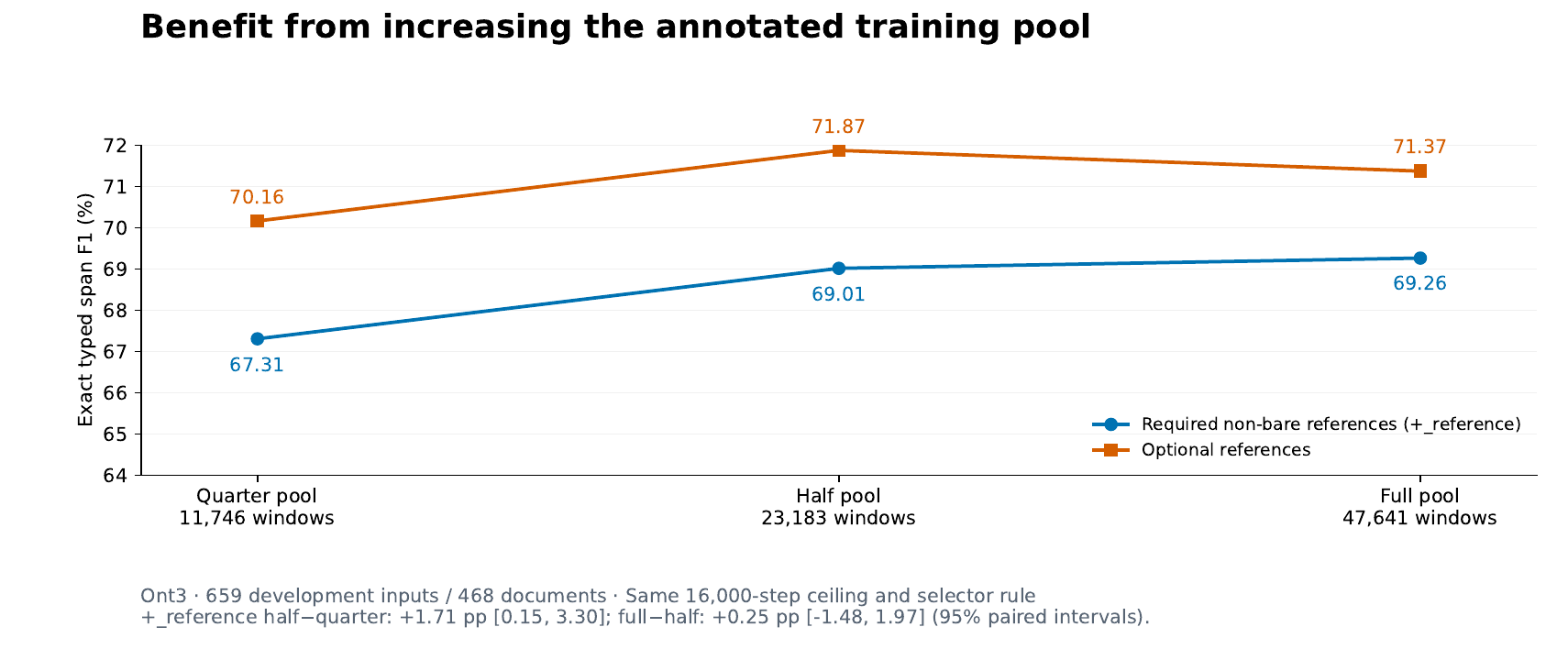}
\else
\includegraphics[alt={Required-reference F1 increases from 67.31 with 11,746 annotated windows to 69.01 with 23,183 and 69.26 with 47,641.}]{figures/ont3-label-support-curve-v1.pdf}
\fi

}

\caption{\label{fig-label-support}Effect of annotated training-data size
on 659 development rows from 468 source documents, with non-bare
references retained in the full view. Quarter and half support are
nested document subsets; language weights are adjusted to meet the same
coverage floors. Every arm uses the same optimization ceiling and
selector rule. Lines connect the three observed training-data sizes. The
intervals concern paired +\texttt{\_\allowbreak{}reference} differences;
they measure resampling uncertainty on reused development data.}

\end{figure*}

To measure the benefit of more annotated training data, we fine-tune all
layers of the large encoder on nested training sets of 11,746, 23,183
and 47,641 windows, with the same 16,000-update ceiling
(Figure~\ref{fig-label-support}). Increasing the training data from
quarter to half size improves +\texttt{\_\allowbreak{}reference} exact
span F1 on the 659-row development set by 1.7 points. Increasing it from
half to full size adds 0.25 points; the paired interval includes zero.

\subsection{Annotation volume, sampling and
cost}\label{sec-o4-annotation-recipe}

O4 continues the isolated-input branch of the Ont3 encoder for 4,000
updates on the expanded, title-corrected training mixture, with 50\% of
sampling mass assigned to mapped gold. Its architecture and typed
objective are unchanged. The selected single checkpoint is seed
20260927; the two-seed weight average differs by only 0.04 points on the
selection criterion, so we retain the simpler single-model lineage. The
criterion pools language-weighted counts from fine typed Silver-dev
(18\%), exact Silver-dev redaction regions (12\%) and gold regions
(70\%), using the 659 Silver-dev rows and 23,251 gold test examples
repeatedly used in development. For selection, optional reference labels
are excluded. The paper's fine-span scores additionally exempt a
named-entity prediction matching an optional reference at exactly the
same offsets; this does not change the selected model.

Continuing training beyond the selected 4,000 updates did not improve
the recipe.

The expanded training data contains 90,339 silver windows and 80,829
mapped-gold windows. Exact deduplication by language and text leaves
88,512 and 80,099 respectively. Compared with O3, 49,812 silver texts
are new, containing 1,652,949 XLM-R subword tokens, of which 378,431
intersect an annotated span. No mapped-gold texts are added. Annotation
changes affect 117 existing silver and 131 mapped-gold texts, mainly
title conventions; these changes do not add texts. Counts concern
sentence-oriented windows, not source documents; repeated draws do not
multiply annotation volume. Optional reference spans count as
supervision in these training totals.

We sought annotation candidates in two ways. Domain-near retrieval ranks
crawl paragraphs by similarity to examples from the evaluation domains,
subtracting their similarity to other candidate paragraphs. This
correction downweights generic text that is near everything, rather than
simply taking the nearest neighbors. Rare-entity ``needles'' instead
match surface patterns, sometimes with checksum checks or nearby cue
words. A hit selects the whole paragraph, including sentences without a
hit, for annotation in context. The matches select candidates rather
than supply gold labels. Source-bucket separation, used-text tracking
and lexical/semantic overlap screening precede annotation and admission
to training.

The annotation unions' receipts identify 50,617 source requests and
52,631 recorded response attempts, including format/session retries and
rejected annotations. Every recorded attempt has token usage. The
original teachers are GPT-5.6-Luna (37,595 requests) and GPT-6-Luna
(13,022); repricing them does not change that provenance. At the
standard
\href{https://developers.openai.com/api/docs/models/gpt-6-luna}{GPT-6-Luna
rates} retrieved September 29, 2026, one million tokens cost \$0.10 for
uncached input, \$0.01 for cached input, \$0.125 for cache writes and
\$0.50 for output. Applying the documented long-context multipliers per
attempt gives \textbf{\$52.66} in estimated replacement
charges.\footnote{Usage includes language-specific instructions and
  demonstrations, sentence inputs, generated outputs and repeated
  session history, charging cache reads and writes at their respective
  rates. Prompt caching and consecutive same-document sentence sessions
  amortize the prompt. Saved retries while finding usable, economical
  session lengths are included; growing history is counted on every
  attempt.} This uses the recorded token lengths and cache behavior, not
a counterfactual rerun or an invoice. Transport failures without
returned usage cannot be priced, and model training, data preparation
and evaluation are outside this annotation total. For Silver-dev, we
spent roughly 100 times as much per segment as for training annotation.

\begin{figure*}[tp]

\centering{

\ifXeTeX
\includegraphics[width=1\linewidth,height=6in,alt={Selection F1 is near 83 from 40 to 60 percent mapped-gold sampling. Gold F1 rises while typed Silver-dev changes little.}]{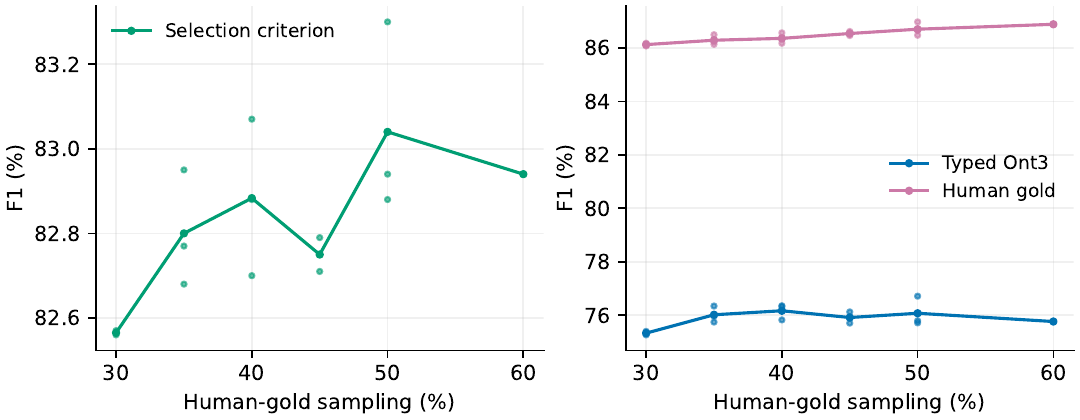}
\else
\includegraphics[alt={Selection F1 is near 83 from 40 to 60 percent mapped-gold sampling. Gold F1 rises while typed Silver-dev changes little.}]{figures/o4-gold-share.pdf}
\fi

}

\caption{\label{fig-o4-gold-share}Gold sampling share at 4,000 updates
on the expanded training data. Dots are individual seeds and lines
connect their means; no fitted trend or uncertainty band is implied. The
selection criterion changes little above 40\%, while higher shares favor
gold over fine Silver-dev. The 60\% setting has only one seed.}

\end{figure*}

Increasing mapped-gold sampling from 30\% to 40\% helps the combined
criterion; 40--60\% is comparatively flat
(Figure~\ref{fig-o4-gold-share}). There are two or three seeds per
setting except 60\%, which has one. More mapped gold mainly trades fine
Silver-dev credit for gold credit. Croatian, Korean and Polish require
language boosts above about 42\% gold; Malay and Vietnamese also require
them at 50\%. All final recipes satisfy the trainer's 0.75\%
per-language floor.

The headline O3--O4 contrast includes annotation, sampling and
input-recipe changes. At fixed decoder bias, pooled fine typed F1
improves more than Silver redaction F1, whose gain remains unresolved.
Paired intervals use 10,000 document resamples within each population,
treating segments independently when their source document is unknown.
They condition on the selected recipes and do not account for the full
development search. The final 2,788-row annotation increment had no
measurable benefit over its predecessor training data in two matched
seeds. Neither that local result nor the sampling plateau determines the
asymptote of training on further data.

\subsection{Other tested approaches}\label{sec-failed-ideas}

\subsubsection{Context-conditioned entity
replacement}\label{sec-character-surface-generation}

When translating annotated text, entity placeholders can be filled with
the original values, corpus examples or generated replacements. We used
Faker for 43\% of placeholder occurrences in the routed translation
training data (Section~\ref{sec-locale-repairs}). Replacing entities can
also vary untranslated training text, but a replacement must fit the
sentence as well as its tag. We explored drawing observed values and
generating them with character models trained on the same corpus, using
sentence context to guide both approaches.

For observed donors, we compared uniform draws within a language and
type with retrieval by masked sentence context. On 140 queries across 35
languages, advisory reviews accepted 97 uniform draws and 107
context-retrieved draws. Character generators conditioned on nearby
text, language and type received 58--63 accepts on the same queries.
Increasing model size from 2.1 to 9.9 million parameters, training
longer and weighting surfaces by corpus frequency did not resolve
malformed values or poor grammatical and semantic fit. These were
reviews of replacement quality, not tagger training results.

A separate experiment included whole-span categories, ordered
name-component kinds and binary predicate values in the generator's
conditioning. Adding a frozen XLM\nobreak\hbox{-}\nobreak{}R encoder of
the masked sentence reduced mean character negative log likelihood from
1.85 to 1.77 on 489 source-disjoint entity sites, but outputs still
violated type and refinement constraints. Predicate conditioning
recorded presence within a span, not the locations of its subspans.
Using the full pattern of span refinements and subspan predicates could
help reject incompatible replacements. The larger native-corpus fits
above did not use refinement conditioning.

We also tested replacement through tagger training. Two Ont3 models
trained for 16,000 updates on the same six training sets. One kept the
original text; the other replaced organization names with observed
donors at probability 0.5 whenever an example was sampled. On 818
development segments, exact organization F1 fell from 75.1 to 68.2, with
more false positives and boundary errors. We therefore kept original
names in that recipe. A shorter screen starting from O3 also found no
benefit. We continued O3 for 1,000 updates while replacing eligible
organization names with observed same-language donors, matched only by
type, at probability 0, 0.25 or 0.5. At the unchanged decoder bias,
redaction F1 stayed within 0.6 points of the probability-0 control on
both Silver-dev and Gold-7, and every paired 95\% interval included
zero. At probability 0.5, Silver-dev organization predictions rose from
343 to 366, while correct ones rose only from 250 to 255. This is
consistent with some reliance on memorized organization names, though
replacing names can also disrupt their fit to the sentence.

Our positive result was narrower: repairing systematic name-order and
date-format errors in translated training text improved tagging
(Section~\ref{sec-locale-repairs}). Replacing intact values in
original-language Ont3 text did not help, so the final recipe keeps
original surfaces. Type-only donor matching did not help, and fully
conditioning a donor on its sentence and annotations leaves only the
original value.

\subsubsection{Consistency
regularization}\label{consistency-regularization}

R\nobreak\hbox{-}\nobreak{}Drop runs each labeled batch twice with
independent dropout masks and penalizes disagreement between the two
token-label distributions. We averaged the supervised losses and added a
symmetric Kullback--Leibler divergence at supervised token positions;
inference still used one pass. The aim was to make predictions less
sensitive to the random perturbations used in training.

Added with penalty coefficient 1.0 to an intermediate continuation,
R\nobreak\hbox{-}\nobreak{}Drop improved language-macro-averaged typed
overlap F1 on the seven-language Fresh7 suite
(Section~\ref{sec-evaluation-sets}) by 0.6 points. Added with the same
coefficient to a later baseline recipe, it reduced application-weighted
typed overlap F1 by 0.5 points (95\% interval {[}−0.80, −0.20{]}) on the
1,267 Fresh20 development documents in twenty languages. The
seven-language gain therefore did not justify adopting this setting in
the later recipe.

\subsubsection{Learned register positions}\label{sec-registers}

We added eight learned input positions that every token can attend to
but that are never scored, each carrying a vector added at every encoder
layer. Against matched controls trained from the same pretrained
encoder, exact typed span F1 on the 1,113 Ont3 development segments
changed by +0.37, +0.61 and −1.44 points across three seeds. We did not
adopt them.

On the expanded training data, eight all-layer registers change the
weighted selection criterion by +0.12 points at 4,000 updates; 32
registers change it by −0.24. Four all-layer registers on O4's full
training data likewise give no resolved gain across four paired seeds
(typed F1 −0.02 points, 95\% interval {[}−0.70, +0.69{]}; matched
controls in Table~\ref{tbl-soft-prompt-controls}). Adding eight
registers to the 40\%-mapped-gold recipe changes it by −0.16. These
differences are small relative to the observed seed variation and do not
support adding registers to O4.

\subsubsection{Targeted annotation}\label{targeted-annotation}

Retrieving annotation candidates toward the evaluation domains with
multilingual sentence embeddings did not beat random intake from the
same languages in matched two-seed comparisons. The last 2,788 added
rows changed the weighted criterion by +0.09 and −0.25 points. More
annotation helped earlier in the expansion, but this final increment did
not resolve a further gain.

\subsubsection{Redaction-loss weighting}\label{redaction-loss-weighting}

Adding a binary entity-versus-outside loss to the typed objective
changed the criterion by −0.10 at weight 0.5 and −0.32 at weight 2. The
larger weight improved Silver-dev redaction regions while losing gold
and fine typed accuracy. These are bounded negative results, with
single-seed changes below roughly half a point difficult to distinguish
from the observed seed variation.

\subsubsection{Weight versus logit
averaging}\label{weight-versus-logit-averaging}

We compared averaging checkpoint weights into a single model (a
``soup'') with averaging the models' token logits before decoding. The
latter retains multiple encoder passes at inference; a soup uses one. We
combined separate training seeds, neighboring checkpoints from one
trajectory, and a continuation with its ancestor. Neither approach
established a reliable gain over the better member; mixing in the
ancestor could hurt. Later multi-seed soups likewise offered no clear
advantage over the selected single O4 checkpoint, which we retained for
its simpler lineage.

\subsubsection{Category-balanced
sampling}\label{category-balanced-sampling}

We tried sampling more examples containing underrepresented labels, but
abandoned this approach after finding no clear benefit. We instead favor
annotating more text.

\subsubsection{Soft prompt tuning}\label{sec-prompt-controls}

Soft prompts prepend learned vectors to the text. With the encoder
frozen, we trained these vectors and the tagging head: ten shared
tokens, or eight shared tokens plus two language-specific tokens.
Separate experiments add register vectors at every encoder layer
(Section~\ref{sec-registers}).

Type-specific prompts use \textbf{?}, \textbf{+} and \textbf{−}
embeddings for unknown, present and absent status. For example, a
sentence containing a person and an organization but no location
receives \textbf{?PER ?ORG ?LOC +PER +ORG −LOC} during training and only
\textbf{?PER ?ORG ?LOC} at inference. Each symbol--type pair denotes its
own learned vector.

The first five arms start from the same PII-trained Ont3 encoder and fit
the affine head for eight epochs on the same 1,076 training sentences.
Evaluation reused 545 development sentences from 453 documents in 35
languages, containing 1,026 reference spans. Supervision of Ont3's
\texttt{\_\allowbreak{}reference} labels was disabled. Decoding is
constrained to legal BIOES label sequences. Neither of the two
shared-token prompt arms improves on the no-prompt arm; both paired
intervals include zero.

\ACLwidetabletrue\ACLcontinuedtablefalse

\ifXeTeX
\begin{longtable}[]{@{}
  >{\raggedright\arraybackslash}p{(\linewidth - 8\tabcolsep) * \real{0.3457}}
  >{\raggedleft\arraybackslash}p{(\linewidth - 8\tabcolsep) * \real{0.1111}}
  >{\raggedleft\arraybackslash}p{(\linewidth - 8\tabcolsep) * \real{0.1728}}
  >{\raggedleft\arraybackslash}p{(\linewidth - 8\tabcolsep) * \real{0.2593}}
  >{\raggedleft\arraybackslash}p{(\linewidth - 8\tabcolsep) * \real{0.1111}}@{}}
\caption{Exact typed-span soft-prompt controls. The first five rows use
545 development sentences, with \texttt{\_\allowbreak{}reference}
supervision and scoring disabled. The O4 rows average four paired
training seeds on all 1,201 Silver inputs, with optional references and
O4's serving chain. Compare ΔF1 within each block; absolute scores have
different populations and training conditions. ΔF1 and paired 95\%
confidence intervals are relative to no prompt, in F1 points. During
training, each entity type receives a \textbf{?} token and a
gold-derived \textbf{+} or \textbf{−} token; at inference, only the
\textbf{?} tokens are visible to the encoder. The 58-token control uses
learned tokens independent of gold status, with only the first 29
visible at inference. For the 545-input control, its interval versus no
prompt is not reported here. Times cover forward computation and
decoding of all 545 sentences; O4 timing was not measured under those
conditions. O4 intervals resample both documents and the four paired
seeds, 10,000 times. }\label{tbl-soft-prompt-controls}\tabularnewline
\toprule\noalign{}
\begin{minipage}[b]{\linewidth}\raggedright
Prompt
\end{minipage} & \begin{minipage}[b]{\linewidth}\raggedleft
F1 (\%)
\end{minipage} & \begin{minipage}[b]{\linewidth}\raggedleft
ΔF1 (points)
\end{minipage} & \begin{minipage}[b]{\linewidth}\raggedleft
Paired 95\% CI
\end{minipage} & \begin{minipage}[b]{\linewidth}\raggedleft
Time (s)
\end{minipage} \\
\midrule\noalign{}
\endfirsthead
\toprule\noalign{}
\begin{minipage}[b]{\linewidth}\raggedright
Prompt
\end{minipage} & \begin{minipage}[b]{\linewidth}\raggedleft
F1 (\%)
\end{minipage} & \begin{minipage}[b]{\linewidth}\raggedleft
ΔF1 (points)
\end{minipage} & \begin{minipage}[b]{\linewidth}\raggedleft
Paired 95\% CI
\end{minipage} & \begin{minipage}[b]{\linewidth}\raggedleft
Time (s)
\end{minipage} \\
\midrule\noalign{}
\endhead
\bottomrule\noalign{}
\endlastfoot
No prompt & 72.53 & 0 & --- & 2.64 \\
10 shared tokens & 71.65 & −0.880 & {[}−2.133, +0.394{]} & 2.59 \\
8 shared + 2 language tokens & 71.94 & −0.587 & {[}−2.012, +0.848{]} &
2.59 \\
58 learned tokens, 29 at inference & 73.16 & +0.63 & --- & 2.91 \\
Gold status: 58 active in train, 29 at inference & 73.34 & +0.807 &
{[}−0.514, +2.210{]} & 2.95 \\
O4: no prompt & 75.74 & 0 & --- & --- \\
O4: 58 learned, 29 at inference & 75.48 & −0.26 & {[}−1.11, +0.59{]} &
--- \\
O4: \textbf{? + −}, 58 active, 29 at inference & 75.61 & −0.13 &
{[}−0.86, +0.64{]} & --- \\
\end{longtable}
\else
\begin{longtable}[]{@{}
  >{\raggedright\arraybackslash}X
  >{\raggedleft\arraybackslash}X
  >{\raggedleft\arraybackslash}X
  >{\raggedleft\arraybackslash}X
  >{\raggedleft\arraybackslash}X@{}}
\caption{Exact typed-span soft-prompt controls. The first five rows use
545 development sentences, with \texttt{\_\allowbreak{}reference}
supervision and scoring disabled. The O4 rows average four paired
training seeds on all 1,201 Silver inputs, with optional references and
O4's serving chain. Compare ΔF1 within each block; absolute scores have
different populations and training conditions. ΔF1 and paired 95\%
confidence intervals are relative to no prompt, in F1 points. During
training, each entity type receives a \textbf{?} token and a
gold-derived \textbf{+} or \textbf{−} token; at inference, only the
\textbf{?} tokens are visible to the encoder. The 58-token control uses
learned tokens independent of gold status, with only the first 29
visible at inference. For the 545-input control, its interval versus no
prompt is not reported here. Times cover forward computation and
decoding of all 545 sentences; O4 timing was not measured under those
conditions. O4 intervals resample both documents and the four paired
seeds, 10,000 times. }\label{tbl-soft-prompt-controls}\tabularnewline
\toprule\noalign{}
\raggedright
Prompt
 & \raggedleft
F1 (\%)
 & \raggedleft
ΔF1 (points)
 & \raggedleft
Paired 95\% CI
 & \raggedleft
Time (s)
 \\
\midrule\noalign{}
\endfirsthead
\toprule\noalign{}
\raggedright
Prompt
 & \raggedleft
F1 (\%)
 & \raggedleft
ΔF1 (points)
 & \raggedleft
Paired 95\% CI
 & \raggedleft
Time (s)
 \\
\midrule\noalign{}
\endhead
\bottomrule\noalign{}
\endlastfoot
No prompt & 72.53 & 0 & --- & 2.64 \\
10 shared tokens & 71.65 & −0.880 & {[}−2.133, +0.394{]} & 2.59 \\
8 shared + 2 language tokens & 71.94 & −0.587 & {[}−2.012, +0.848{]} &
2.59 \\
58 learned tokens, 29 at inference & 73.16 & +0.63 & --- & 2.91 \\
Gold status: 58 active in train, 29 at inference & 73.34 & +0.807 &
{[}−0.514, +2.210{]} & 2.95 \\
O4: no prompt & 75.74 & 0 & --- & --- \\
O4: 58 learned, 29 at inference & 75.48 & −0.26 & {[}−1.11, +0.59{]} &
--- \\
O4: \textbf{? + −}, 58 active, 29 at inference & 75.61 & −0.13 &
{[}−0.86, +0.64{]} & --- \\
\end{longtable}
\fi

The O4 runs update the full encoder on its complete training data for
4,000 steps, with a prompt learning rate of 0.0003, as in the small-data
status recipe. Against matched continuations, neutral and status prompts
change Gold-7 exact-boundary redaction F1 by −0.26 and −0.02 points on
1,283 inputs, and Silver exact-boundary redaction F1 by +0.26 and +0.09.
All paired intervals include zero.

With 29 entity types, the status arm has 29 \textbf{?} tokens and 29
\textbf{+}/\textbf{−} slots, masking the latter at inference. The
58-token control scores 73.2, leaving the status-specific difference
unresolved (+0.18 points, interval {[}−0.359, +0.758{]}).

For the 545-input study, intervals use 10,000 paired document resamples
and are unadjusted. Forward/decode timings cover all 545 examples at
batch width four on an L40S, excluding model loading.

With the richer training data, supervision-status slots again fail to
improve their matched neutral-slot control. At 4,000 updates, neutral
slots change the weighted selection criterion by +0.24 points relative
to the plain recipe; gold-status slots score 1.94 points below neutral
slots (paired 95\% interval {[}−2.62, −1.30{]}). Status is available
during training but hidden at inference. Inspected errors show drift
toward coarse gold conventions, including locality predictions replaced
by generic location and omitted health conditions. We found no clearly
helpful prompt recipe for our XLM\nobreak\hbox{-}\nobreak{}R tagger.
Prompt tuning may still be useful when full encoder training is too
expensive; that constraint does not apply here.

\subsubsection{Binary entity-presence
supervision}\label{sec-entity-presence}

In an Ont2 experiment using the
Gemma\nobreak\hbox{-}\nobreak{}4-31B-annotated training data
(Section~\ref{sec-teacher-omission-controls}), binary entity-presence
supervision trained the model to distinguish entity tokens from
background without requiring a particular entity type.
Positive-unlabeled training used known entity tokens and estimated
entity prevalence from fully annotated text instead of treating every
token the teacher left unmarked as background. At approximately equal
exposure to complete and positive-only annotations, this objective
raised native recall but lowered precision and F1. We therefore
continued to ignore unmarked tokens in this Gemma-annotated training
data.

\subsection{Serving measurement
conditions}\label{sec-serving-conditions}

The CPU throughput measurement in Table~\ref{tbl-serving-cost} used 100
English legal segments (3,250 tokens) on a shared AMD EPYC 7R13 host;
each figure is the median of three timed repetitions after one warmup,
excluding model loading. Both systems used model batch size 1;
XLM\nobreak\hbox{-}\nobreak{}R served 16 concurrent requests, while
GLiNER2 served one request with 16 intra-operation threads. The GPU
measurement used 487 documents in seven languages on an NVIDIA L40S, in
single-document batches, with one uncontended pass after eight warmup
documents per language.

\end{document}